\documentclass{article}

\usepackage{microtype}
\usepackage{graphicx}
\usepackage{booktabs} 

\usepackage{hyperref}

\usepackage{amsmath}
\usepackage{amssymb}
\usepackage{mathtools}
\usepackage{amsthm}

\usepackage[capitalize,noabbrev]{cleveref}

\theoremstyle{plain}
\newtheorem{theorem}{Theorem}[section]
\newtheorem{proposition}[theorem]{Proposition}
\newtheorem{lemma}[theorem]{Lemma}
\newtheorem{corollary}[theorem]{Corollary}
\theoremstyle{definition}
\newtheorem{definition}[theorem]{Definition}

\theoremstyle{remark}

\usepackage[textsize=tiny]{todonotes}

\usepackage{xspace}
\usepackage[noend]{algorithmic}
 \usepackage{caption}
 \usepackage{subcaption}
\usepackage{algorithm}%

\usepackage[accepted]{icml2024}

\usepackage{tikz}
\usetikzlibrary{positioning}
\tikzset{>=stealth}

\usetikzlibrary{positioning, calc, shapes.geometric, shapes, shapes.multipart, arrows.meta, arrows, decorations.markings, external, trees}

\tikzset{
    -Latex,auto,node distance =1 cm and 1 cm,semithick,
    state/.style ={ellipse, draw, minimum width = 0.7 cm},
    point/.style = {circle, draw, inner sep=0.04cm,fill,node contents={}},
    bidirected/.style={Latex-Latex,dashed},
    el/.style = {inner sep=2pt, align=left, sloped}
}
\tikzstyle{bag} = [align=center]
\usetikzlibrary{positioning}
\usepackage{graphicx}
\usepackage{wrapfig}

\usepackage{import}
\usetikzlibrary{positioning}
\usetikzlibrary{3d} 

\usetikzlibrary{arrows.meta}
\usetikzlibrary{math} 
\usepackage{listofitems} 
\tikzstyle{mynode}=[thick,draw=blue,fill=blue!20,circle,minimum size=22]
\tikzstyle{confnode}=[thick,draw=blue,fill=blue!50,circle,minimum size=22]

\usepackage{xstring}
\usetikzlibrary{fit,positioning}
\usetikzlibrary{shapes,decorations,arrows,calc,arrows.meta,fit,positioning}
\tikzset{
    edge/.style={-Latex,node distance =1 cm and 1 cm, semithick},
    state/.style ={ellipse, draw, minimum size = 1.5 cm},
    point/.style = {circle, draw, inner sep=0.04cm,fill,node contents={}},
    bidirected/.style={Latex-Latex,dashed},
    el/.style = {inner sep=2pt, align=left, sloped},
}
\usepackage{xcolor}

\newcommand{\Li}{$\mathcal{L}_1$\xspace}
\newcommand{\Lii}{$\mathcal{L}_2$\xspace}

\newcommand{\WG}{Modular-DCM\xspace}
\newcommand{\ABr}{DCM-Rep\xspace}

\newcommand{\emp}{\emptyset}
\newcommand{\IG}{InterFaceGAN\xspace}
\newcommand{\CLa}{\texttt{TrainDomain}\xspace}
\newcommand{\CLb}{\texttt{Intervention}\xspace}
\newcommand{\CLc}{\texttt{Augmented}\xspace}
\newcommand{\G}{\mathbb{G}\xspace}

\newcommand{\gray}{\textcolor{gray}}

\newcommand{\wh}{\textcolor{white}}
\newcommand{\nc}{rgb:red,1;green,2;blue,5}

\newcommand{\layer}{\mathcal{L}}
\newcommand{\doo}{\text{do}}

\newcommand{\indep}{\perp \!\!\! \perp}
\newcommand{\ID}{\rotatebox[origin=c]{-90}{$\perp$}}
\usepackage{bbm}

\usepackage{natbib}
\usepackage{microtype}

\icmltitlerunning{Modular Learning of Deep Causal Generative Models for High-dimensional Causal Inference}

\begin{document}

\twocolumn[
\icmltitle{Modular Learning of Deep Causal Generative Models \\for High-dimensional Causal Inference}



\icmlsetsymbol{equal}{*}

\begin{icmlauthorlist}
\icmlauthor{Md Musfiqur Rahman}{yyy}
\icmlauthor{Murat Kocaoglu}{yyy}
\end{icmlauthorlist}

\icmlaffiliation{yyy}{School of Electrical and Computer Engineering, Purdue University, West Lafayette, IN, USA}

\icmlcorrespondingauthor{Md Musfiqur Rahman}{rahman89@purdue.edu}

\icmlkeywords{Machine Learning, ICML}

\vskip 0.3in
]



\printAffiliationsAndNotice{}  

\begin{abstract}    


Sound and complete algorithms have been proposed to compute identifiable causal queries 
using the causal structure and data. However, most of these algorithms assume accurate estimation of the data distribution, which is impractical for high-dimensional variables such as images. On the other hand, modern deep generative architectures can be trained to sample from high-dimensional distributions. However, training these networks are typically very costly. Thus, it is desirable to leverage pre-trained models to answer causal queries using such high-dimensional data. To address this, we propose modular training of deep causal generative models that not only makes learning more efficient, but also allows us to utilize large, pre-trained conditional generative models. To the best of our knowledge, our algorithm, Modular-DCM is the first algorithm that, given the causal structure, uses adversarial training to learn the network weights, and can make use of pre-trained models to provably sample from any identifiable causal query in the presence of latent confounders. With extensive experiments on the Colored-MNIST dataset, we demonstrate that our algorithm outperforms the  baselines. We also show our algorithm's convergence on the COVIDx dataset and its utility with a causal invariant prediction problem on CelebA-HQ. 
\end{abstract}

\begin{figure}[t!]
	\centering
\begin{subfigure}{0.70\linewidth}
  \begin{tikzpicture}[
      observable/.style={circle, draw=none, fill=none, very thick, minimum size=7mm},
      scale=0.5, transform shape
      ]
      \tikzstyle{connection}=[ultra thick,every node/.style={sloped,allow upside down},draw=\edgecolor,opacity=0.7]
       \tikzstyle{every node}=[font=\Large]

      \node[align=left] (X) {Covid-19\\Symptom ($C$)};
      \node[align=left,right =0.9cm of X ] (Z) {Chest\\X-ray($X$)};
      \node[align=left,right =0.9cm of Z] (Y) {Pneumonia\\Diagnosis ($N$)} ;
      \node[rectangle,draw,align=left, fill=lightgray, above=1.5cm of Z] (U) {Hospital\\Location};
      \draw[->] (X.east) -- (Z.west);
      \draw[->] (Z.east) -- (Y.west);
      \draw[->] (U.west)to [out=185,in=80](X.north);
      \draw[->] (U.east)to [out=-5,in=100](Y.north);
\end{tikzpicture}
\caption{ Causal graph}
		\label{fig:xray-a}
	\end{subfigure}
	\hspace{1em}
\begin{subfigure}{1\linewidth}
    \centering
    \includegraphics[width=1\linewidth]{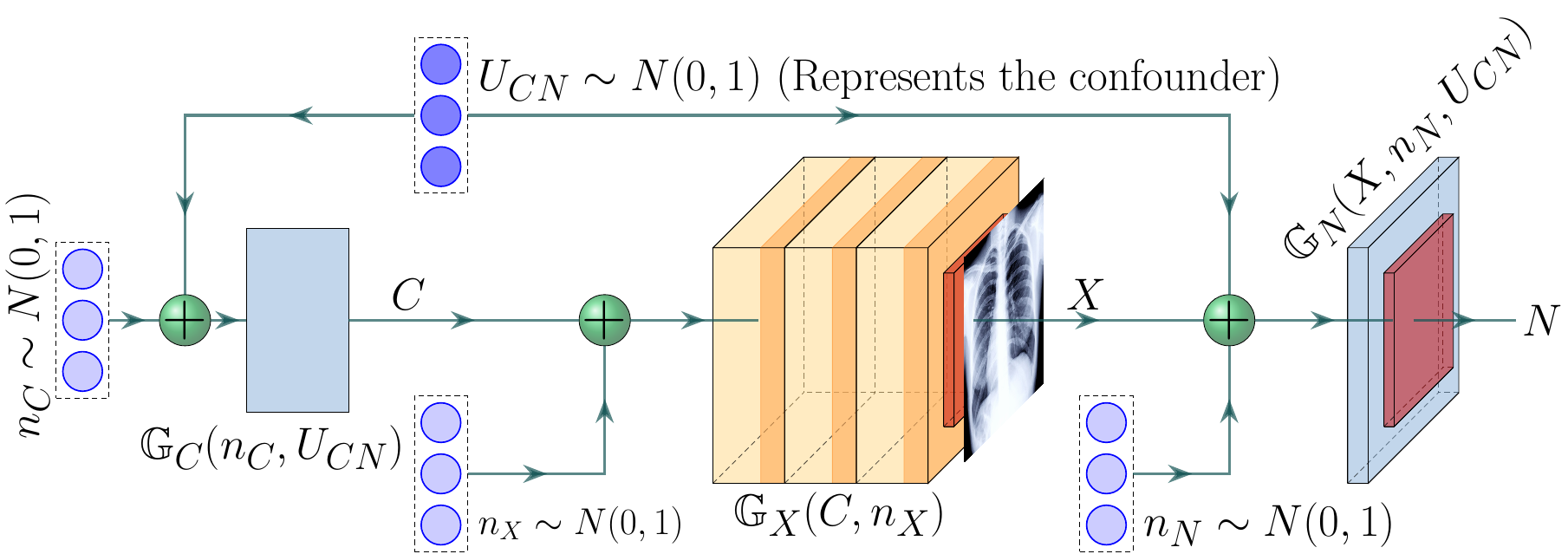}
    \caption{Deep Causal Generative Model
    }
    \label{fig:xray-b}
\end{subfigure}
\caption{Causal graph for the $\mathrm{XrayImg}$ example (top) and its deep causal generative model (bottom). For each variable, an NN ($\mathbb{G}_C,\mathbb{G}_X, \mathbb{G}_N$) is trained to mimic the true mechanism.
}
\label{fig:xray-neural-net}
\end{figure}
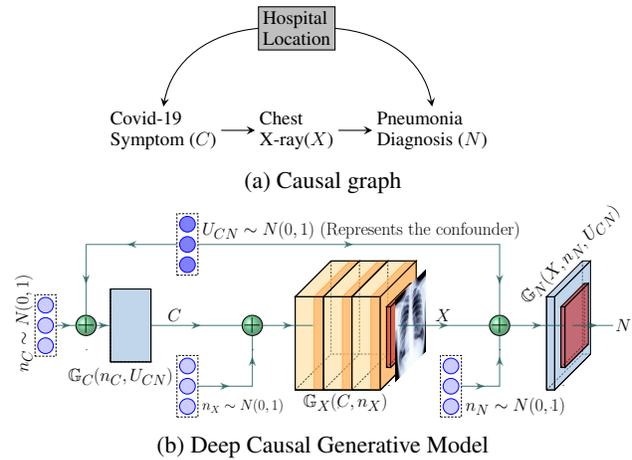

\section{Introduction}
 %
Evaluating the causal effect of an intervention
on a system of interest is one of the fundamental questions that arise across disciplines. 
Pearl's structural causal models (SCMs) provide a principled approach to answering such queries from data. Using SCMs, today we have a clear understanding of which causal queries can be answered from data, 
and which cannot without further assumptions~\cite{pearl1995causal, shpitser2008complete, huang2012pearl,bareinboim2012causal}. These \emph{identification algorithms} find a closed-form expression for an interventional distribution using only observational data, by making use of the causal structure via do-calculus rules~\cite{pearl1995causal}. 
%
Today, most of our datasets contain high-dimensional variables, such as images. 
Modern deep learning architectures can handle high-dimensional data and solve \emph{non-causal} machine learning problems such as classification, detection or generation.   
The existing causal inference algorithms can answer any identifiable causal question, but they 
cannot handle high-dimensional variables as they require access to the joint distribution, which is not practical with image data. 

As an example, consider a healthcare dataset of Covid symptoms ($C$), Pneumonia diagnosis ($N$) and chest X-rays ($X$) of patients from hospitals in two cities with different socio-economic status, which we do not observe to ensure patient privacy. 
Then, hospital location acts as a latent confounder for both $C$ and $N$ since it might have effect on how likely patients are getting sick on average and if they have access to better trained doctors.
In this scenario, the data-generating process can be summarized by the causal graph in Figure \ref{fig:xray-neural-net}. We would like to understand how likely an average person across the two cities is to be diagnosed with pneumonia if they have Covid symptoms, i.e., the causal effect of Covid symptoms on pneumonia diagnosis, $P_{C}(N)$. 
The causal effect can be computed from the observational distribution as $\int_{x,c'} p(x|c)p(n|x,c')p(c')$ through the \emph{front-door adjustment}~\citep{pearl2009causality}. However, it is not possible to reliably estimate $P(x|c)$ or marginalize over all possible $\mathrm{XrayImage}$ due to its high dimensionality.
To the best of our knowledge, no existing algorithm can address this 
problem.

%
%
\par 
Although the use of such structured deep generative models has been explored recently, existing solutions either assume no unobserved confounders~\cite{kocaoglu2018causalgan, pawlowski2020deep}, consider specific graphs~\cite{louizos2017causal,zhang2021treatment} or are effective for low-dimensional variables~\cite{xia2021causal,xia2023neural} due to the challenge of learning high-dimensional joint distribution.
Furthermore, these methods do not have the flexibility to use pre-trained models without affecting their weights. This is useful since state-of-the-art deep models, such as large image generators, can only be successfully trained by a few industrial research labs with expensive resources.

In this paper, we propose a modular, sampling-based solution to address the high-dimensionality challenge the existing algorithms face while learning deep causal generative models. 
Our solution uses deep learning architectures that mimic the causal structure of the system such as in Figure \ref{fig:xray-b}. 
We offer efficient and flexible training facilitated by our key contribution: 
the ability to identify which parts of the deep causal generative models can be trained separately (such as $\{G_X\}$ in Figure \ref{fig:xray-b}), and which parts should be trained together ($\{G_S, G_D\}$). 
We show that after this modularization, there is a \emph{correct} training order for each sub-network (ex:$\{G_X\} \rightarrow \{G_S, G_D\}$). Our algorithm follows such an order to train the sub-networks while freezing weights of already trained ones in the previous steps.
After training, our method can be used to obtain samples from the interventional distribution (ex: $P_{C}(N)$), which is implicitly modeled. 
Thus the modularity in our method enables the flexibility to plug in pre-trained generative networks, and paves the way to utilize large pre-trained models for causal inference. 
%
%
The following are our main contributions.
\begin{itemize}
\item We propose an adversarial learning algorithm 
for training deep causal generative models with latent confounders for high-dimensional variables. 
We show that, after convergence, our model can produce high-dimensional samples according to interventional queries that are identifiable from the data distributions. 
\item To the best of our knowledge, ours is the first algorithm that can modularize the training process in the presence of latent confounders while preserving their theoretical guarantees, thereby enabling the use of large pre-trained models for causal effect estimation.
\item With extensive experiments on Colored-MNIST, we demonstrate that \WG converges better compared to the closest baselines and can correctly generate interventional samples.
We also show our convergence on COVIDx CXR-3 and
solve an invariant prediciton problem on CelebA-HQ.
\end{itemize}
\section{Related Works}
In recent years, a variety of neural causal methods have been developed in the literature to answer interventional queries~\citep{gao2022causal,jerzak2022image,dash2022evaluating, qin2021causal, castro2020causality}.
\citet{shalit2017estimating, louizos2017causal, nemirovsky2020countergan} offer to solve the causal inference problem using deep generative models. 
Yet, they do not offer theoretical guarantees of causal estimation in general, but for some special cases. 
\par
Researchers have recently focused on imposing causal structures within neural network architectures.
Particularly, \citet{kocaoglu2018causalgan} introduced a deep causal model that produces interventional image samples after training on observational data. ~\citet{chao2023interventional} offer similar contribution with diffusion-based~\cite{song2020denoising} approaches. ~\citet{pawlowski2020deep, ribeiro2023high} apply normalizing flows and variational inference to predict exogenous noise for counterfactual inference.
\citet{dash2022evaluating} use an encoder and a generator to produce counterfactual images in order to train a fair classifier.
A major limitation of these works is the causal sufficiency assumption, i.e., each variable is caused by \emph{independent} unobserved variables. 
Semi-Markovian models~\cite{tian2006characterization} which allow unobserved confounders affect pairs is more practical.
\par
For semi-Markovian models, \citet{xia2021causal, balazadeh2022partial, xia2023neural} follow a similar approach as~\citet{kocaoglu2018causalgan} to arrange neural models as a causal graph. They propose a minimization-maximization method to identify and estimate causal effects. 
\citet{xia2023neural} extend these to identify and estimate counterfactual queries.

\par
Most of the existing methods described above can handle only discrete or low-dimensional~\citep{bica2020estimating, yang2021causalvae} variables and it is not clear how to extend their results to continuous high-dimensional image data.
Moreover, if these methods are given a pre-trained neural network model, they do not have the ability to incorporate them in their training. To the best of our knowledge, our approach is the first to address this problem in the presence of unobserved confounders, unlocking the potential of large pre-trained models for causal inference. 

%
%
%
%
%
%
%
\section{Background}
%
\begin{definition}[Structural causal model (SCM)~\citep{pearl2009causality}]
An SCM $\mathcal{M}$ is a $5$-tuple 
$ \mathcal{M}=(\mathcal{V}, \mathcal{N}, \mathcal{U}, \mathcal{F}, P(.) )$, where each observed variable $V_i\in\mathcal{V}$ is realized as an evaluation of the function $f_i\in\mathcal{F}$ which looks at a subset of the remaining observed variables $Pa_i\subset \mathcal{V}$, an unobserved exogenous noise variable $E_i\in \mathcal{N}$, and an unobserved confounding (latent) variable $U_i\in\mathcal{U}$. $P(.)$ is a product joint distribution over all unobserved variables $\mathcal{N}\cup\mathcal{U}$. \end{definition}

Each SCM induces a directed graph called the \emph{causal graph},
or acyclic directed mixed graph (ADMG)
with $\mathcal{V}$ as the vertex set. The directed edges are determined by which variables directly affect which other variable by appearing explicitly in that variable's function. Thus the causal graph is $G=(V,E)$ where $V_i\rightarrow V_j$ iff $V_i\in Pa_j$. The set $Pa_j$ is called the parent set of $V_j$. We assume this directed graph is acyclic (DAG). Under the semi-Markovian assumption, each unobserved confounder can appear in the equation of exactly two observed variables. We represent the existence of an unobserved confounder between $X,Y$ in the SCM by adding a bidirected edge $X\leftrightarrow Y$ to the causal graph. These graphs are no longer DAGs although still acyclic.

$V_i$ is called an ancestor for $V_j$ if there is a directed path from $V_i$ to $V_j$. Then $V_j$ is said to be a descendant of $V_i$. The set of ancestors of $V_i$ in graph $G$ is shown by $An_G(V_i)$. A do-intervention $do(v_i)$ replaces the functional equation of $V_i$ with $V_i=v_i$ without affecting other equations. The distribution induced on the observed variables after such an intervention is called an interventional distribution, shown by $P_{v_i}(\mathcal{V})$. $P_{\emptyset}(\mathcal{V})=P(\mathcal{V})$ is called the observational distribution. 
In this paper, we use $\mathcal{L}_1$ and $\mathcal{L}_2$ as notation for observational and interventional distributions, respectively. 
\begin{definition}[c-components]
\label{def:cc}
A subset of nodes is called a c-component if it is a maximal set of nodes in $G$ that are connected by bi-directed paths.
\end{definition}
\section{Deep Causal Generative Model with Latents}
%
%

\begin{figure*}[t!]
    \centering

    \begin{subfigure}{0.6\textwidth}
            \begin{minipage}[c]{1\linewidth}
        \begin{tikzpicture}[scale=0.8, transform shape]


 \node  (z3) {$Z_3$};
 \node [right =0.4cm of z3] (z1) {$Z_1$};
 \node [right =0.4cm of z1] (z2) {$Z_2$};

 \node [below =0.4cm of z3] (x1) {$X_1$};
 \node [right =1.4cm of x1] (x2) {$X_2$};

 \node [below =1.2cm of z1] (text) {$\text{Causal Graph}$};

 \draw[ thick] (z3) to  (z1); 
 \draw[ thick] (z1) to  (z2); 
 \draw[ thick] (x1) to  (z1); 
 \draw[ thick] (z1) to  (x2); 
 \path[bidirected] (z3) edge[bend left=45] (z1);
 \path[bidirected] (z1) edge[bend left=45] (z2);
 \path[bidirected] (x1) edge[bend right=25] (x2);


   \node [rectangle, draw, right =0.4 cm of z2] (fc1) {$C_{Z_3Z_1Z_2}$};
    \node [rectangle, draw, below =0.4 cm of fc1] (fc2) {$C_{X_1,X_2}$};
  \draw[ thick] (fc1) to  (fc2);

\node [bag, above =1.2cm of fc2]  {\text{Step (1/ 3):}\\
\text{Build H-graph:}};

  \node [circle, fill=\nc,  right =1.2cm of fc1] (z3) {$\wh{Z_3}$};
 \node [circle, fill=\nc,right =0.4cm of z3] (z1) {$\wh{Z_1}$};
 \node [circle, fill=\nc,right =0.4cm of z1] (z2) {$\wh{Z_2}$};

 \node [below =0.4cm of z3] (x1) {$X_1$};
 
 \draw[ thick] (z3) to  (z1); 
 \draw[ thick] (z1) to  (z2); 
 \draw[ thick] (x1) to  (z1); 
 \path[bidirected] (z3) edge[bend left=45] (z1);
 \path[bidirected] (z1) edge[bend left=45] (z2);
\node [bag, above =0.1cm of z1]  {Step (2/ 3): Train $Z_1,Z_2,Z_3$ \\
Or, use pre-trained $Z_1,Z_2,Z_3$};

\node [bag, below =1.2cm of z1]  {Match $P(z_1,z_2,z_3|x_1)$ \\
$= Q(z_1,z_2,z_3|do(x_1))$};

  \node [right =1.2cm of z2] (z3) {$Z_3$};
 \node [right =0.4cm of z3] (z1) {$Z_1$};

 \node [circle, fill=\nc, below =0.4cm of z3] (x1) {$\wh{X_1}$};
 \node [circle, fill=\nc, right =1.2cm of x1] (x2) {$\wh{X_2}$};
 
 \draw[ thick] (z3) to  (z1); 
 \draw[ thick] (x1) to  (z1); 
 \draw[ thick] (z1) to  (x2); 
 \path[bidirected] (z3) edge[bend left=45] (z1);
 \path[bidirected] (x1) edge[bend right=25] (x2);

\node [bag, above =0.1 cm of z1]  {Step (3/3):\\ Train $X_1, X_2$};
 \node [bag, below =1.2cm of z1]  {Match $P(x_1,x_2,z_1, z_3)$  \\
$= Q(x_1,x_2,z_1, z_3)$};

\end{tikzpicture}
        \end{minipage}
        \caption{
         Step 1: Since $P(x_1,x_2| do(z_1)) \neq P(x_1,x_2| z_1)$, add edge from c-component $H_1: [Z_1,Z_2,Z_3]$ to c-component $H_2: [X_1,X_2]$. Step 2: train only networks in $H_1$ to match $P(z_1,z_2,z_3|x_1) = Q(z_1,z_2,z_3|do(x_1))$. Step 3: train only $H_2$ while using pre-trained  $Z_1,Z_3$ to match $P(x_1,x_2,z_1, z_3) = Q(x_1,x_2,z_1, z_3)$.
       }
       \label{fig:modular-simulation}
    \end{subfigure}
\hspace{5mm}
\begin{subfigure}  {0.35\textwidth}  
\begin{minipage}[c]{1\linewidth}
       \begin{tikzpicture}[scale=0.8, transform shape]


 \node  [right =1.5cm of z1] (1) {$Z$};
 \node [below left =0.5cm and -0.2cm of 1] (2) {$W$};
 \node [below left =0.5cm and -0.2cm of 2] (3) {$X$};
 \node [right =1cm of 3] (4) {$Y$};
 
 \draw[ thick] (1) to  (2); 
 \draw[ thick] (2) to  (3); 
 \draw[ thick] (3) to  (4); 
 \draw[ thick] (1) to  (4); 
 \path[bidirected] (1) edge[bend right=45] (3);

 \node [rectangle, draw, right =of 1] (h1) {$C_W$};
  \node [rectangle, draw, below =0.5 cm of h1] (h2) {$C_{XZ}$};
 \node [rectangle, draw, below =0.2 cm of h2] (h3) {$C_Y$};
  \draw[ thick] (h1) to  (h2); s

\node  [right =0.2cm of h1] (x0) {$X_0$};
 \node [right =0.5cm of x0] (w0) {$W_0$};
 \node [right =0.5cm of w0] (y0) {$Y_0$};

 \node [below =0.5cm of x0] (x1_) {$X_1$};
 \node [below =0.5cm of w0] (x2_) {$X_2$};
 \node [below =0.5cm of y0] (w1) {$W_1$};

 \node [below =0.5cm of w1] (y1) {$Y_1$};

 \draw[ thick] (x0) to  (w0); 
 \draw[ thick] (w0) to  (y0); 
 \draw[ thick] (x1_) to  (x2_); 
 \draw[ thick] (x2_) to  (w1); 
\draw[ thick] (x0) to  (w1); 
\draw[ thick] (w1) to  (y0); 
\draw[ thick] (w1) to  (y1); 

\path[bidirected] (x0) edge[bend left=35] (y0);
\path[bidirected] (x1_) edge[bend right=45] (x2_);
\path[bidirected] (x2_) edge[bend right=25] (y1);

\node [rectangle, draw, right =0.15 cm of y0] (c1_) {$C_{W_0}$};
\node [rectangle, draw, below =0.65 cm of c1_] (c2_) {$C_{X_0,Y_0}$};
\node [rectangle, draw, below =1 cm of c2_ ] (c3) {$C_{W_1}$};
\node [rectangle, draw, left =0.5 cm of c3 ] (c4) {$C_{X_1,X_2,Y_1}$};
\draw[ thick] (c1_) to  (c2_); 
\draw[ thick] (c3) to  (c2_); 
\draw[ thick] (c3) to  (c4);

 \node  [below = 0.5cm of 4](fx) {$X$};
 \node  [right = 0.5cm of fx](fz) {$Z$};
 \node  [right = 0.5cm of fz](fy) {$Y$};

 \draw[ thick] (fx) to  (fz); 
 \draw[ thick] (fz) to  (fy); 
 \path[bidirected] (fx) edge[bend left=35] (fy);

   \node [rectangle, draw, below right = 0.1cm and -0.5cm of fx] (fc1) {$C_{Z}$};
    \node [rectangle, draw, right =0.5 cm of fc1] (fc2) {$C_{XY}$};
  \draw[ thick] (fc1) to  (fc2); 

\end{tikzpicture} 
\end{minipage}
\caption{
 The rectangular box represents each h-node and which networks we have to train together. The partial order of the h-graph represents in which order we should train the mechanisms
}
\label{fig:h-graph_examples}
    \end{subfigure}
\caption{
(Left:)
 Modular training in 3 steps.
 (Right:) Causal graphs and their h-graphs showing modularization of the training process.
}
\label{fig:modular-training-h-graphs}
\end{figure*}
%
%
%

Suppose the ground truth data-generating SCM is made up of functions $X_i=f_i(Pa_i,E_i)$. If we have these equations, we can simulate an intervention on, say $X_5=1$, by evaluating the remaining equations. However, we can never hope to learn the true functions and unobserved noise terms from data. 
The fundamental observation of Pearl is that even then there are some causal queries that can be uniquely identified as some \emph{deterministic function} of the causal graph and the joint distribution between observed variables, e.g., $p(n|do(c))=\xi(G,p(c,x,n))$ in Figure~\ref{fig:xray-a} for some deterministic $\xi$.
This means that, if we can, somehow, train a causal model made up of neural networks that fits the $data\sim p(\mathbf{v})$, and has the same causal graph, then it has to induce the same interventional distribution $p(d|do(s))$ as the ground truth SCM, \emph{irrespective of what functions the neural network uses}. This is a very strong idea that allows mimicing the causal structure, and opens up the possibility of using deep learning algorithms for performing causal inference through sampling even with high-dimensional variables.
This is the basic idea behind \cite{kocaoglu2018causalgan} without latents. Motivated by their work, we define a \emph{deep causal generative model} for semi-Markovian model and show identifiability
\footnote{Identifiability here refers to our ability to uniquely sample from an interventional distribution. See Definition~\ref{appex:def-id} for details.}
results. 
Now, we formalize the above simple observations.
\begin{definition}[DCM]
\label{def:scm}
	A neural net architecture $\mathbb{G}$ is called a deep causal generative model (DCM) for an ADMG $G=(\mathcal{V},\mathcal{E})$ if it is composed of a collection of neural nets, one  $\mathbb{G}_i$ for each $V_i\in\mathcal{V}$ such that 
		i) \emph{each $\mathbb{G}_i$ accepts a sufficiently high-dimensional noise vector $N_i$,} 
		ii) \emph{the output of $\mathbb{G}_j$ is input to $\mathbb{G}_i$ iff $V_j\in Pa_G(V_i)$,}
		iii) \emph{$N_i=N_j$ iff $V_i\leftrightarrow V_j$. }
\end{definition}
We define $Q$ as the distribution induced by the DCM. Noise vectors $N_i$ replace both the exogenous noises and the unobserved confounders in the true SCM. They are of sufficiently high dimension to induce the observed distribution. We say that a DCM is \emph{representative enough for an SCM} if the neural networks have sufficiently many parameters to induce the observed distribution induced by the SCM. 
For the neural architectures of variables in the same c-component, we can consider conditional GANs~\citep{mirza2014conditional}, as they are effective in matching the joint distribution by feeding the same prior noise $N_i=N_j$ (as confounders) into multiple generators. For variables that are not confounded ($N_i\neq N_j$), we can use conditional models such as diffusion models~\cite{ho2022classifier}.
With Defintion\ref{def:scm}, we have the following, similar to \cite{xia2021causal}:
\begin{theorem}
	\label{th:identifiability}
	Consider any SCM $\mathcal{M}=(G, \mathcal{N}, \mathcal{U}, \mathcal{F}, P(.) )$.  A DCM $\mathbb{G}$ for $G$ entails the same identifiable interventional distributions as the SCM $\mathcal{M}$ if it entails the same observational distribution.  
	\end{theorem}

Thus, even with high-dimensional variables in the true SCM, given a causal graph, in principle, any identifiable interventional query can be sampled from, with a DCM that fits the observational distribution.
However, to learn the DCM, \cite{kocaoglu2018causalgan,xia2021causal} suggest training all neural nets $\mathbb{G}$ in the DCM together. Such an approach to match the joint distribution containing all low and high-dimensional variables is empirically challenging in terms of convergence. Any modularization not only is expected to help train more efficiently for better solution quality, but also allow the flexibility to use pre-trained image generative models. 
Now, we focus on uncovering how to achieve such modularization and how it contributes along the two aspects. 

\subsection{Modular-DCM Intuitive Explanation}
%
Consider the graph $G$ in Figure \ref{fig:modular-simulation}. 
Suppose, we have an observational dataset $D\sim P(\mathcal{V})$.
Based on Theorem \ref{th:identifiability},
we can to sample from different \Lii distributions such as $P(x_2|do(x_1))$ and $P(z_2|do(x_1))$ by training a DCM $\mathbb{G}$ that is consistent with $G$ and fits the observational data $P(\mathcal{V})$. 
The DCM will contain one feed-forward neural net per observed variable, i.e., $\mathbb{G}=\{\mathbb{G}_{Z_1},\mathbb{G}_{Z_2},\mathbb{G}_{Z_3},\mathbb{G}_{X_1},\mathbb{G}_{X_2}\}$.
If we follow the naive way and jointly train all networks in $\mathbb{G}$ together, we have to match $P(x_1,x_2,z_1,z_2,z_3)$ containing 
all low and high dimensional variables in a single training phase. 
Matching this joint distribution by training all models at the same time could be difficult, since we are attempting to minimize a very complicated loss function. 
Thus, the question we are interested in is, which neural nets can be trained separately, and which need to be trained together \emph{to be able to fit the joint distribution}.

\par Suppose we first train the causal generative model $\mathbb{G}_{Z_1}$, i.e., learn a mapping that can sample from $P(z_1|z_3,x_1)$.
Even if we provide the unobserved confounder $N_1$ (Definition \ref{def:scm}), which also affects $Z_2$ and $Z_3$, the neural network might learn a mapping that later makes it impossible to induce the correct dependence between $Z_1,Z_2$, or $Z_1,Z_3$ no matter how $\mathbb{G}_{Z_2}$ or $\mathbb{G}_{Z_3}$ are trained later. This is because fitting the conditional $P(Z_1|Z_3,X_1)$ does not provide any incentive for the model $\mathbb{G}_{Z_1}$ to induce the correct confounding dependency (through the latent variables) with $Z_2$ and $Z_3$. 
If the model ignores the confounding dependence, it cannot induce the dependence between $Z_3$ and $Z_2$ conditioned on $Z_1$ ($Z_3 \not\indep Z_2|Z_1$). 
This observation suggests that 
the causal mechanisms of variables that are in the same c-component should be trained together. 
Therefore, we have to train $[\mathbb{G}_{Z_1},\mathbb{G}_{Z_2}, \mathbb{G}_{Z_3}]$
together; similarly $[\mathbb{G}_{X_1}, \mathbb{G}_{X_2}]$ together. 

To match the joint $P(\mathcal{V})$ for semi-Markovian models 
while preserving the integrity of c-components, we propose using Tian's factorization~\citep{tian2002general}.
It factorizes $P(\mathcal{V})$ into c-factors: the joint distributions of each c-component $C_j$ intervened on their parents, i.e., $P_{pa(c_j)}(c_j)$.
%
%
\begin{equation}
	\label{eq:fake-intv-vs-true-intv}
	\begin{split}
		P(v)&= P(x_1,x_2|\doo(z_1))  P(z_1,z_2,z_3|\doo(x_1))\\
	\end{split}
\end{equation}
Due to this factorization, fitting $P(\mathcal{V})$ is equivalent to fitting each of the c-factors. 
If we had access to the 
\Lii distributions from $\doo(z_1)$ and $\doo(x_1),$
$\forall z_1x_1$,
we could intervene on $\mathbb{G}_{Z_1}$ and $\mathbb{G}_{X_1}$ in the DCM to obtain $\doo(z_1)$ and $\doo(x_1)$ samples and
train the models to match these \Lii distributions. 
However, we only have access to the $P(\mathcal{V})$ dataset.

\par Note that it is very difficult to \emph{condition} in feedforward models during training, which is the case in a DCM. To sample from 
$Q(x_2|z_1)$ 
it is not sufficient to feed 
$z_1$ 
to the network 
$\mathbb{G}_{X_2}$.
In fact, observe that this is exactly the intervention operation, and would give us a sample from $Q(x_2|do(z_1))$.
%
It is trivial to intervene on the inputs to a neural network, but highly non-trivial to condition since feedforward models cannot easily be used to correctly update the posterior via backdoor paths. 
Thus, we need to find some interventional distribution such that the DCM can generate samples for this distribution and
can be trained by comparing them with some equivalent true observational samples. 

%

Our key idea is to \textbf{leverage the do-calculus rule-$2$}~\citep{pearl1995causal} to use observational samples and pretend that they are from these \Lii distributions. This gives us a handle on how to modularize the training process of 
c-components. For example, in Figure~\ref{fig:modular-simulation}, c-factor $P(z_1,z_2,z_3|\doo(x_1))=P(z_1,z_2,z_3|x_1)$ since do-calculus rule-$2$ applies, i.e., intervening on $X_1$ is equivalent to conditioning on $X_1$.
We can then use the conditional distribution as a proxy/alternative to the c-factor to learn $Q(z_1,z_2,z_3|\doo(x_1))$ with the DCM. 
However, $P(x_1,x_2|\doo(z_1))\neq P(x_1,x_2|z_1)$. 
To overcome this issue, we seek to fit a joint distribution that implies this c-factor, i.e., we find a superset of $X_1,X_2$ on which rule-$2$ applies. We can include $Z_1$ into the joint distribution that needs to be matched together with $X_1,X_2$ and check if the parent set of $\{X_1,X_2,Z_1\}$ satisfy rule-$2$. We continue including variables until we reach the joint $P(x_1,x_2,z_1,z_3)$ to be the alternative distribution for $\{X_1,X_2\}$'s c-factor.

\par After identifying which sub-networks of the DCM can be trained separately, we need to decide \emph{a valid order} in which they should be trained. For the same example, we can first train $[\mathbb{G}_{Z_1}, \mathbb{G}_{Z_2}, \mathbb{G}_{Z_3}]$ together to induce $Q(z_1,z_2,z_3|do(x_1))= P(z_1,z_2,z_3|x_1)=P(z_1,z_2,z_3|do(x_1))$. This is shown in step (2/3) in Figure~\ref{fig:modular-simulation}: We can produce samples from the mechanisms of $Z_1, Z_2, Z_3$ by intervening on their parent $X_1$ with real observations from dataset {$D$}. Thus, we do not need  $\mathbb{G}_{X_1}$  to be pre-trained.
Now, we train mechanisms of the next c-component $[\mathbb{G}_{X_1}, \mathbb{G}_{X_2}]$  in our training order (step 3/3). As discussed, we need to ensure $Q(x_1,x_2,z_1,z_3|do(
\emptyset))=P(x_1,x_2,z_1,z_3|do(
\emptyset))=P(x_1,x_2,z_1,z_3|\emptyset)$. Since mechanisms of $Z_1,Z_3$ were trained in the previous step, we can freeze the network weights of $[\mathbb{G}_{Z_1}, \mathbb{G}_{Z_3}]$. These are used to correctly sample from $Z_1$ given $X_1$, and feed this correctly sampled value into the network of $X_2$. In Appendix \ref{append:main_paper_example}, we show that the c-factors in Equation~\ref{eq:fake-intv-vs-true-intv} will correctly match the true c-factors after fitting these two conditional probabilities in this order. 
Therefore, DCM matches the joint distribution $P(\mathcal{V})$ as well. On the other hand, if we first trained the networks $[\mathbb{G}_{X_1}, \mathbb{G}_{X_2}]$, it would not be possible to match the joint $P(x_1,x_2,z_1,z_3)$ as the mechanisms of $Z_1,Z_3$ are not yet trained. Thus, this order would not work. 
\subsection{Training Algorithm for Modular-DCM}
In this section, we generalize the discussed ideas, into a modular algorithm that has mainly two phases: 1) arranging the c-components in a valid training order and 2) training (sets of) c-components to match their c-factors.

\emph{\textbf{Arranging the c-components}}: Consider a c-component $C_t$. When should a c-component $C_s$ be trained before $C_t$? Since 
we need rule $2$ of do-calculus to hold on the parents of $C_t$ for training, if 
$C_s$ contains some parents of $C_t$ that are located on the backdoor paths between any two variables in $C_t$, then $C_s$ must be pre-trained before $C_t$. 
Conditioning and intervening on those parents of $C_t$ is not the same, i.e., $P(C_t|do(pa(C_t) \cap C_s)\!\neq\! P(C_t|pa(C_t) \cap C_s)$. 
Thus we include $pa(C_t) \cap C_s$ in the joint distribution that we want to match for $C_t$, which requires those parents in $C_s$ to be pre-trained.
For the front-door graph in Figure~\ref{fig:h-graph_examples}, we observe that $P(X,Y|\doo (Z))\neq P(X,Y|Z)$. Thus, we train $\mathbb{G}_X$, $\mathbb{G}_Z$, $\mathbb{G}_Y$ in $[C_Z:\{\G_Z\}] \rightarrow [C_{XY}:\{\G_X, \G_Y\}]$ order.
To obtain a partial order among all c-components, we construct a directed graph structure called $\mathcal{H}$-graph that contains c-components as nodes. 
While adding edges, if any cycle is formed, 
we merge c-components on that cycle into a single h-node indicating that they will need to be trained jointly. Thus some h-nodes may contain more than one c-component. 
The final structure is a DAG and gives us a valid partial order  $\mathcal{T}$ for modular training (Proposition~\ref{appex-lemma:consistent-hgraph}). 
Formally, an $\mathcal{H}$-graph is defined as:
\begin{definition}[$\mathcal{H}$-graph]
\label{def:H-graph}
Given a causal graph $G$ with c-components  $\mathcal{C}=\{C_1,\hdots C_n\}$, let $\{H_k\}_k$ be some partition  of $\mathcal{C}$. The directed graph $(V_{\mathcal{H}},E_{\mathcal{H}})$ where $V_{\mathcal{H}}= \{H_k\}_k$ and 
 $H_s \rightarrow H_t \in E_{\mathcal{H}}$ iff $P(H_t|\doo(pa_{G}(H_t) \cap H_s))\neq$ $P(H_t|pa_{G}(H_t) \cap H_s)$, is called an $\mathcal{H}$-graph for $G$ if  acyclic. 
 \end{definition}
 %
%
We run Algorithm~\ref{construct-H-graph}:$\mathrm{Contruct\_Hgraph()}$ to build an $\mathcal{H}$-graph by checking the edge condition on line 5.
In Figure~\ref{fig:modular-training-h-graphs} and Appendix~\ref{appex:sec-long-hgraph}, we provide some examples of $\mathcal{H}$-graphs.
Note that we only use the $\mathcal{H}$-graph to obtain a partial training order of h-nodes.
For any h-node $H_k$,
$An(H_k)$ and $Pa(H_k)$ below 
refer to ancestors and parents in the causal graph $G$, not in the $\mathcal{H}$-graph.
\emph{\textbf{Training c-components}}:
We follow $\mathcal{H}$-graph's topological order and train the c-components in an h-node $H_k$.
If we can match $P_{pa(H_k)}(H_k)$, it will ensure that the DCM will learn their corresponding c-factors $P_{pa(C_j)}(C_j), \forall C_j\in H_k$ as well. 
As mentioned earlier, we can generate fake interventional samples from $Q_{pa(H_k)}(H_k)$ induced by the DCM, but they cannot be used to train $\mathbb{G}_{H_k}$ as we do not have access to real data samples from the interventional distribution $P_{pa(H_k)}(H_k)$.
Thus, we train $\mathbb{G}_{H_k}$ to learn a larger joint distribution that can be obtained from the observational dataset as an alternative to its c-factors.
We search for a set $\mathcal{A}_k$ that can be added to the joint with $H_k$ such that
$P_{Pa(H_k , \mathcal{A}_k)}(H_k, \mathcal{A}_k) 
	= P(H_k, \mathcal{A}_k| Pa(H_k , \mathcal{A}_k))$, i.e., 
true interventional and conditional distribution are the same. This enables us to take conditional samples from the input dataset and use them as true interventional samples to match
them with the DCM-generated fake interventional samples from $Q_{Pa(H_k , \mathcal{A}_k)}(H_k, \mathcal{A}_k)$ and train $\mathbb{G}_{H_k}$.
The above condition is generalized as a  \textbf{modularity condition}:
\begin{definition}
	\label{def:modularity-condition}
 Let $H_k$ be an h-node in the $\mathcal{H}$-graph. A set $\mathcal{A}_k \subseteq An_{G}(H_k)\setminus H_k$ satisfies the modularity condition if it is the smallest set with 
	$P(H_k, \mathcal{A}_k|do(Pa(H_k , \mathcal{A}_k)))
	= P(H_k, \mathcal{A}_k| Pa(H_k , \mathcal{A}_k))$.
\end{definition}
%
%
\setlength{\textfloatsep}{5pt}
\begin{algorithm}[t!]
\footnotesize
\caption{\footnotesize Modular Training($G, \mathbf{D}$)}
\begin{algorithmic}[1]
   \STATE {\bfseries Input:} Causal Graph $G$, Dataset $\mathbf{D}$.
\STATE Initialize DCM $\mathbb{G}$
\STATE $\mathcal{H}\leftarrow$ $\mathrm{Construct\_Hgraph}$($G$) \label{line:const-hgraph}
\FOR{{\bfseries each} $H_k \in \mathcal{H} $ in partial order \label{line:iter-hnodes}
}
\STATE Initialize $\mathcal{A}_k \leftarrow \emptyset$  
\WHILE{$\mathrm{IsRule2}(H_k, \mathcal{A}_k)=0$}   
{
\STATE $\mathcal{A}_k \leftarrow  Pa_{G}(H_k,\mathcal{A}_k)   $ \label{line:assign-A}
}
\ENDWHILE
\STATE $\mathbb{G}_{H_k}\leftarrow$$\mathrm{TrainModule}$$({\mathbb{G}_{H_k}, G}, H_k, \mathcal{A}_k, \mathbf{D} $)\label{line:call-trmod}
\ENDFOR
   \STATE {\bfseries Return:} $\mathbb{G}$
\end{algorithmic}
\label{alg:train-by-components}
\end{algorithm}

As mentioned earlier, such \Li and \Lii distributional equivalence holds when the do-calculus rule-$2$ applies:
\begin{equation}
\begin{split}
     P_{Pa(H_k , \mathcal{A}_k)}(H_k, \mathcal{A}_k)
	= P(H_k, \mathcal{A}_k| Pa(H_k , \mathcal{A}_k)), \hspace{1em} \\
  \text{\hspace{0mm} if  $(H_k, \mathcal{A}_k \indep Pa(H_k , \mathcal{A}_k))_{G_{\underline{Pa(H_k , \mathcal{A}_k)}}}$}
\end{split}
\end{equation}
This suggests a graphical criterion to find such a set $\mathcal{A}_k$ and we apply it at line 6 in Algorithm~\ref{alg:train-by-components}. 
Intuitively, if the outgoing edges of $Pa(H_k , \mathcal{A}_k)$ are deleted ($G_{\underline{Pa(H_k , \mathcal{A}_k)}}$) and they become d-separated from $\{H_k , \mathcal{A}_k\}$, then there exists no backdoor path from $Pa(H_k , \mathcal{A}_k)$ to $\{H_k, \mathcal{A}_k\}$ in $G$.
%
%
Therefore, for a specific $H_k$, 
we start with $\mathcal{A}_k=\emptyset$ and check if $Pa(H_k , \mathcal{A}_k)$ satisfies the conditions of the rule-$2$ for $\{H_k , \mathcal{A}_k\}$. If not, we add parents of $\{H_k , \mathcal{A}_k\}$ to $\mathcal{A}_k$.
We include ancestors, since only they can affect  $H_k$'s mechanisms from outside of the c-component.
We continue the process until $Pa(H_k , \mathcal{A}_k)$ satisfies rule-$2$.
%
Finally, finding a set $\mathcal{A}_k$ satisfying the modularity condition implies that we can train $\mathbb{G}_{H_k}$ by matching:
\begin{equation}
\label{eq:hnode-mod-cond}
\begin{split}
    Q_{pa(H_k , \mathcal{A}_k)}(H_k ,  \mathcal{A}_k)
    = P(H_k , \mathcal{A}_k |  {Pa(H_k , \mathcal{A}_k)}) \\
    \text{; Now training: $H_k$, Pre-trained: $\mathcal{A}_k$}
\end{split}
\end{equation}
We utilize adversarial training to train the generators in $\mathbb{G}_{H_k}$ on observational dataset $\mathbf{D}$ to match the above. This is done by Algorithm~\ref{alg:modular-train-DCM}: $\mathrm{TrainModule()}$ called in line 8.
More precisely, this sub-routine uses all mechanisms in $\{H_k , \mathcal{A}_k\}$ to produce samples but only updates the mechanisms in $\mathbb{G}_{H_k}$ corresponding to the current h-node and returns those models after convergence.
Even though we will train only $\mathbb{G}_{H_k}$ i.e., $\mathbb{G}_V, {\forall V \in H_k}$, $\mathcal{A}_k$ appears together with $H_k$ in the joint distribution that we need to match. Thus, we use pre-trained causal mechanisms of $\mathcal{A}_k$, i.e., $\mathbb{G}_V, {\forall V \in \mathcal{A}_k}$ here.
 The partial order of $\mathcal{H}$-graph ensures that we have already trained $\mathcal{A}_k$ before $H_k$.

Training $\mathbb{G}_{H_k}$ to match the distribution
in Equation~\ref{eq:hnode-mod-cond}, is sufficient
to learn the c-factors $P_{Pa(C_j)}(C_j), \forall C_j \in H_k$.
After training each $\mathbb{G}_{H_k}$ according to the partial order of $\mathcal{H}$-graph, \WG will learn a DCM that induces  
$Q(\mathcal{V})\!=\! P(\mathcal{V})$.
%
Finally, the trained DCM can sample from 
interventional $\mathcal{L}_2$ distributions  identifiable from  $P(\mathcal{V})$.
These are formalized in Theorem~\ref{th:main:modular-train-converges}.
Proofs are in Appendix \ref{appex:converge-alg-3-proof}.
%
%
%
%
%
%
%
%
%
%
%
%

\textbf{Assumptions:} 1. The true ADMG is known. 
2. The causal model is semi-Markovian. 
3. The data distribution is strictly positive.
4. Each conditional generative model $\mathbb{G}_i, \forall i$ in the DCM can correctly learn the target conditional distribution.
%
%
%
\begin{theorem}
\label{th:main:modular-train-converges}
Consider any SCM $\mathcal{M}=(G, \mathcal{N}, \mathcal{U}, \mathcal{F}, P(.) )$.  
Suppose Assumptions 1-4 hold. Algorithm \ref{alg:train-by-components} on $(G,\mathbf{D})$ returns a 
DCM $\mathbb{G}$ 
that entails $i)$ the same observational distribution, and $ii)$ the same identifiable interventional distributions as the SCM $\mathcal{M}$.
\end{theorem}
 
\begin{figure*}[t!]
  \begin{minipage}[c]{0.30\textwidth}
\begin{subfigure}{1\linewidth}
\includegraphics[width=0.9\textwidth]{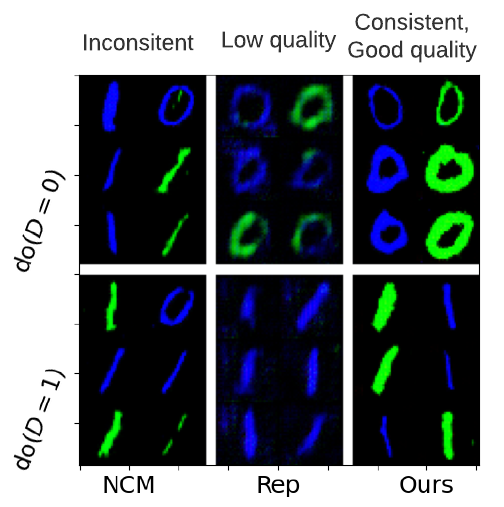}
\caption{Generated Images}
\label{fig:digit-benchmarks}
\end{subfigure}
  \end{minipage}
\begin{minipage}[c]{0.65\textwidth}
\begin{subfigure}{1\linewidth}
  \caption{TVD for frontdoor graph}
 \label{fig:tvd-frontdoor}
 \begin{minipage}[c]{0.75\textwidth}
 \centering
 \scriptsize
\begin{tabular}{|c|ccc|ccc|ccc|}
\hline
\begin{tabular}[c]{@{}c@{}}Total Variation  Distance (TVD) $\downarrow$ \end{tabular}     & \multicolumn{3}{c|}{$P(D,A)$}                                & \multicolumn{3}{c|}{$P(A|\doo(D=0))$}                              & \multicolumn{3}{c|}{$P(A|\doo(D=1))$}                              \\ \hline
Epochs  & \multicolumn{1}{c|}{25}   & \multicolumn{1}{c|}{150}  & \textbf{300}  & \multicolumn{1}{c|}{25}   & \multicolumn{1}{c|}{150}  & \textbf{300}  & \multicolumn{1}{c|}{25}   & \multicolumn{1}{c|}{150}  & \textbf{300}  \\ \hline
NCM     & \multicolumn{1}{c|}{0.14} & \multicolumn{1}{c|}{0.16} & 0.17 & \multicolumn{1}{c|}{0.48} & \multicolumn{1}{c|}{0.42} & 0.43 & \multicolumn{1}{c|}{0.08} & \multicolumn{1}{c|}{0.11} & 0.11 \\ \hline
DCM-Rep & \multicolumn{1}{c|}{0.38} & \multicolumn{1}{c|}{0.17} & 0.16 & \multicolumn{1}{c|}{0.36} & \multicolumn{1}{c|}{0.45} & 0.36 & \multicolumn{1}{c|}{0.22} & \multicolumn{1}{c|}{0.1}  & 0.11 \\ \hline
DCM (Ours)     & \multicolumn{1}{c|}{0.27} & \multicolumn{1}{c|}{0.17} & \textbf{0.08} & \multicolumn{1}{c|}{0.43} & \multicolumn{1}{c|}{0.36} & \textbf{0.13} & \multicolumn{1}{c|}{0.14} & \multicolumn{1}{c|}{0.1}  & \textbf{0.04} \\ \hline
\end{tabular}
 \end{minipage}
 \end{subfigure}
 \begin{subfigure}{0.45\linewidth}
   \caption{FID for frontdoor graph}
 \label{fig:fid-frontdoor}
    \scriptsize
\begin{tabular}{|c|ccc|}
\hline
                                                          & \multicolumn{3}{c|}{\begin{tabular}[c]{@{}c@{}}Frechet Inception Distance (FID) $\downarrow$ \end{tabular}} \\ \hline
Epochs                                                    & \multicolumn{1}{c|}{25}                & \multicolumn{1}{c|}{150}               & \textbf{300}              \\ \hline
NCM                                                       & \multicolumn{1}{c|}{61.40}             & \multicolumn{1}{c|}{59.28}             & 59.59            \\ \hline
\begin{tabular}[c]{@{}c@{}}NCM\\ Pre-trained\end{tabular} & \multicolumn{1}{c|}{26.60}             & \multicolumn{1}{c|}{184.40}            & 192.27           \\ \hline
DCM-Rep                                                 & \multicolumn{1}{c|}{151.41}            & \multicolumn{1}{c|}{78.69}             & 80.65            \\ \hline
DCM (Ours)                                                & \multicolumn{1}{c|}{\textbf{27.02}}    & \multicolumn{1}{c|}{\textbf{27.37}}    & \textbf{27.20}   \\ \hline
\end{tabular}
 \end{subfigure}
 \hspace{6mm}
 \begin{subfigure}{0.55\linewidth}
    \caption{FID for Diamond graph}
 \label{fig:fid-diamond}
    \scriptsize
\begin{tabular}{|c|ccc|}
\hline
           & \multicolumn{3}{c|}{Frechet Inception Distance FID($I_1$) $\downarrow$}                                  \\ \hline
Epochs     & \multicolumn{1}{c|}{25}     & \multicolumn{1}{c|}{150}   & \textbf{300}   \\ \hline
NCM        & \multicolumn{1}{c|}{101.00} & \multicolumn{1}{c|}{67.82} & 80.54 \\ \hline
DCM (Ours) & \multicolumn{1}{c|}{79.96}  & \multicolumn{1}{c|}{35.74} & \textbf{32.88} \\ \hline
           & \multicolumn{3}{c|}{Frechet Inception Distance FID($I_2$) $\downarrow$}                                  \\ \hline
NCM        & \multicolumn{1}{c|}{112.61} & \multicolumn{1}{c|}{67.56} & 65.17 \\ \hline
DCM (Ours) & \multicolumn{1}{c|}{22.22}  & \multicolumn{1}{c|}{15.47} & \textbf{11.80} \\ \hline
\end{tabular}
  \end{subfigure}
\end{minipage} 
      \caption{
      For the frontdoor graph in Figure~\ref{fig:frontdoor-graph}, NCM produces good images but not consistent with $\doo(D)$. \WG without modular training (DCM-rep) produces consistent but low-quality images. Our modular approach (DCM) with training order:  $\{I\}\rightarrow \{D, A\}$ produces consistent, good images and converges faster (as shown in Figure~\ref{fig:digit-benchmarks},~\ref{fig:tvd-frontdoor},~\ref{fig:fid-frontdoor}). In Figure~\ref{fig:fid-diamond}, we show our performance for the graph in Figure~\ref{fig:diamond-graph} that contains two image variables. 
    %
    %
    }
   \label{fig:colored-mnist-1-experiment}
\end{figure*}
\section{Experimental Evaluation}
We present \WG performance on two semi-synthetic Colored-MNIST experiments and training convergence on a real-world COVIDx CXR-3 dataset provided in Appendix~\ref{sec:appex-covid}.
We also propose a solution to an invariant prediction problem for classification in CelebA-HQ.
For distributions and image quality comparison, we use metrics such as the total variation distance (TVD), the KL-divergence, and the Frechet Inception Distance (FID).
%
We share our implementation at \url{https://github.com/Musfiqshohan/Modular-DCM}.
%
%
%
%
%
%
%

\subsection{Semi-Synthetic Colored-MNIST Experiments}  
\label{sec:image-mediator}
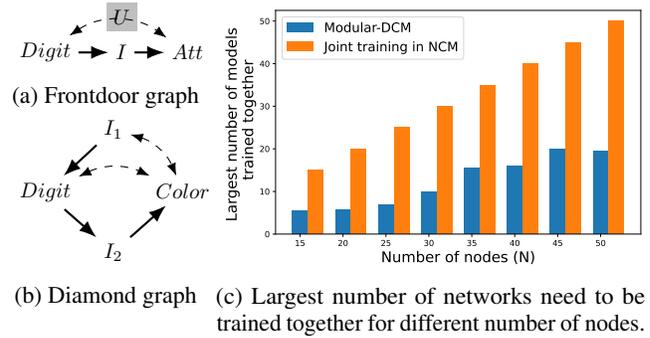
\begin{figure}[t!]
\begin{minipage}[c]{0.30\linewidth}
\begin{subfigure}{1\linewidth}
\centering
\begin{tikzpicture}[scale=0.8, transform shape]
 \node  (X) {$Digit$};
 \node  [right = 0.5cm of X](Z) {$I$};
 \node  [fill=lightgray, above = 0.1cm of Z] (U) {$U$};
 \node  [right = 0.5cm of Z](Y) {$Att$};
 \draw[ thick] (X) to  (Z); 
 \draw[ thick] (Z) to  (Y); 
 \path[bidirected] (X) edge[bend left=35] (Y);
\end{tikzpicture}
\caption{Frontdoor graph}
\label{fig:frontdoor-graph}
\end{subfigure}
\begin{subfigure}{1\linewidth}
\centering
\begin{tikzpicture}[scale=0.8, transform shape]
\node  [] (i1){$I_1$};
 \node  [below left = 0.5cm and 0.3cm of i1](D) {$Digit$};
 \node  [below right= 0.5cm and 0.3cm of i1](C) {$Color$};
 \node  [below = 1.5cm of i1](i2) {$I_2$};
 \draw[ thick] (i1) to  (D); 
 \draw[ thick] (D) to  (i2);
 \draw[ thick] (i2) to  (C);
 \path[bidirected] (i1) edge[bend left=25] (C);
 \path[bidirected] (D) edge[bend left=25] (C);
\end{tikzpicture}
\caption{Diamond graph}
\label{fig:diamond-graph}
\end{subfigure}
  \end{minipage} 
  \hspace{1mm}
  \begin{minipage}[c]{0.69\linewidth}
  \begin{subfigure}{1\linewidth}
\includegraphics[width=1.1\linewidth]{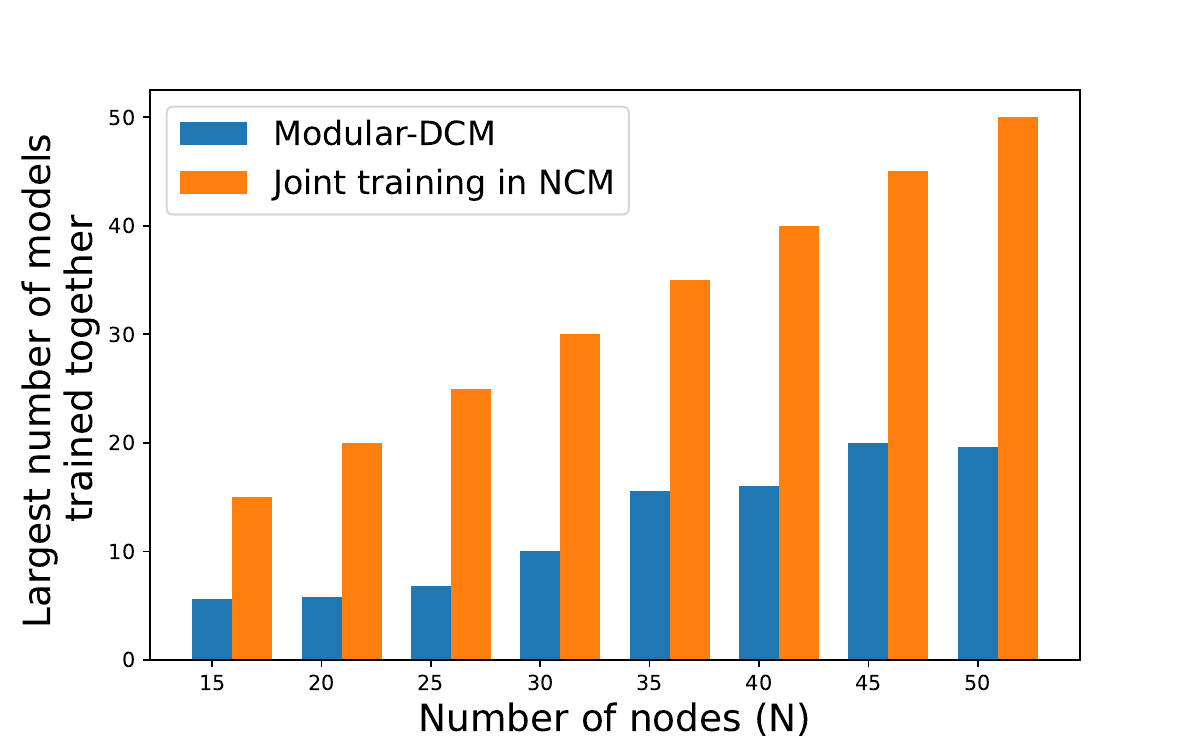}
  \caption{Largest number of networks need to be trained together for different number of nodes.}
  \label{fig:arbitrary-plot}
  \end{subfigure}
  \end{minipage} 
\caption{
Modular DCM on specific and arbitrary graphs.
}
\label{fig:mnist-exp-2}
\end{figure}

\textbf{MNIST frontdoor graph:}
We constructed a synthetic SCM that induces the graph in Figure~\ref{fig:frontdoor-graph}. 
Image variable $I$ shows an image of the digit value of $Digit (D)$.
We pick some random projection of the image as $Attribute (A)$ such that $P(A|do(D=0)) \neq P(A|do(D=1))$ holds, ensuring a strong causal effect.
A hidden variable $U$ affects both $D$ and $A$ such that $P(A|do(D))\neq P(A|D)$.
Suppose we are given a dataset $\mathcal{D}$ sampled from $P(D, A, I)$. 
Our goal is to estimate the causal effect $P(A|do(D))$.
We can use the backdoor criterion~\citep{pearl1993bayesian},
to measure the ground truth $P(A|do(D))$ $= \int_{U} P(A|D,U) P(U)$.
%
\par 
To estimate $P(A|do(D))$ by training on the observational dataset $\mathcal{D}[D,A,I]$, we construct the \WG architecture with a neural network $\mathbb{G}_D$ having fully connected layers to produce $D$, a CNN-based generator $\mathbb{G}_I$ to generate images, and a classifier $\mathbb{G}_A$ to classify MNIST images into variable $A$ such that $D$ and $A$ are confounded. 
Now, if we can train all mechanisms in the DCM to match $P(D, A, I)$, we can produce correct samples from $P(A|do(D))$. 
For this graph, the corresponding $\mathcal{H}$-graph is $[I]\rightarrow[D,A]$. 
Thus, we first train $\mathbb{G}_{I}$ by matching $P(I|D)$. Instead of training $\mathbb{G}_{I}$, we can also employ a pre-trained generative model that takes digits $D$ as input and produces an MNIST image showing $D$ digit in it. 
Next, we freeze $\mathbb{G}_{I}$ and train $\mathbb{G}_D$ and $\mathbb{G}_A$, to match the joint distribution $P(D, A, I)$ since $\{I\}$ is ancestor set $\mathcal{A}$ for c-component $\{D, A\}$.
Convergence of generative models becomes difficult using the loss of this joint distribution since the losses generated by both low and high dimensional variables are non-trivial to compare and re-weight (see Appendix~\ref{sub:appex-img-med}). Thus, we map samples of $I$ to a low-dimensional representation, $RI$ with a trained encoder and match $P(D, A, RI)$ instead of the joint $P(D, A, I)$. 
\par
\textbf{Evaluation}:
In Figure~\ref{fig:colored-mnist-1-experiment}, we compare our method with~\cite{xia2023neural}: NCM and a version of our method: \ABr that does not use modular training (to serve as ablation study).
First, we evaluate how each method matches the \Li and \Lii distributions in Figure~\ref{fig:tvd-frontdoor}.
Since NCM trains all mechanisms with the same loss function involving both low and high-dimensional variables, it learns marginal distribution $P(I)$ but does not fully converge to match  $P(A|\doo(D=0))$ finishing with TVD$=0.43$ at epoch 300.
DCM-Rep uses a low-dim representation of images: $RI$ and matches the joint distribution $P(D, A, RI)$ as a proxy to $P(D, A, I)$ without modularization. 
We observe DCM-Rep to converge slower (TVD$=0.36$ at epoch 300 for $P(A|\doo(D=0))$) compared to the original \WG.
Finally, \WG matches $P(D, A, RI)$ and converges faster with TVD$=0.08$ for $P(D,A)$ and $0.13$ for $P(A|do(D=0))$.
In Figure~\ref{fig:digit-benchmarks}, we show generated images of each method for ${\doo(D=0)}$(top) and ${\doo(D=1)}$(bottom) and evaluate their quality with FID scores in Figure ~\ref{fig:fid-frontdoor}.
We observe that NCM produces good-quality images (Figure~\ref{fig:digit-benchmarks} left, FID$=59.59$ at epoch 300) but is inconsistent with $do(D)$ intervention since it learns only marginal $P(I)$. \ABr generates consistent but low-quality images (Figure~\ref{fig:digit-benchmarks} middle, FID$=80.65$). \WG equipped with a pre-trained model produces good-quality (FID$=27.20$), consistent $P(I|do(D))$ images (Figure~\ref{fig:digit-benchmarks} right). To justify the necessity of modularity, we add another method (NCM Pre-trained) in our ablation study, where we equip it with a pre-trained model but have to match the original joint by training all mechanisms together, the same as NCM. Note that it starts with a low FID ($26.60$) but ends up worsening the image quality with FID$= 192.27$ (Figure~\ref{fig:fid-frontdoor}, row 2).

For a more rigorous evaluation, we use the effectiveness metric proposed in \cite{monteiro2023measuring} and
employ a classifier to map all images generated according to $P(I|\doo(D))$ back to discrete digits $D$. Next, we compute the exact likelihoods and compare with the true uniform intervention $\doo(D): [0.5, 0.5]$ that we perform for $P(I|\doo(D))$. We observe that the results are consistent with Figure~\ref{fig:digit-benchmarks}. 
NCM generated images are classified as $[0.19, 0.81]$, implying that NCM learns only marginal $P(I)$. On the other hand, DCM-Rep and DCM generated images are classified as uniform distribution with $98\%$ and $99\%$ accuracy.
%
%
%
%
%
%
\par \textbf{MNIST diamond graph:}
We illustrate our performance with a second synthetic SCM for the graph in Figure~\ref{fig:diamond-graph}. It contains multiple image nodes $\{I_1,I_2\}$ and discrete variables $Digit (D)$ and $Color (C)$. We consider a hidden confounders between $\{I_1, C\}$ and one between $\{D,C\}$. 
Baseline NCM matches $P(I_1, D, I_2, C)$ by training mechanisms of all variables at the same time. Whereas we utilize the modularity offered by c-components $\{I_1, D, C\}$ and $\{I_2\}$.
We i) first train $I_2$ to match $P(I_2|D)$ and then ii) train $I_1, D,C$ to match $P(I_1, D, I_2, C)$ while freezing weights of $I_2$. At the first step, $I_2$ trains well. In the 2nd step, since we don’t
have to train $I_2$ anymore, we optimize a less complex loss function compared to NCM. In Figure~\ref{fig:fid-diamond}, we compare the FID scores of $I_1$ and $I_2$ generated by \WG and NCM. 
We observe that while both matches $P(D,C)$ (thus TVD omitted), DCM achieves 
FID$(I_1)=32.88$ and FID$(I_2)=11.80$ while NCM achieves FID$(I_1)=80.54$ and FID$(I_2)=65.17$ after running for 300 epochs.
%

\textbf{Arbitrary graphs:}
In this experiment, we showcase the benefit of modularization over a random ensemble of graphs. We numerically visualize the largest number of mechanisms we have to update and train together compared to the full training of existing works. 
We sample random DAGs with a varying number of nodes ($N \in [15 - 50]$) keeping the arc ratio and the number of latents equal to $N/3$.
We call Algorithm~\ref{construct-H-graph}: {Construct\_H-graph(.)} to find the largest training component.
We took the average of five runs and plot it in Figure~\ref{fig:arbitrary-plot}. This plot demonstrates the number of networks that are trained together in a single training phase.
\begin{figure*}[t!]
\label{tb:celeb-graph-dist}
  \begin{minipage}[b]{0.20\linewidth}
\begin{tikzpicture}[scale=0.7, transform shape]
\tikzstyle{every node}=[]
 \node [ellipse, fill=lightgray] (d) {$Domain$};
  \node [text=red, below =0.6cm of d] (sex) {$Sex$};
  \node [ right =0.6cm of sex]     (eye) {$Eyeglass$};
 \node [ below left  =0.4cm and -0.5cm of eye] (i) {$Image$};
 \draw[ thick] (eye) to  (i);
 \draw[ thick] (sex) to  (i); 
 \draw[ thick] (d) to  (sex); 
 \path[bidirected] (eye) edge[bend right=35] (sex);
\end{tikzpicture}
  \end{minipage} 
\begin{minipage}[b]{0.83\linewidth}
\hspace{-7.5mm}
\includegraphics[width=1.0\linewidth]{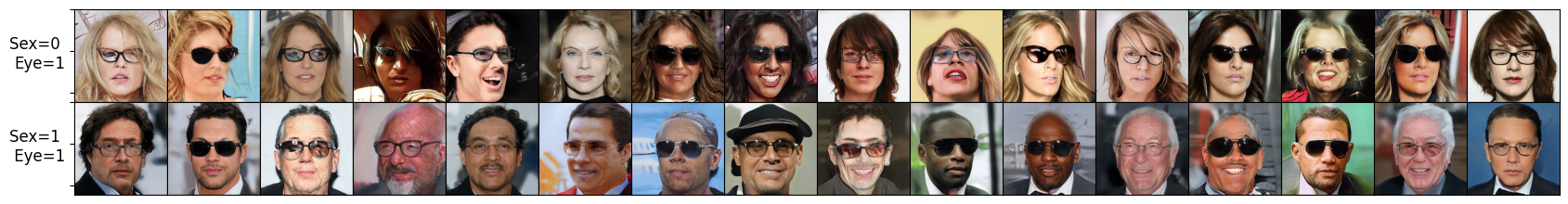}
 \end{minipage}

\begin{subfigure}{.20\textwidth}
\centering
\scriptsize
\begin{tabular}{|c|cc|}
\hline
          & \multicolumn{2}{c|}{Train}                 \\ \hline
          & \multicolumn{1}{c|}{EG=0} & EG=1 \\ \hline
Sex=0   & \multicolumn{1}{c|}{0.60}       & 0.018    \\ \hline
Sex=1 & \multicolumn{1}{c|}{ 0.18}       & 0.20     \\ \hline
          & \multicolumn{2}{c|}{Test}                  \\ \hline
Sex=0   & \multicolumn{1}{c|}{0.31}       & 0.14     \\ \hline
Sex=1 & \multicolumn{1}{c|}{0.47}       & 0.08     \\ \hline
\end{tabular}
\caption{Train \& Test distribution}
\label{fig:train-test-dist}
\end{subfigure}
\begin{subfigure}{.75\textwidth}
\scriptsize
\begin{tabular}{|l|l|l|l|l|l|l|l|l|}
\hline
\begin{tabular}[c]{@{}l@{}}Classifier/\\ Accuracy $\uparrow $\end{tabular}   & Sex=0 & Sex=1 & Eyeglass=0 & Eyeglass=1 & \begin{tabular}[c]{@{}l@{}}Sex=0\\ Eyeglass=0\end{tabular} & \begin{tabular}[c]{@{}l@{}}Sex=0\\ Eyeglass=1\end{tabular} & \begin{tabular}[c]{@{}l@{}}Sex=1\\ Eyeglass=0\end{tabular} & \begin{tabular}[c]{@{}l@{}}Sex=1\\ Eyeglass=1\end{tabular} \\ \hline
\texttt{TrainDomain}   & 0.948   & 0.946   & 0.971          & 0.847          & 0.998                                                          & 0.810                                                          & 0.953                                                          & 0.90                                                           \\ \hline
\texttt{IRM}   & 0.880   & 0.866   & 0.918          & 0.681          & 0.980                                                          & 0.600                                                          & 0.877                                                          & 0.800                                                           \\ \hline
\begin{tabular}[c]{@{}l@{}}\texttt{Intvention} \\ \texttt{(Ours)}\end{tabular} 
& 0.913   & 0.944   & 0.993          & 0.674          & 1.000                                                          & 0.669                                                          & 0.988                                                          & 0.68                                                           \\ \hline
\begin{tabular}[c]{@{}l@{}}\texttt{Augmented} \\ \texttt{(Ours)}\end{tabular}
& 0.970   & 0.981   & \textbf{0.994 }         & \textbf{0.905}          & 0.995                                                          & \textbf{0.901 }                                                         & 0.993                                                          & 0.91                                                           \\ \hline
\end{tabular}
\caption{Image samples and classifier accuracy for different sub-population}
\label{fig:table-accuracy}
\end{subfigure}
  \caption{
  (Top-Left): Invariant prediction causal graph.
  (Top-right) Images generated by \IG from $P(I|Sex,Eyeglass=1)$.
  (Bottom-left): Joint distribution of $P(Sex,Eyeglass)$.
  (Bottom-right): Eyeglass prediction accuracy of 3 classifiers in different sub-populations. Three classifiers are trained on the training dataset, the interventional dataset, and the augmented dataset (combined both). Note that the \texttt{Augmented} has better accuracy in the $Sex=0,Eyeglass=1$ sub-population which was our target to achieve.
  }
  \label{fig:celeb-graph-dist}
\end{figure*}
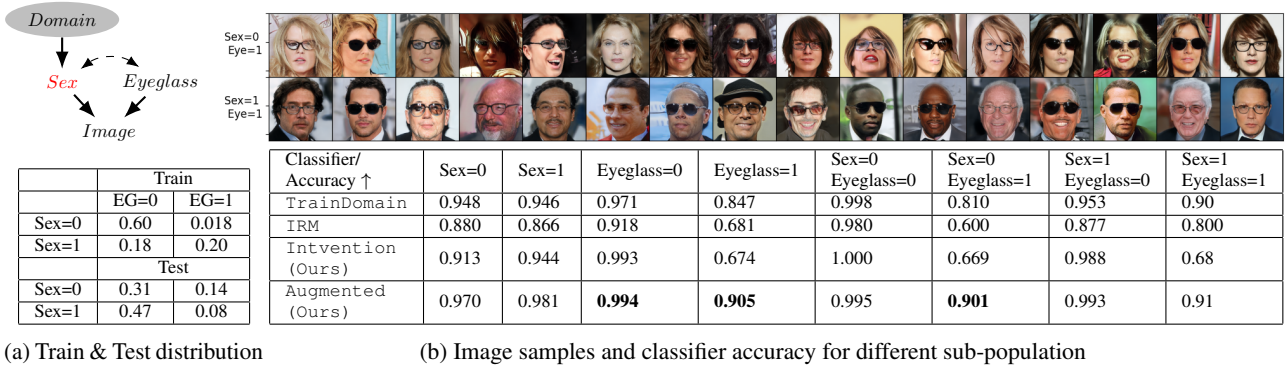
%

We observe that the number of models NCM trains together increases linearly with respect to $N$ whereas the growth in our method is relatively smaller since it does not depend on the number of nodes, but rather on the number of latents. 
For a graph of $N=50$ variables, NCM updates the $50$ networks corresponding to the mechanisms of all variables with a common loss function. Whereas, we train the same set of neural networks but modularly c-component by c-component with average max size of $20$ (for the setting in Figure~\ref{fig:arbitrary-plot}). We achieve better convergence since we minimize a less complex loss function at each training phase. We experience such convergence for Colored-MNIST with a low total variation distance compared to NCM (Figure~\ref{fig:fid-frontdoor}).
We discuss complexity evaluation of our algorithm in Appendix~\ref{appex:complex-eval}.
\subsection{Invariant Prediction on CelebA-HQ}
%
%
%
%
In this section, we design a causal invariant classifier $f$ for the high-dimensional image dataset, CelebA-HQ~\cite{CelebAMask-HQ} such that its specific attribute classification $Eyeglass = f(Image)$; does not experience low accuracy with domain shift.

\par \textbf{Motivation:}
Among all attributes of CelebA-HQ, some attributes, such as $Sex$ and $Eyeglass$ have spurious correlations between them~\cite{shen2020interpreting} (men are more likely to wear eyeglasses, correlation coefficient 0.47). 
A classifier trained on this dataset might consider the facial features of a male as an indicator to predict the presence of eyeglasses. 
As a result, if there is a shift in the sex distribution in the test domain, i.e., $P(Sex|domain=test$) $\neq$ $P(Sex|domain=train)$ (Figure~\ref{fig:train-test-dist}), and the classifier has to predict more images of females, it might have low eyeglass accuracy. 
For example, in Figure~\ref{fig:table-accuracy} (row-1), the accuracy for such a classifier in the $Sex=0,Eyeglass=1$ sub-population is 0.81 (comparatively lower).
We model the above scenario with the causal graph in Figure~\ref{fig:train-test-dist} (top-left).
We assume Eyeglass and Sex attributes determine how the Image variable would look like (shown with directed edges). The spurious correlation between the attributes is represented with a bi-directed edge. We reflect the distribution shift in $P(Sex)$, with an edge from the $Domain$ variable. 

To make the CelebA-HQ attribute classification independent of the domain shift, we employ causal invariant prediction.
Causal invariant prediction refers to the problem of learning a predictive model which is invariant to specific distribution shifts. 
According to~\citep{subbaswamy2019preventing, lee2023finding}, to build a causal invariant predictor,
we need to train it on an interventional dataset where the target attribute is independent of spuriously correlated/sensitive attributes due to the intervention performed. 
In our context, we need interventional samples from the high-dimensional interventional distribution $P(Eyeglass, Image|\doo( Sex))$ to train the invariant classifier, since the connection between $Domain$ and $Eyeglass$ is cut off by $\doo(Sex)$. 
The first step to obtain these interventional samples is to train a deep causal generative model and learn the observational distribution $P(Sex, Eyeglass, Image)$. 
For this purpose, we utilize our algorithm \WG.
%
%
%
%

\textbf{Dataset:}
The original CelebA-HQ dataset contains 1468 images of $Eyeglasses=1$. 
We distribute these samples among  5380 train samples and 1280 test samples maintaining the joint distribution in Figure~\ref{fig:train-test-dist}
such that the distribution shift in $P(Sex)$ is reflected across domains.

\textbf{Training:}
 Here we discuss three classifiers that we trained.
\texttt{TrainDomain}: 
This classifier is trained on the training dataset.
\CLb: 
According to ~\cite{subbaswamy2019preventing}, if we can generate a dataset $\mathcal{D}[Eyeglass, Image]\sim P(Eyeglass,Image|\doo(Sex))$ and train a classifier on this dataset, its prediction will be invariant to the $Domain$ and $Sex$, since intervention on $Sex$ removes their influence. To generate this high-dimensional interventional dataset, \WG employs neural networks $(\mathbb{G}_{Eyeglass}, \mathbb{G}_{Sex}, \mathbb{G}_I)$ for each of $Eyeglass, Sex$ and $Image$ and connects them according to the causal graph in Figure~\ref{fig:train-test-dist}. 
First, we train $\{\mathbb{G}_{Eyeglass}, \mathbb{G}_{Sex}\}$ together (same c-component).
Now, for $\mathbb{G}_I$, \WG's flexibility to incorporate pre-trained networks in its causal generative models allows us to utilize \IG~\citep{shen2020interpreting},
that uses StyleGAN~\citep{karras2019style} under the hood to produce realistic human faces. This pre-trained model plays an important role for a classifier since it needs to see realistic images and
training a model from scratch to match the true $P(I|Eyeglass,Sex)$ will be costly.
Next, we uniformly intervene on $Sex=0$ and $Sex=1$ and push forward through the trained models to generate 10k samples $\mathcal{D}\sim P(Eyeglass, Image|\doo(Sex))$  of both females and males. Finally, we train a new classifier \CLb on this dataset for eyeglass prediction.
$\texttt{Augmented}:$
To obtain the benefits of both classifiers, we create an augmented dataset by combining the training dataset and the interventional dataset generated by \WG. We train a new classifier $\texttt{Augmented}$ on it. 

%
%
\textbf{Evaluation}:
 We evaluate the accuracy 
of the classifiers in the test domain (Figure~\ref{fig:table-accuracy}). 
Since \CLa (row-1) learns the $Male$-$Eyeglass$ bias, it achieves accuracy= 0.90  in the $Sex=1, Eyeglass=1$ sub-population but it performs badly for $Sex=0, Eyeglass=1$ (accuracy 0.81). 
\CLb (row-3) is trained on images generated by \IG,  but due to the large support of the image manifold, samples generated by \IG from $P(Image|Eyeglass,Sex)$ might not represent all types of images that are present in the original CelebA-HQ dataset. For example, CelebA-HQ contains more variety of sunglasses compared to the \IG generated images (Figure~\ref{appex:i_given_eye} vs ~\ref{appex:eye1-sample}). As a result, \CLb does not perform very well in the $Eyeglass=1$ sub-population (0.67). However, the generated interventional dataset contains images of both $Sex=0,1$ wearing eyeglasses which are free from the training dataset bias. Thus, when we combine both datasets, samples from the training dataset introduce the \CLc classifier (row-4) to different varieties of $Eyeglass=1$ images and samples from the interventional dataset enforce it to focus on only eyeglass property in an image. We observe that \CLc improves accuracy in the three sub-populations (bolded in Figure~\ref{fig:table-accuracy}) : i)$\{Sex=0\}: 0.948 \rightarrow 0.97$ ii)$\{Eyeglass=1\}: 0.847 \rightarrow 0.905$ and iii)$\{Sex=0,Eyeglass=1\}: 0.810 \rightarrow 0.901$.
Thus, even though we have access to only the biased observational dataset, \WG offered a bias-free interventional dataset and enabled us to train a domain invariant classifier.

Note that we also evaluate the Invariant Risk Minimization (IRM)~\cite{InvariantRiskMinimization} method for this specific experiment although the problem setup of IRM and our proposed method are different. We train IRM on data from two environments: $Sex=0$ and $Sex=1$. We provide its accuracy in Figure~\ref{fig:table-accuracy} (row-2) and show that the classifiers trained according to our approach outperform it in most cases.
\section{Conclusion}
We propose a modular adversarial training algorithm for learning deep causal generative models and estimate causal effects with high-dimensional variables in the presence of confounders. 
After convergence, \WG can generate high-dimensional samples from identifiable interventional distributions. 
We assume the causal model to be semi-Markovian which aim to relax in our future work. 
Some potential application of our algorithm includes:
continual learning~\cite{busch2023continually} of deep causal generative models or high-dimensional interventional sampling in federated setting~\cite{vo2022adaptive}.
%

%
%
%

\section*{Acknowledgements}
This research has been supported in part by NSF CAREER 2239375, IIS 2348717, Amazon Research Award and Adobe Research.

\section*{Impact Statement}
Our proposed algorithm \WG, can sample from high-dimensional observational and interventional distributions.
As a result, it can be used to explore different creative directions such as producing realistic interventional images that we can not observe in real world.
We can train \WG models on datasets and perform an intervention on sensitive attributes to detect any bias towards them or any unfairness against them~\citep{xu2019achieving,van2021decaf}.
However, an adversary might apply our method to produce realistic images that are causal. As a result, it will be harder to detect fake data generated by DCM.

\bibliographystyle{icml2024}
\bibliography{references}

\clearpage
\appendix
\onecolumn

\section{Limitations and Future work}
Similar to most causal inference algorithms, we had to make the assumption of having a fully specified causal graph with latents, as prior.  We also assume each confounder to cause only two observed variables, which is considered as semi-Markovian in the literature. 
Another limitation of our work is that same as our close baselines, we do not consider conditional sampling. \WG can perform rejection sampling, which is practical if the evidence variables are low-dimensional.
With the advancements in causal discovery with latents, it might be possible to reliably learn part of the structure and leverage the partial identifiability results from the literature. Indeed, this would be one of the future directions we are interested in. We aim to extend our work for non-Markovian causal models where confounders can cause any number of observed variables.  We aim to resolve these limitations in our future work.

\section{\WG Training on Interventional Datasets}
Although Theorem~\ref{th:main:modular-train-converges} focuses training on observational data and sampling from interventional distributions, it can trivially be generalized to z-identifiability~\cite{bareinboim2012causal}.
That is we can generate samples from other interventional distributions that are non-identifiable from only observational data but are identifiable from a combination of both observational and interventional data. 
This can be trivially done by expanding the notion of identifiability to use a given collection of
interventional distributions, and requiring \WG to entail the same interventional distributions for the said collection. Thus in the appendix we provide proofs for both setups.

\section{ Appendix: Modular-DCM: Adversarial Training of Deep Causal Generative Models}
\label{appex:theo-discuss}
\begin{definition}[Identifiability~\citep{shpitser2007counterfactuals}]
\label{appex:def-id}
Given a causal graph, $G$, let $\textbf{M}$ be the set of all causal models that induce $G$ and objects $\phi$ and $\theta$ are computable from each model in $\textbf{M}$.
We define that $\phi$ is $\theta$-identifiable in $G$, 
 if there exists a deterministic function $g_G$ determined by the graph structure, such that $\phi$ can be uniquely computable as
$\phi = g_G(\theta)$ in any $M \in \textbf{M}$.
\end{definition}
\begin{definition}[Causal  Effects z-Identifiability]
Let $\textbf{X,Y,Z}$ be  disjoint  sets  of  variables in the causal graph $G$.
If $\phi=P_\textbf{x}(\textbf{y})$ is the causal effect  of  the action $\doo(\textbf{X=x})$ on the variables in $ \textbf{Y}$,  and $\theta$ contains $P(\bf{V})$ and interventional distributions $P(\bf{V}\setminus \bf{Z}'|\doo(\bf{Z}'))$, for all $\bf{Z}'\subseteq \bf{Z}$,
where $\phi$ and $\theta$ satisfies the definition of Identifiability, we define it as z-identifiabililty.
\citep{10.5555/3020652.3020668} proposes a z-identification algorithm to derive $g_G$ for these $\phi$ and $\theta$
\end{definition}

\subsection{\WG Interventional Sampling after Training}
\label{appex:wg-sampling}
\begin{figure}[H]
  \centering
\hspace*{-5mm}  
    \includegraphics[width=0.7\linewidth]{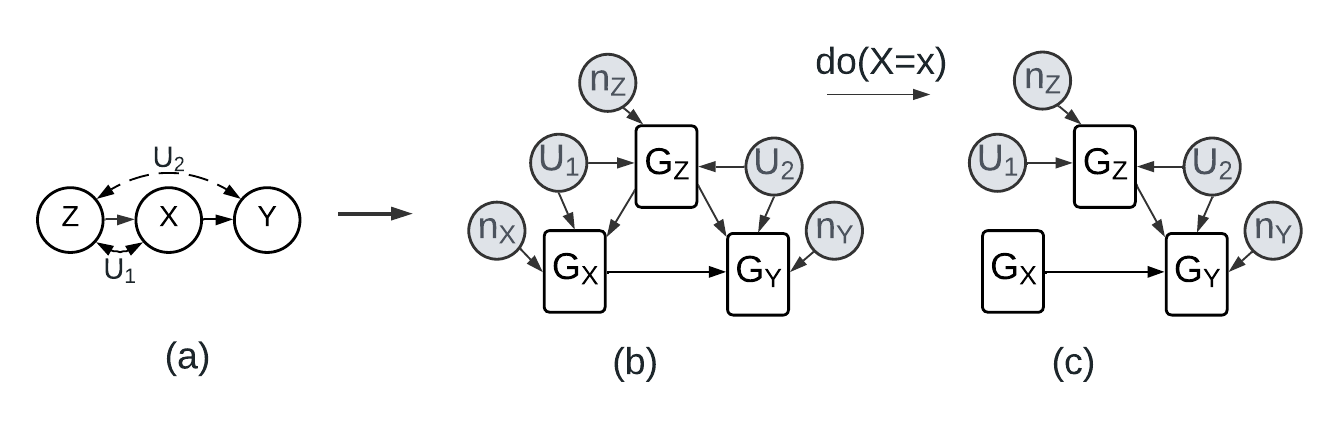}
   \caption{(a) Causal Graph with latents. (b), (c) DCM before and after intervention. }
   \label{appex:fig-wg-sampling}
\end{figure}

After \WG training, to perform hard intervention and produce samples accordingly, we manually set values of the intervened variables instead of using their neural network. Then, we feed forward those values into its children's mechanisms and generate rest of the variable like as usual. 
Figure~\ref{appex:fig-wg-sampling}(b) is the \WG network for the causal graph in Figure~\ref{appex:fig-wg-sampling}(a). Now, in Figure~\ref{appex:fig-wg-sampling}(c), we performed $\doo(X=x)$. Exogenous variables $U_1$ and $n_X$ are not affecting $X$ anymore as we manually set $X=x$. 

\subsection{Modular-DCM: Adversarial Training of Deep Causal Generative Models (Full Training)}
\label{appex:sec-identification}
Full training of the DCM indicates the setup when we update all mechanisms in the causal graph with the same common loss.
In this section, we prove that a trained DCM can sample from identifiable causal queries from any causal layer.
We assume $M_1$ as true SCM and $M_2$ as DCM of \WG.

\begin{theorem}
\label{th:appex-identifiability}
Let $\mathcal{M}_1 = (G=(\mathcal{V},\mathcal{E}), \mathcal{N}, \mathcal{U}, \mathcal{F}, P(.) )$ be an SCM. If a causal query $\mathcal{K}_{\mathcal{M}_1}(\mathcal{V})$ is identifiable from a collection of observational and/or interventional distributions $\{P_i(\mathcal{V})\}_{i\in[m]}$ for graph $G$, 
then any SCM $\mathcal{M}_2 = (G, \mathcal{N}', \mathcal{U}', \mathcal{F}', Q(.))$ entails the same answer to the causal query if it entails the same input distributions. Therefore, 
for any identifiable query $\mathcal{K}$, if $\{P_i(\mathcal{V})\}_{i\in[m]} \ID \mathcal{K}_{\mathcal{M}_1}(\mathcal{V})$ and 
$P_i(\mathcal{V})= Q_i(\mathcal{V}), \forall i\in [m]$, then $\mathcal{K}_{\mathcal{M}_1}(\mathcal{V})= \mathcal{K}_{\mathcal{M}_2}(\mathcal{V})$.
\begin{proof}
By definition of identifiability, we have that $\mathcal{K}_{\mathcal{M}_1} = g_G(\{P_i(\mathcal{V})\}_{i\in[m]})$ for some deterministic function $g_G$ that is determined by the graph structure. Since $\mathcal{M}_2$ has the same causal graph, the query $\mathcal{K}_{\mathcal{M}_2}$ is also identifiable and through the same function $g_G$, i.e., $\mathcal{K}_{\mathcal{M}_2}=g_G(\{Q_i(\mathcal{V})_{i\in [m]}\})$. Thus, the query has the same answer in both SCMs, if they entail the same input distributions over the observed variables, i.e., $P_i(\mathcal{V})=Q_i(\mathcal{V}), \forall i$. 
\end{proof}
\end{theorem}

\begin{figure}[H]

\begin{subfigure}{0.30\linewidth}
\centering
		\includegraphics[width=0.8\linewidth]{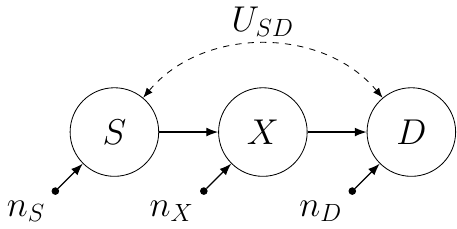}
		\caption{Front-door graph}
		\label{fig:appex-G-frontdoor}
	\end{subfigure}
\begin{subfigure}{0.37\linewidth}
\centering
		\includegraphics[width=0.8\linewidth]{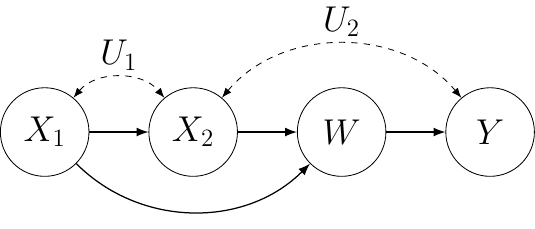}
		\caption{$\{P(\mathcal{V})$, $P_{x_1}(\mathcal{V})\} \ID P_{x_1,x_2}(Y|x_1',x_2')$}
		\label{fig:appex-G-CF}
	\end{subfigure}
\begin{subfigure}{0.30\linewidth}
		\includegraphics[width=0.8\linewidth]{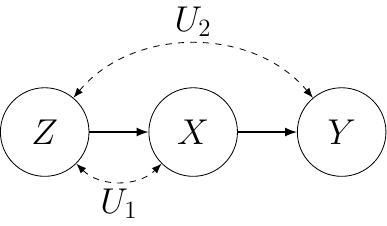}
		\caption{$\{P(\mathcal{V})$, $P_z(X,Y)\} \ID P_{x}(Y)$}
		\label{fig:appex-G-zidentify}
	\end{subfigure}
 \caption{Causal graphs with latents and respective identifiable causal queries. $\theta$ identifies $\phi$ :$\theta \ID \phi$}
   \label{fig:appex-theory-explain-graph}
\end{figure}

\begin{corollary}
\label{cor:id-obs}
 Let $\mathcal{M}_1 = (G=(\mathcal{V},\mathcal{E}), \mathcal{N}, \mathcal{U}, \mathcal{F}, P(.) )$ and 
 $\mathcal{M}_2 = (G, \mathcal{N}', \mathcal{U}', \mathcal{F}', Q(.))$ be two SCMs. 
 If $\{P(\mathcal{V})\}\ID P_x(Y)$ for $X,Y\subset \mathcal{V}$, $X\cap Y=\emptyset$ and 
 $P(\mathcal{V})= Q(\mathcal{V})$
 then $P_x(Y)=  Q_x(Y)$ 
\end{corollary}  
For example, in Figure \ref{fig:appex-theory-explain-graph}(b), the interventional query $P_{x_1,x_2}(W)$ is identifiable from $P(\mathcal{V})$.
According to the Corollary~\ref{cor:id-obs}, after training on $P(\mathcal{V})$ dataset, \WG will produce correct interventional sample from $P_{x_1,x_2}(W)$ and along with other queries in $\layer_2(P(\mathcal{V}))$.
\par
\cite{bareinboim2012causal} showed that we can identify some \Lii-queries with other surrogate interventions and \Li-distributions. Similarly, we can apply  Theorem~\ref{th:appex-identifiability}:
\begin{corollary}
\label{id-intv}
 Let $\mathcal{M}_1 = (G=(\mathcal{V},\mathcal{E}), \mathcal{N}, \mathcal{U}, \mathcal{F}, P(.) )$
 and $\mathcal{M}_2 = (G, \mathcal{N}', \mathcal{U}', \mathcal{F}', Q(.))$
 be two SCMs and $X,Y$ be disjoint, and $\{S_i\}_i$ arbitrary subsets of variables.
 If  $i) \{P(\mathcal{V}), P_{s_1}({\mathcal{V}}),P_{s_2}({\mathcal{V}})\hdots\}\ID P_x(Y)$, 
 $ii) P(\mathcal{V}) = Q(\mathcal{V})$ and
$iii) P_{{s_i}}({\mathcal{V}}) 
=Q_{s_i}({\mathcal{V}})$, $\forall i,s_i$ 
then $P_x(Y)= Q_x(Y)$.
\end{corollary}
In Figure \ref{fig:appex-theory-explain-graph}(c), the interventional query $P_{x}(Y)$ is identifiable from $P(\mathcal{V})$ and $P_z(X,Y)$. Therefore, after being trained on datasets sampled from these distributions,
\WG will produce correct interventional sample from $P_{x}(Y)$ and all other queries in $\layer_2(P(\mathcal{V})$, $P_z(X,Y))$.

\setlength{\textfloatsep}{5pt}
\begin{algorithm}[t!]
\small
	\caption{\WG Training on Multiple Datasets}
    \label{alg:train_DCMmechanisms}
\begin{algorithmic}[1]
   \STATE {\bfseries Input:} Causal Graph $G=(\mathcal{V}$, $\mathcal{E})$, Interventional datasets= $(\mathbf{I}, \mathcal{D})$,  DCM $\mathbb{G}$, Critic $\mathbb{D}$,  Parameters= $\theta_1, \hdots, \theta_n$, $\lambda = 10$
   
   \WHILE{$\theta_1, \hdots, \theta_n$ has not converged}
   \FOR{{\bfseries each} $(X, D) \in (\mathbf{I}, \mathcal{D})$}  
\STATE   $compare\_var = \mathcal{V}$ 
\STATE Sample real data $\mathbf{v_x^r}\sim D$ 
following the distribution $\mathbb{P}_x^r$ with intervention $X$.
\STATE $\mathbf{x} \gets X.values$ \hspace{3mm} \gray{//$X$=($keys, values$)}
\STATE $\mathbf{v}_x^f =$ RunGAN($G, \mathbb{G}, X, compare\_var, \emptyset)$
\STATE $\hat{\mathbf{v}}_x= \epsilon \mathbf{v}_x^r + (1 - \epsilon)\mathbf{v}_x^f $
\STATE $L_{x} = \mathbb{D}_{w_x} (\mathbf{v}_x^f) - \mathbb{D}_{w_x} (\mathbf{v}_x^r)
* \lambda(\left\Vert \nabla_{\hat{\mathbf{v}}_x}\mathbb{D}_{w_x}(\hat{\mathbf{v}}_x)\right\Vert_2 - 1)^2 $
\STATE $G_{loss} = G_{loss}  + \mathbb{D}_{w_x} (\mathbf{v}_x^f)$
   \STATE $w_x = Adam(\nabla_{w_x} \frac{1}{m} \sum_{j=1}^m L_{\mathbb{D}_x}, w_x, \alpha, \beta_1, \beta_2)$
      \ENDFOR
   \FOR{$\theta \in \theta_1, \hdots \theta_n$}
   \STATE $\theta = Adam(\nabla_{\theta} -G_{loss}
,\theta, \alpha, \beta_1, \beta_2 )$
   \ENDFOR
   \ENDWHILE
   \STATE {\bfseries Return:} $\theta_1, \hdots \theta_n$
\end{algorithmic}
\end{algorithm}
\subsection{Training with Multiple Datasets
}
We propose a method in Algorithm~\ref{alg:train_DCMmechanisms} for training \WG with both $\mathcal{L}_1$ and $\mathcal{L}_2$ datasets. We use Wasserstein GAN with penalized gradients (WGAN-GP)~\citep{gulrajani2017improved} for adversarial training. We can also use more recent generative models such as diffusion models when variables are not in any c-component. $\mathbb{G}$ is the DCM, a set of generators and $\{\mathbb{D}_{x}\}_{ X\in \mathbf{I}}$ are a set of discriminators for each intervention value combinations.
The objective function of a two-player minimax game would be
\begin{equation*}
\label{eq:gan-loss}
    \begin{split}
        &\min_{\mathbb{G}}
        \sum_x 
        \max_{\mathbb{D}_x} L(\mathbb{D}_x,\mathbb{G}),
        \\  L(\mathbb{D}_x,\mathbb{G})=& \mathop{\mathbb{E}}_{v\sim \mathbb{P}_{x}^{r}}[\mathbb{D}_x(\mathbf{v})] - 
        \mathop{\mathbb{E}}_{ \mathbf{z}\sim \mathbb{P}_{Z},
        \mathbf{u}\sim \mathbb{P}_{U}}
        [\mathbb{D}_x(\mathbb{G^{(\mathbf{x})}}(\mathbf{z}, \mathbf{u}))]
    \end{split}
\end{equation*}
Here, for intervention $\text{do}(X=x), X\in \mathbf{I}$, $\mathbb{G}^{(x)}({\mathbf{z,u}})$ are generated samples and $v\sim \mathbb{P}^{r}_{x}$ are real $\mathcal{L}_1$ or $\mathcal{L}_2$ samples.
 We train our models by iterating over all datasets and learn $\mathcal{L}_1$ and $\mathcal{L}_2$ distributions (line 3). We produce generated interventional samples by intervening on the corresponding node of our architecture. For this purpose, we call Algorithm~\ref{alg:RunGAN} RunGAN(), at line 7. 
 We compare the generated samples with the input $\mathcal{L}_1$ or $\mathcal{L}_2$ datasets. 
 For each different combination of the intervened variables $x$, $\mathbb{D}_{x}$ will have different losses, $L_{X=x}$ from each discriminator (line 9).
At line 10, we calculate and accumulate the generator loss over each dataset.
If we have $V_1,\hdots, V_n \in \mathcal{V}$, then we update each variable's model weights based on the accumulated loss (line 13). 
 This will ensure that after convergence, \WG models will learn distributions of all the available datasets and according to Theorem~\ref{th:appex-identifiability}, it will be able to produce samples from same or higher causal layers queries that are identifiable from these input distributions.
Following this approach, \WG Training in Algorithm~\ref{alg:train_DCMmechanisms}, will find a DCM solution that matches to all the input distributions, mimicking the true SCM. 
Finally, we describe  sampling method for \WG after training convergence in Appendix~\ref{appex:wg-sampling}.
%
%
\begin{proposition}
\label{appex:alg1-matches-pIv}
Let $\mathcal{M}_1$ be the true SCM and Algorithm~\ref{alg:train_DCMmechanisms}:  \textbf{\WG Training} converges
after being trained on datasets: $\mathbf{D}= \{\mathcal{D}_i\}_i$, outputs the DCM $\mathcal{M}_2$. If for any causal query $\mathcal{K}_{\mathcal{M}_1}(\mathcal{V})$ identifiable from $\mathbf{D}$ then $\mathcal{K}_{\mathcal{M}_1}(\mathcal{V})= \mathcal{K}_{\mathcal{M}2}(\mathcal{V})$
\end{proposition}
\begin{proof}
Let $\mathcal{M}_1 = (G=(\mathcal{V},\mathcal{E}), \mathcal{N}, \mathcal{U}, \mathcal{F}, P(.) )$ be the true SCM and $\mathcal{M}_2 = (G, \mathcal{N}', \mathcal{U}', \mathcal{F}', Q(.))$ be the deep causal generative model represented by Modular-DCM. \WG Training converges implies that $Q_i(\mathcal{V})=P_i(\mathcal{V}), \forall i\in [m]$ for all input distributions.
Therefore, according to Theorem~\ref{th:appex-identifiability}, \WG is capable of producing samples from correct interventional distributions that are identifiable from the input distributions.
\end{proof}

\subsection{Non-Markovianity}
\label{non-Markov-to-semi-Markov}

Note that, one can convert a non-Markovian causal model $M_1$ to a semi-Markovian causal model $M_2$ by taking the common confounder among the observed variables and splitting it into new confounders for each pair. 
Now, for a causal query to be unidentifiable in a semi-Markovian model $M_2$, we can apply the Identification algorithm ~\citep{shpitser2008complete} and check if there exists a hedge. The unidentifiability of the causal query does not depend on the confounder distribution. Thus, if the causal query is unidentifiable in the transformed semi-Markovian model $M_2$, it will be unidentifiable in the original non-Markovian model $M_1$ as well.

Besides Semi-Markovian, Theorem~\ref{th:identifiability} and Theorem~\ref{th:appex-identifiability} holds for Non-Markovian models, with latents appearing anywhere in the graph and thus can be learned by \WG training.
\cite{jaber2019causal} performs causal effect identification on equivalence class of causal diagrams, a partial ancestral graph (PAG) that can be learned from observational data. Therefore, we can apply their method to check if an interventional query is identifiable from observational data in a Non-Markovian causal model and express the query in terms of observations and obtain the same result as Theorem~\ref{th:appex-identifiability}. We aim to explore these directions in more detail in our future work.

\section{ Appendix: \WG Modular Training}
\label{appex:sec-Modular Training}

\subsection{ Tian's Factorization for Modular Training}
\label{appex:tian-fact-modular-train}
\label{append:main_paper_example}
\begin{figure}[t!]
  \centering
        \includegraphics[width=0.5\linewidth]{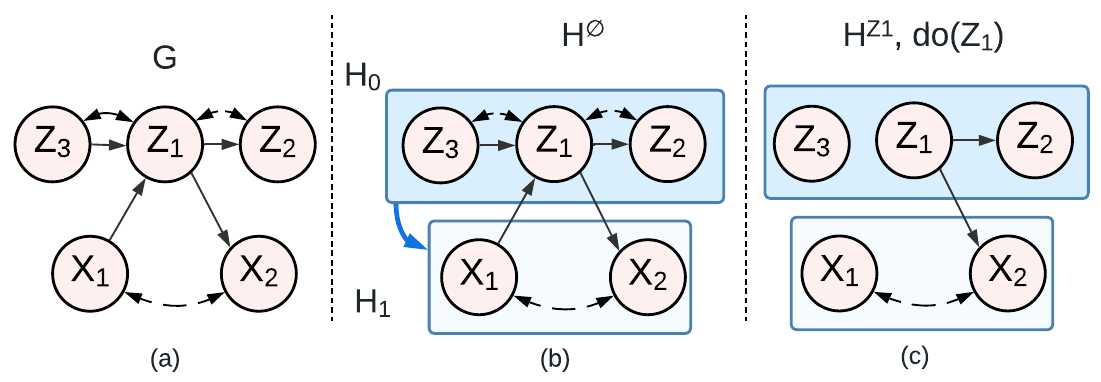}
   \caption{(a) Causal graph $G$, (b) $\mathcal{H}^\emptyset$-graph, (c) $H^{Z_1}$-graph
   }
   \label{appex:mechanism-graphsx}
\end{figure}

In Figure~\ref{appex:mechanism-graphsx}(a), We apply Tian's factorization~\citep{tian2002general} to get,
%

\begin{equation}
    \label{eq:appex-tian-correct-fact}
    P(v)= P(x_1,x_2|\doo(z_1))  P(z_1,z_2,z_3|\doo(x_1))
\end{equation}
We need to match the following distributions with the DCM.
\begin{equation}
\label{eq:appex-tian-fake-intv-vs-true-intv}
    \begin{split}
        P(x_1,x_2|\doo(z_1))= Q(x_1,x_2|\doo(z_1))\\
        P(z_1, z_2, z_3| \doo(x_1))= Q(z_1, z_2, z_3| \doo(x_1))
    \end{split}
\end{equation}

With modular training, we matched the following alternative distributions:
\begin{equation}
\label{eq:matched-modular}
    \begin{split}
            P(z_1,z_2,z_3|x_1)&= Q(z_1,z_2,z_3|\doo(x_1))\\
    P(x_1,x_2,z_1,z_3)&= Q(x_1,x_2,z_1,z_3)\\
    \end{split}
\end{equation}

Now, for the graph in Figure~\ref{appex:mechanism-graphsx}(a),
\begin{equation}
    \begin{split}
     P(x_1,x_2,z_1,z_2, z_3)&=  P(x_1,x_2|\doo(z_1)) \times P(z_1,z_2,z_3|\doo(x_1)\\
         &=\frac{P(x_1,x_2,z_1, z_3)}{P(z_1,z_3|\doo(x_1))} \times P(z_1,z_2, z_3|x_1) \hspace{5mm} \text{[C-factorization of $P(x_1,x_2,z_1, z_3)$]}\\ 
         &=\frac{P(x_1,x_2,z_1,z_3)}{P(z_1,z_3|x_1)} \times P(z_1,z_2,z_3|x_1)
         \hspace{5mm} \text{[Do-calculus rule-2 applies]}\\ 
         &=\frac{P(x_1,x_2,z_1,z_3)}{\sum_{z_2}P(z_1,z_2,z_3|x_1)} \times P(z_1,z_2,z_3|x_1)\\
    &=\frac{Q(x_1,x_2,z_1,z_3)}{\sum_{z_2}Q_{x_1}(z_1,z_2,z_3)} \times Q_{x_1}(z_1,z_2,z_3)
  \hspace{5mm}   \text{[According to Equation~\ref{eq:matched-modular}]}
    \\
         &=Q(x_1,x_2,z_1,z_2) \hspace{5mm} \text{[We can follow the same above steps as $P(.)$ for $Q(.)$]}
    \end{split}
\end{equation}
Therefore, if we match the distributions in Equation~\ref{eq:matched-modular} with the DCM, it will match $P(\mathcal{V})$ as well.

\subsection{ Modular Training for Interventional Dataset}
\subsubsection{Modular Training  Basics}

\begin{figure}[H]
  \centering
        \includegraphics[width=1\linewidth]{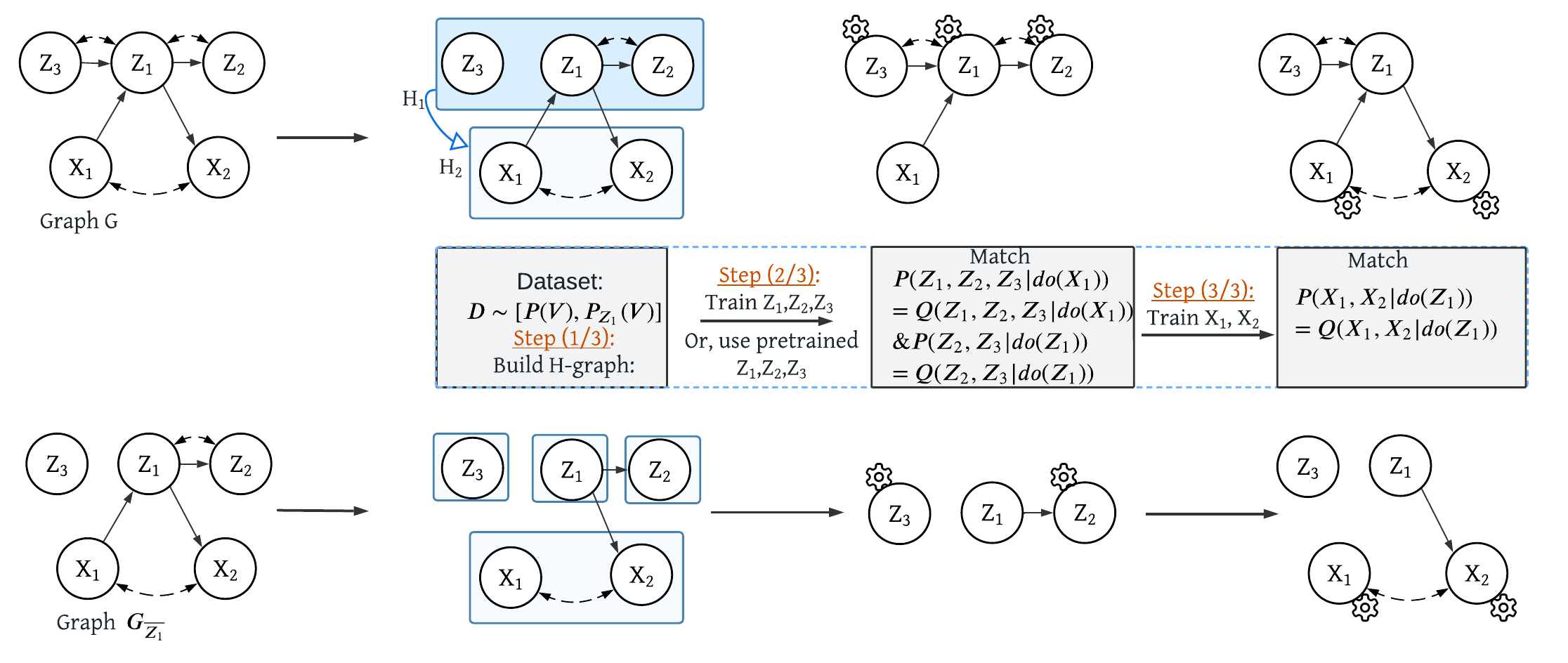}
   \caption{
   Modular training on H-graph: $H_1:[Z_1,Z_2,Z_3]\rightarrow H_2:[X_1,X_2]$ with dataset $D\sim P(V)$.
   }
   \label{fig:modular-simulation-I}
\end{figure}

Suppose, for the graph in Figure~\ref{fig:modular-simulation-I}, we have two datasets $D^{\emp} \sim P(\mathcal{V})$ and $D^{z_1}\sim P_{Z_1}(V)$, i.e., intervention set $\mathcal{I}=\{\emptyset, Z_1\}$.
Joint distributions in both dataset factorize like below:
\begin{equation}
\label{eq:appex-fake-intv-vs-true-intv}
    \begin{split}
        P(v)&= P_{z_1}(x_1,x_2)  P_{x_1}(z_1,z_2,z_3)\\
        P_{z_1}(v)&= P_{z_1}(x_1,x_2)  P(z_3) P_{z_1}(z_2)\\
                &= P_{z_1}(x_1,x_2)  P_{z_1}(z_2,z_3)  \text{[Since $Z_2,Z_3$ independent in $G_{\overline{Z_1}}$ graph]}\\
                 &= P_{z_1}(x_1,x_2)  P_{x_1, z_1}(z_2,z_3) \text{[Ignores intervention using do calculus rule-2]}\\
    \end{split}
\end{equation}

We change the c-factors for $P_{z_1}(V)$ to keep the variables in each c-factor same in all distributions.
This factorization suggests that to match $P(\mathcal{V})$ and $P_{Z_1}(V)$ we have to match each of the c-factors using $D^{\emp}$ and $D^{z_1}$ datasets. 
In Figure~\ref{fig:modular-simulation} graph $G$, $P_{x_1}(z_1,z_2,z_3)=P(z_1,z_2,z_3|x_1)$ since do-calculus rule-$2$ applies. And in $G_{\overline{Z_1}}$, $P(z_3) P_{z_1}(z_2)$ can be combined into $P_{x_1, z_1}(z_2,z_3)$.
Thus we can use these distributions to train part of the DCM: $\mathbb{G}_{Z_1}, \mathbb{G}_{Z_2}, \mathbb{G}_{Z_3}$ to learn both $Q(z_1,z_2,z_3|\doo(x_1))$ and  $Q(z_2,z_3|\doo(x_1,z_1))$. 
However, $P(x_1,x_2|\doo(z_1))\neq P(x_1,x_2|z_1)$ in $P(\mathcal{V})$. But we have access to $P_{z_1}(\mathcal{V})$. Thus, we can train $\mathbb{G}_{X_1}, \mathbb{G}_{X_2}$ with only dataset $D^{Z_1} \sim P_{z_1}(\mathcal{V})$ (instead of both $D^{\emp},D^{Z_1}$) and learn $Q(x_1,x_2|\doo(z_1))$. This will ensure the DCM has matched both $P(\mathcal{V})$ and $P_{z_1}(\mathcal{V})$ distribution.

Thus, we search proxy distributions to each c-factor corresponding to both $P(\mathcal{V})$ and $P_{Z_1}(V)$ dataset, to train the mechanisms in a c-component $\mathbf{Y}$. 
For each of the c-factors corresponding to $\mathbf{Y}$ in $P(\mathcal{V})$ and $P_{z_1}(V)$, we search for two ancestor sets $\mathcal{A}_{\emp},\mathcal{A}_{Z_1}$ in both $P(\mathcal{V})$ and $P_{z_1}(V)$ datasets such that 
the parent set $Pa(\mathbf{Y} \cup \mathcal{A}_{\emp})$ 
satisfies rule-$2$ for the joint $\mathbf{Y} \cup \mathcal{A}_{\emp}$
and $Pa(\mathbf{Y} \cup \mathcal{A}_{Z_1})$ satisfies rule-$2$ for the joint $\mathbf{Y} \cup \mathcal{A}_{Z_1}$ with $\doo(Z_1)$ intervention.

We update Definition~\ref{def:modularity-condition}as \textbf{modularity condition-I} for multiple interventional datasets as below:

\begin{definition}[Modularity condition-I]
	\label{def:modularity-condition-I}
Given a causal graph $G$, an intervention $I\in \mathcal{I}$  and a c-component variable set  $\mathbf{Y}$, a set $\mathcal{A}\subseteq An_{G_{\overline{I}}} (\mathbf{Y})\setminus \mathbf{Y}$ is said to satisfy the modularity condition if it is the smallest set that satisfies 
$P(\mathbf{Y}\cup\mathbf{X}| \doo(Pa(\mathbf{Y} \cup \mathbf{X})), \doo(I)) 
= P(\mathbf{Y}\cup\mathbf{X}| Pa(\mathbf{Y} \cup \mathbf{X}), \doo(I))$, i.e., do-calculus rule-$2$~\citep{pearl1995causal} applies. 
\end{definition}

Unlike before, we have access to $P_{z_1}(\mathcal{V})$ and we can use that to match $P_{z_1}(x_1,x_2)$ in both cases.
To match the \Li and \Lii joint distributions according to (\ref{eq:appex-fake-intv-vs-true-intv}), we train each c-component one by one.
For each c-component, we identify the modularity conditions 
of all c-factors $P_{pa(\mathbf{Y}) \cup I}(\mathbf{Y})$, $\forall I\in \mathcal{I}$  and use them to train $\mathbf{Y}$. 
We train the mechanisms in $\mathbf{Y}$ to learn an alternative to each c-factor $P_{pa(\mathbf{Y}) \cup I}(\mathbf{Y})$, $\forall I\in \mathcal{I}$. 
For some ancestor set $\mathcal{A}_I$, the alternative distribution is in the form $P(\mathbf{Y}\cup \mathcal{A}_I| \doo(Pa(\mathbf{Y} \cup \mathcal{A}_I)), \doo(I))$ which should be equivalent to $P(\mathbf{Y}\cup \mathcal{A}| Pa(\mathbf{Y} \cup \mathcal{A}), \doo(I))$. 
We will find an $\mathcal{A}_I$ from the $D^I, \forall I\in \mathcal{I}$ such that we do not require $Pa(\mathbf{Y} \cup \mathcal{A})$ to be intervened on.
\par
Now, to match $P(\mathbf{Y}\cup\mathcal{A}| Pa(\mathbf{Y} \cup \mathcal{A}), \doo(I))$ = $Q(\mathbf{Y}\cup\mathcal{A}| \doo(Pa(\mathbf{Y} \cup \mathcal{A})), \doo(I))$ with our generative models, we pick the observations of $Pa(\mathbf{Y} \cup \mathcal{A})$ from $D^I$ dataset and intervene in our DCM with those values besides intervening on $\mathbb{G}_{I}$.
Since we do not need generated samples for $Pa(\mathbf{Y} \cup \mathcal{A})$ from DCM, rather their observations from the given $D^I$ dataset, we do not require them 
to be trained beforehand. 
However, the order in which we train c-components matters and we follow the partial order found for $P(\mathcal{V})$ dataset even thought we train with multiple datasets.
\par 
For example,  in Figure~\ref{fig:modular-simulation}, we have two graphs $G$  and $G_{\overline{Z_1}}$.
We follow $G$'s training order for both graphs to train the c-components, i.e., $[\mathbb{G}_{Z_1}, \mathbb{G}_{Z_2}, \mathbb{G}_{Z_3}] \rightarrow [\mathbb{G}_{X_1},\mathbb{G}_{X_2}]$.
Here, for the c-component $\mathbf{Y}=\{Z_1,Z_2, Z_3\}$, we match  $P(\mathcal{V})$ c-factor $P_{x_1}(z_1,z_2,z_3)$ and $P_{z_1}(\mathcal{V})$ c-factor $P_{x_1,z_1}(z_2,z_3)$ thus have to find alternative distribution for them. We find the smallest ancestor set $\mathcal{A}_{\emp}, \mathcal{A}_{Z_1}$ for these c-factors in both 
$D^{\emp}$ and $D^{Z_1}$ datasets.
$\mathcal{A}_{\emptyset}=\emptyset$ satisfies modularity condition for $P(\mathcal{V})$ c-factor  and their $Pa(\mathbf{Y} \cup \mathcal{A}) = \{X_1\}$. $\mathcal{A}_{Z_1}=\emptyset$ satisfies modularity condition for $P_{z_1}(\mathcal{V})$ c-factor and their $Pa(\mathbf{Y} \cup \mathcal{A}) = \emptyset$.
At step (2/3) in Figure~\ref{fig:modular-simulation}, 
We do not need  $\mathbb{G}_{X_1}$  to be pre-trained.
$[\mathbb{G}_{Z_1}, \mathbb{G}_{Z_2}, \mathbb{G}_{Z_3}]$ converges by matching both $P(z_1,z_2,z_3|x_1)= Q_{x_1}(z_1,z_2,z_3)$ and $P_{x_1,z_1}(z_2,z_3)= Q_{x_1,z_1}(z_2,z_3)$.

\par 
Now, we train mechanisms of the next c-component $[\mathbb{G}_{X_1}, \mathbb{G}_{X_2}]$  in our training order (step 3/3). 
We have to match  $P(\mathcal{V})$ c-factor $P_{z_1}(x_1,x_2)$ and $P_{Z_1}(\mathcal{V})$ c-factor $P_{z_1}(x_1,x_2)$.
Ancestor set $\mathcal{A}_{\emp}= \{Z_1,Z_3\}$ satisfies the modularity condition for $\mathbf{Y}=\{X_1,X_2\}$ with $P(\mathcal{V})$ dataset but  $\mathcal{A}_{Z_1}= \emptyset$ a smaller set, satisfies the modularity condition for same c-factor with $P_{z_1}(\mathcal{V})$ dataset. Also, $P_{z_1}(\mathcal{V})$ c-factor is $P_{z_1}(x_1,x_2)$. Thus if we train $[\mathbb{G}_{X_1},\mathbb{G}_{X_2}]$ with only $P_{Z_1}(\mathcal{V})$ dataset, it will learn both c-factors and converge with $P_{z_1}(x_1,x_2)= Q_{z_1}(x_1,x_2)$. Since we have matched all the c-factors, our DCM will match both $P(\mathcal{V})$ and $P_{Z_1}(\mathcal{V})$ distributions.
 During training of $[\mathbb{G}_{X_1}, \mathbb{G}_{X_2}]$, we had $\mathcal{A}=\emptyset$ for both observation and interventional c-factors. Therefore, we do not need any pre-trained mechanisms, rather we can directly use the observations from  $P_{Z_1}(\mathcal{V})$ dataset as parent values.
We define $\mathcal{H}^I$-graph for each $I\in \mathcal{I}$ as below:
\begin{definition}[$\mathcal{H}^I$-graph]
\label{def:HI-graph}
For a post-interventional graph $G_{\overline{I}}$, let the set of c-components in $G_{\overline{I}}$ be $\mathcal{C}=\{C_1,\hdots C_t\}$. Choose a partition  $\{H^I_k\}_k$ of $\mathcal{C}$ such that the $\mathcal{H}^I$-graph $\mathcal{H}^I=(V_{\mathcal{H}^I},E_{\mathcal{H}^I})$, defined as follows, is acyclic: 
$V_{\mathcal{H}^I}= \{H^I_k\}_k$ and 
for any $s,t$, $H^I_s \rightarrow H^I_t \in E_{\mathcal{H}^I}$, iff
$P(H^I_t|\doo(pa(H^I_t) \cap H^I_s)) \neq P(H^I_t|pa(H^I_t) \cap H^I_s)$, i.e., do-calculus rule-$2$ does not hold. Note that one can always choose a partition of $\mathcal{C}$ to ensure $\mathcal{H}^I$ is acylic: The $\mathcal{H}^I$ graph with a single node $H^I_1 =\mathcal{C}$ in $G_{\overline{I}}$.
Even though $\mathcal{H}^I$ for different $I$ might have different partial order, during training, every $\mathcal{H}^I$ follows the partial order of $\mathcal{H}^{\emptyset}$.
 Since its partial order is valid for other H-graphs as well (Proposition~\ref{appex-lemma:consistent-hgraph}).
\end{definition}

Here, $\mathcal{H}^{I}$ is the $\mathcal{H}$-graph constructed from $G_{\overline{I}}$, for $I\in \mathcal{I}$ where $\mathcal{I}$ is the intervention set. 
$\mathcal{H}^{\emptyset}$ is the $\mathcal{H}$-graph constructed from $G$ for observational training.
$H_k^I \coloneqq$ $k$-th h-node in the $\mathcal{H}^{I}$ graph.
During $H^I$-graphs construction, we resolve cycles by combining c-components on that cycle into a single h-node. Please check example in Figure~\ref{appex:fig-long-hgraph}. 
After merging all such cycles, $H^I, \forall I \in \mathcal{I}$ will become directed acyclic graphs.
The partial order of this graph will indicate the training order that we can follow to train all variables in $G$. 
For example in Figure~\ref{fig:modular-simulation-I}, two given datasets $D_1$ and $D_2$, imply two different graphs $G$ and $G_{Z_1}$ respectively. $[Z_3]\rightarrow [Z_1,Z_2] \rightarrow [X_1,X_2]$ is a valid training order for $\mathcal{H}_{Z_1}$, we follow the same order as $\mathcal{H}^{\emptyset}$. We follow : $[Z_1,Z_2,Z_3] \rightarrow [X_1,X_2]$.

\begin{algorithm}[t!]
  \footnotesize
\caption{ \footnotesize TrainModule($\mathbb{G}, G, H_{*}, \mathcal{A}, \mathbf{D}$)}
    \label{alg:modular-train-DCM}
\begin{algorithmic}[1]
   \STATE {\bfseries Input:} DCM $\mathbb{G}$, Graph $G(\mathcal{V}$, $\mathcal{E})$, h-node $H_*$,
Ancestor set $\mathcal{A}$, Data $\mathbf{D}$,  Params $\theta_H, \lambda = 10$
   \WHILE{$\theta_{H_*}$ has not converged}
   \FOR{ {\bfseries each} $(\mathcal{A}_i, X_i, D_i) \in (\mathcal{A}, \mathbf{D})$} \label{line:iter-dist}
\STATE $V_r= H_* \cup \mathcal{A}_i \cup Pa(H_* \cup \mathcal{A}_i)\cup X_i$ \label{line:all-data-var}
\STATE Initialize critic $\mathbb{D}_{w_i}$
\FOR{$t=1,\hdots,m$ \COMMENT{$m$ samples}} 
\STATE Sample real data $\mathbf{v_x^r}\sim D_i$  \label{line:sample-vars}
\STATE $\mathbf{x}^r \gets$ get\_intv\_values($X_i ,D_i$) \label{line:get-intvs}
\STATE $\mathbf{v}_x^f =$ RunGAN($\mathbb{G}, \mathbf{x}^r, V_r, \theta_{H_*})$  \label{line:run-gan}
\STATE $\hat{\mathbf{v}}_x= \epsilon \mathbf{v}_x^r + (1 - \epsilon)\mathbf{v}_x^f $
\STATE $L_{i}^{(t)} = {\mathbb{D}_{w_i}} (\mathbf{v}_x^f) - \mathbb{D}_{w_i} (\mathbf{v}_x^r) 
* \lambda(\left\Vert \nabla_{\hat{\mathbf{v}}_x}\mathbb{D}_{w_i}(\hat{\mathbf{v}}_x)\right\Vert_2 - 1)^2 $
\label{line:crit-loss}
\ENDFOR
\STATE $w_i = Adam(\nabla_{w_i} \frac{1}{m} \sum_{t=1}^m L_{i}^{(t)}, w_i)$  \label{line:upd-crit}
\STATE $G_{loss} = G_{loss} + \frac{1}{m} \sum_{j=1}^m - \mathbb{D}_{w_i} (\mathbf{v}_x^f)$ \label{line:gen-loss}
\hspace{5mm} 
   \ENDFOR
 \FOR{$\theta \in \theta_{H_*}$ \COMMENT{All hnode mechanisms}}
\STATE $\theta = Adam(\nabla_{\theta} G_{loss}
,\theta)$ \label{line:upd-gen}
\ENDFOR
   \ENDWHILE
   \STATE {\bfseries Return:} $\theta_1, \hdots \theta_n$
   \end{algorithmic}
   \end{algorithm}
   
\begin{algorithm}[t!]
\footnotesize
\caption{\footnotesize IsRule2($Y,X, I=\emptyset$ (by default))}
\begin{algorithmic}[1]
   \STATE {\bfseries Input:} Variable sets $Y$ and $X$, Intervention $I$.
   \STATE {\bfseries Return:} 
   \IF{$P(Y \cup X|\doo(Pa(Y\cup X)), \doo(I))= P(Y\cup X|Pa(Y\cup X),\doo(I))$}
   \STATE {\bfseries Return:1} 
   \ELSE
   \STATE {\bfseries Return:0}
   \ENDIF
\end{algorithmic}
\label{alg:IsRule2}
\end{algorithm}
\begin{algorithm}[t!]
\small
	\caption{Construct-$\mathcal{H}^I$-graph($G,\mathcal{I}$)}
\begin{algorithmic}[1]
 \STATE {\bfseries Input:} Causal Graph $G$, Intervention set, $\mathcal{I}$
 \FOR{{\bfseries each} $I \in \mathcal{I}$}
   \STATE $\mathcal{C} \leftarrow \mathrm{get\_ccomponents}(G_{\overline{I}})$
   \STATE Construct graph $\mathcal{H}^I$ by creating nodes $H_j^I$ as  $H_j^I=C_j$,  $\forall C_j \in \mathcal{C} $
\ENDFOR
\FOR{{\bfseries each} $I \in \mathcal{I}$}
\FOR{{\bfseries each} $H_s^I,H_t^I \in \mathcal{H}^I$ such that $H_s^I\neq H_t^I$}
\IF{ 
$P(H_t^I|\doo(pa(H_t^I) \cap H_s^I)) \neq P(H_t^I|pa(H_t^I) \cap H_s^I)$
}
\STATE $\mathcal{H}^I.add(H_s^I \rightarrow H_t^I)$
\ENDIF
\ENDFOR
\STATE $\mathcal{H}^I \leftarrow$  merge($\mathcal{H}^I$, $cyc$) $\forall cyc \in Cycles(H^I)$
\ENDFOR
\FOR{{\bfseries each} $I \in \mathcal{I}$}
\FOR{{\bfseries each} $H_j^{\emptyset} \in \mathcal{H}^{\emptyset}$}
\STATE $H_j^{I}= \bigcup \limits_k H_k^I$ such that $\mathcal{V}(H_k^I) \subseteq \mathcal{V}(H_j^{\emptyset})$
\hspace{5mm} \text{[All variables in $H_k^I$ h-node is contained in $H_j^{\emptyset}$ h-node.] }
\ENDFOR
\ENDFOR
\STATE {\bfseries Return:} $\{\mathcal{H}^I: I \in \mathcal{I}\}$
\end{algorithmic}
\label{appex:construct-H-graph}
\end{algorithm}
\begin{algorithm}[t!]
\footnotesize
\caption{\footnotesize Modular Training-I($G, \mathcal{I}, \mathbf{D}$)}
\begin{algorithmic}[1]
   \STATE {\bfseries Input:} Causal Graph $G$, Intervention set $\mathcal{I}$, Dataset $\mathbf{D}$.
\STATE Initialize DCM $\mathbb{G}$
\STATE $\mathcal{H}^I\leftarrow$ \textbf{Construct-H-graphs}($G,\mathcal{I}$) \label{line:appex-const-hgraph}
\FOR{{\bfseries each} $H_k \in \mathcal{H}^\mathbf{\emptyset} $ in partial order \label{line:appex-iter-hnodes}
}
\STATE $\mathcal{A}_{\emp} \leftarrow \mathcal{V}$  
\hspace{4mm} \text{ //Initialize with all nodes}
\FOR{{\bfseries each} $S \subseteq An_{G}(H_k)$}   \label{line:appex-iter-anc-1}
{
\IF{$\mathrm{IsRule2} (H_k, {S}, \emp)=1$ \\and $|S| < |\mathcal{A}_{\emp}|$} \label{line:appex-assign-cond-1}
\STATE $\mathcal{A}_{\emp} \leftarrow  {S}  \label{line:appex-assign-A-1}
$ 
\ENDIF
}
\ENDFOR
\FOR{{\bfseries each} $I\in \mathcal{I} \cap {H}_k$}   \label{line:iter-intv}
\STATE $\mathcal{A}_{I} \leftarrow \mathcal{V}$  
\hspace{4mm} \text{ //Initialize with all nodes}
\FOR{{\bfseries each} $S \subseteq An_{G_{\overline{I}}}(H_k)$}   \label{line:appex-iter-anc-2}
{
\IF{$\mathrm{IsRule2} (H_k, {S}, I)=1$ \\and $|S| < |\mathcal{A}_{I}|$} \label{line:appex-assign-cond-2}
\STATE $\mathcal{A}_{I} \leftarrow  {S}  \label{line:appex-assign-A-2}
$ 
\ENDIF
}
\ENDFOR
\ENDFOR
\STATE $\mathbb{G}_{H_k} \leftarrow$TrainModule(${\mathbb{G}_{H_k}, G}, H_k, \mathcal{A} , \mathbf{D} $) \label{line:appex-call-trmod}
\ENDFOR
   \STATE {\bfseries Return:} $\mathbb{G}$
\end{algorithmic}
\label{alg:appex-train-by-components}
\end{algorithm}
\par
We run the subroutine \textbf{Contruct-$\mathcal{H}^I$-graph()} in Algorithm~\ref{appex:construct-H-graph} to build $\mathcal{H}$-graphs.
We check the edge condition at line~\ref{alg2:do-calculus-check} and merge cycles at line~\ref{alg:h-graph-mergecycles} if any.
In Figure~\ref{fig:modular-simulation} step (1/3), we build the $\mathcal{H}$-graph $H_1:[Z_1,Z_2,Z_3]\rightarrow H_2:[X_1,X_2]$ for $G$.

\subsubsection{Training Process of Modular \WG}
We construct the $\mathcal{H}^I$-graph for each $I\in \mathcal{I}$ at Algorithm~\ref{alg:appex-train-by-components} line~\ref{line:appex-const-hgraph}. Next, we train each h-node $H_k^{\emptyset}$ of $\mathcal{H}^{\emptyset}$, according to its partial order $\mathcal{T}$. Since we follow the partial order of $\mathcal{H}^{\emptyset}$-graph, we remove the superscript to address the hnode.
Next, we match alternative distributions for $P_I(\mathcal{V})$ c-factors that correspond to the c-components in $H_k$. (lines~\ref{line:appex-iter-hnodes}-\ref{line:appex-call-trmod}) 
%
%
%
We initialize a set 
$\mathcal{A}_I=\{V:V\subseteq An_{{G}_{\overline{I}}}(H_k)\}, \forall I\in \mathcal{I}$
to keep track of the joint distribution we need to match to train  each h-node $H_k$ from $D^I$ datasets.
We iterate over each intervention and search for the smallest set of ancestors $\mathcal{A}_I$ in $G_{\overline{I}}$  such that $\mathcal{A}_I$ satisfies the modularity condition for $H_k$ in $D^I$ dataset tested by Algorithm~\ref{alg:IsRule2}: \textbf{$\mathrm{IsRule2}(.)$} (line~\ref{line:appex-assign-cond-1})

$\mathcal{A}_I, \forall I\in \mathcal{I}$ implies a set of joint distributions in Equation~\ref{eq:appex-hnode-mod-cond}, which is sufficient for training the current h-node $H_k$ 
to learn the c-factors $P_{Pa(C_i) \cup I}(C_i), \forall C_i \in H_k, \forall I\in \mathcal{I}$.

\begin{equation}
\label{eq:appex-hnode-mod-cond}
\begin{split}
    Q(H_k \cup & \mathcal{A}_I |  \doo( pa(H_k \cup \mathcal{A}_I)), \doo({{I}}))
    = P(H_k \cup \mathcal{A}_I |  {pa(H_k \cup \mathcal{A}_I)}, \doo(I)), \text{in }{G_{\overline{I}}}, \forall I\in \mathcal{I}.\\
    & \text{Training: $H_k$, Pre-trained: $\mathcal{A}_I$, From $D^I$ dataset: $pa(H_k \cup \mathcal{A}_I)$, Intervened: $\doo(I)$}
\end{split}
\end{equation}

%
 Training $H_k$ with the  $\mathcal{A}_{\emp}$ found at this step, is sufficient to learn the c-factors $P_{Pa(C_i)}(C_i), \forall C_i \in H_k$.
Similarly, if we have an interventional dataset with $I\in H_k$ i.e., the intervened variable lies in the current h-node, we have to match c-factors $P_{Pa(C_i)\cup I}(C_i), \forall C_i \in H_k$. 
To find alternatives to these c-factors, we look for the ancestor set $\mathcal{A}_{I}$ in the $D^I$ dataset.
 For each $\mathcal{A}_{I}$, we train $H_k$ to match the interventional joint  distribution in Equation~\ref{eq:appex-hnode-mod-cond}.
We ignore intervention on any descendants of $H_k$ since the intervention will not affect c-factors differently than the c-factor in the $D^{\emp}$ observational dataset.

\subsubsection{Learn \texorpdfstring{$P(\mathcal{V})$} c-factors from Interventional Datasets}
When we need the alternative distribution for $P(\mathcal{V})$ c-factor, we search for the smallest ancestor set in $D^{\emptyset}$ dataset. However,
when we have a dataset $D^I$  with $I \in An_G(H_j^\emptyset)$,
we can search for an ancestor set in $An_{G_{\overline{I}}}(H_k)$ and train on $D^I$ to match a distribution that would be a proxy to $P(\mathcal{V})$ c-factor.
This is possible because when we factorize $P_I(\mathcal{V})$ for $I \in An_G(H_j^\emptyset)$, the c-factors corresponding to the descendant c-components of $I$ are similar to $P(\mathcal{V})$ c-factors of the same c-components.

\par
We update Theorem~\ref{th:main:modular-train-converges} for interventional datasets as below.
\begin{theorem}
\label{th:appex-intv-identifiability}
Let $\mathcal{M}_1 = (G=(\mathcal{V},\mathcal{E}), \mathcal{N}, \mathcal{U}, \mathcal{F}, P(.) )$ be the true SCM and $\mathcal{M}_2 = (G, \mathcal{N}', \mathcal{U}', \mathcal{F}', Q(.))$ be the DCM.
Suppose Algorithm \ref{alg:appex-train-by-components}: \textbf{\WG Modular Training-I} on observational and interventional datasets $\mathbf{D}^I\sim P_I(\mathcal{V}), \forall I\in \mathcal{I}$ converges
for each h-node in the $\mathcal{H}^\emptyset$-graph constructed from $G=(\mathcal{V},\mathcal{E})$
and DCM induces the distribution $Q_I(V), \forall I\in \mathcal{I}$.  Then, we have $i) P_I(V)=Q_I(V)$, and $ii)$ for any interventional
causal query $\mathcal{K}_{\mathcal{M}_1}(\mathcal{V})$ that is identifiable from $\mathbf{D}^I, \forall I\in \mathcal{I}$, we have  $\mathcal{K}_{\mathcal{M}_1}(\mathcal{V})= \mathcal{K}_{\mathcal{M}2}(\mathcal{V})$.
\end{theorem}
We provide the proof in Appendix~\ref{appex:converge-alg-3-proof}.


\subsection[Training following the H-graph]{Training following the $\mathcal{H}$-graph}

\WG utilizes conditional generative models such as diffusion models and Wasserstein GAN with penalized gradients ~\citep{gulrajani2017improved} for adversarial training on $\mathcal{L}_1$ and $\mathcal{L}_2$ datasets in Algorithm~\ref{alg:modular-train-DCM}.
$\mathbb{G}$ is the DCM, a set of generators and 
$\{\mathbb{D}_{w_i}\}$
are a set of critics for each intervention dataset.
Here, for intervention $\text{do}(X=x), X\in \mathbf{I}$, $\mathbb{G}^{(x)}({\mathbf{z,u}})$ are generated samples and $v\sim \mathbb{P}^{r}_{x}$ are real $\mathcal{L}_1$ or $\mathcal{L}_2$ samples.
 We train our models by iterating over all datasets and learn $\mathcal{L}_1$ and $\mathcal{L}_2$ distributions (lines~\ref{line:iter-dist}-\ref{line:upd-gen}). 
 We produce fake interventional samples at line~\ref{line:run-gan} by intervening on the corresponding node of our architecture with Algorithm~\ref{alg:RunGAN} RunGAN().
 Each critic $\mathbb{D}_{x}$ will obtain different losses, $L_{X=x}$ by comparing the generated samples with different true datasets (line~\ref{line:crit-loss}).
 Finally, at line~\ref{line:upd-gen}, we update each variable's model weights located at the current hnode 
 based on the accumulated generated loss over each dataset at line~\ref{line:gen-loss}. 
After calling Algorithm~\ref{alg:modular-train-DCM}: \textbf{TrainModule()} for each of the h-nodes according to the partial order of $\mathcal{H}^\emptyset$-graph, \WG will find a DCM equivalent to the true SCM that matches all dataset distributions. According to, Theorem~\ref{th:appex-identifiability}, it will be able to produce correct $\mathcal{L}_1, \mathcal{L}_2$ samples identifiable from these input distributions (Appendix~\ref{appex:wg-sampling}). \\

\subsection{Essential Theoretical Statements Required for Distributions Matching by \WG Modular Training
}

\begin{figure}[t!]
  \centering
     \includegraphics[width=1\linewidth]{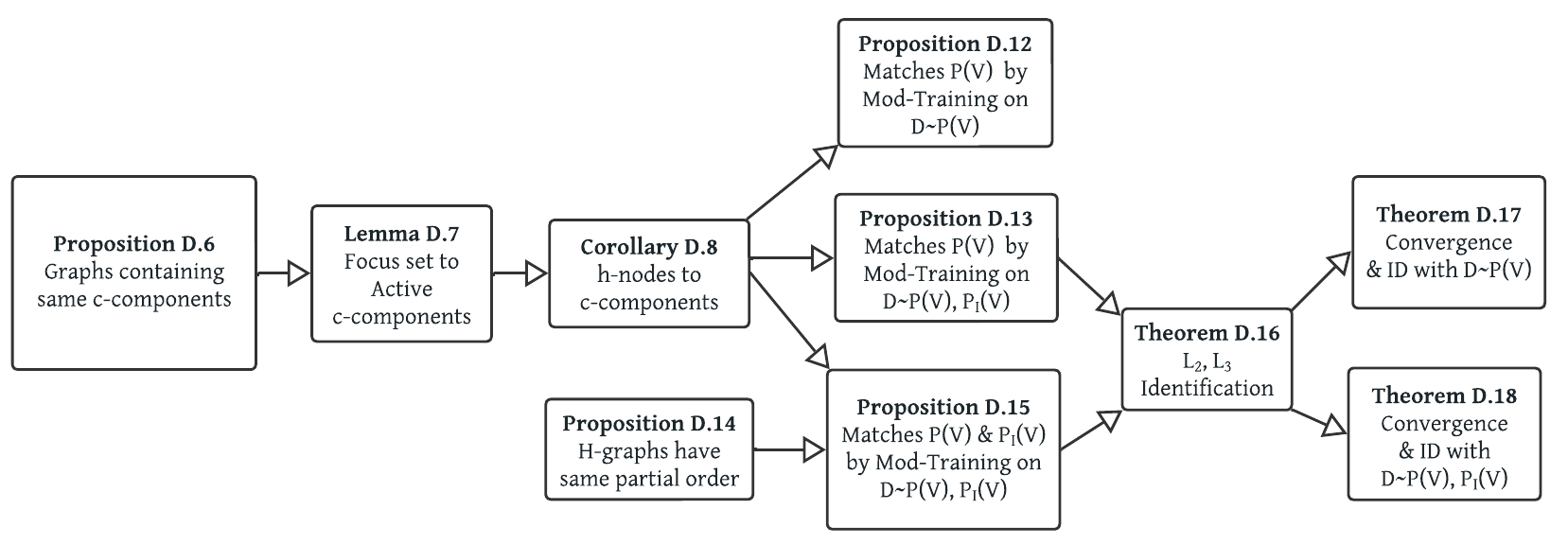}
   \caption{Flowchart of proofs }
   \label{proof-flow-init}
\end{figure}

In this section and the following section, we prove some theoretical statements required for our algorithm. Figure~\ref{proof-flow-init}, illustrates the statements we have to prove and the route we have to follow.

In proposition~\ref{prop:same-cc}, we prove the property of a sub-graph having the same set of c-components although we intervene on their parents outside that sub-graph.
We use this proposition in Lemma~\ref{thm:sub-graph-factorization} to show that we can apply c-component factorization for any sub-graph under appropriate intervention. Therefore, in Corollary~\ref{thm-set-of-c-factorization}, we can show that c-component factorization can be applied for h-nodes of the $\mathcal{H}$-graph as well. 

We build the above theoretical ground  and utilize the statements in section~\ref{appendix:theoretical_proofs}.
We show that c-factorization works for  the $\mathcal{H}$-graph and Modular Training on h-nodes matches those c-factors. Thus, \WG  will be able to match 
i) the observational joint distribution $P(\mathcal{V})$ after training on observational data (Proposition~\ref{appex-th:obs-dist-matches-obsdata}) 
ii) the observational joint distribution $P(\mathcal{V})$ after training on partial observational data and interventional data (Proposition~\ref{appex-th:obs-dist-matches}) 
 and iii) the interventional joint distribution $P_I(\mathcal{V}), \forall I \in \mathcal{I}$ after training on  observational data and interventional data. (Proposition~\ref{appex-th:intv-dist-matches}). The last proposition requires the proof that for all intervention $I \in \mathcal{I}$, the generated $\mathcal{H}^I$ graphs follows the same partial order.
 
 \WG modular training can now match $P(\mathcal{V})$ and $P_I(\mathcal{V}), \forall I \in \mathcal{I}$ according to Proposition~\ref{appex-th:obs-dist-matches} and Proposition~\ref{appex-th:intv-dist-matches}.
 We can now apply Theorem~\ref{th:appex-identifiability} to say that
  \WG modular training can sample from identifiable interventional distributions after training on $D\sim P(V)$.
  Finally, Theorem~\ref{th:modular-train-converges-obsdata} for observational case is a direct application of Proposition~\ref{appex-th:obs-dist-matches-obsdata} and Theorem\ref{appex-th:intervention-identifiability} while Theorem~\ref{th:appex-modular-train-converges} for interventional case is a direct application of Proposition~\ref{appex-th:obs-dist-matches}, Theorem~\ref{appex-th:intv-dist-matches} and Theorem~\ref{appex-th:intervention-identifiability}.

We start with some definitions that would be required during our proofs.
\begin{definition}[Intervention Set, $\mathcal{I}$]
Intervention Set, $\mathcal{I}$ represents the set of all available interventional variables such that after performing intervention $I\in \mathcal{I}$ on $G$, we observe $G_{\overline{I}}$. $\mathcal{I}$ includes $I=\emptyset$, which refers to "no intervention" and implies the original graph $G$ and the  observational data $P(\mathcal{V})$.
\end{definition}

\begin{definition}[Sub-graph, $(G_{\overline{I}})_V$]
Let $G_V$ be a sub-graph of $G$ containing nodes in $V$ and all arrows between such nodes. $(G_{\overline{I}})_V$ refers to the sub-graph of $G_{\overline{I}}$ containing nodes in $V$ only, such that variable $I$ is intervened on i.e., all incoming edges to $I$ is cut off.
\end{definition}

\begin{proposition}
\label{prop:same-cc}
Let $V\in \mathcal{V}, I\in\mathcal{I}$ be some arbitrary variable sets. 
The set of c-components formed from a sub-graph $(G_{\overline{I}})_V$ with intervention $I$ is not affected by additional interventions on their parents from outside of the sub-graph. Formally, $(G_{\overline{Pa(V)\cup I}})_V$ and $(G_{\overline{I}})_V$ has the same set of c-components. 
\end{proposition}
\begin{proof}
Let $C((G_{\overline{I}})_V)$ be the c-components which consists of nodes of $V$ in graph $(G_{\overline{I}})_V$. In sub-graph $(G_{\overline{Pa(V)\cup I}})_V$, no extra intervention is being done on any node in $V$ rather only on $Pa(V)$ where $V$ and $Pa(V)$ are two disjoint sets. Therefore, the c-components can be produced from this sub-graph will be same as for $G_{\overline{I}}$. i.e., $C(G_{V\overline{Pa(V)\cup I}}) = C(G_{\overline{I}}).$
\end{proof}
%
\begin{lemma}  
\label{thm:sub-graph-factorization}

Let $V'$ be a set called focus-set. $V'$ and intervention $I$ be arbitrary subsets of observable variables $\mathcal{V}$ and
$\{C_i\}_i$ be the set of c-components in $G_{\overline{I}}$.
Let $\{Pa(V') \cup I\}$ be a set called action-set.
and $S$ be a set called remain-set, defined as $S \coloneqq \mathcal{V}\backslash\{V'\cup Pa(V') \cup I\}$, 
$S(i)$ as $S(i)= S \cap C_i$ i.e., some part of the remain-set that are located in c-component $C_i$.
Thus,
$S= \bigcup \limits_i S(i)$. We also define active c-components $C^+_i$ as $C^+_i \coloneqq  C_i\setminus \{S(i) \cup Pa(V') \cup I\}$ i.e., the variables in focus-set that are located in c-component $C_i$. 
Given these sets, Tian's factorization can be applied to a sub-graph under proper intervention. Formally, we can factorize as below:
\begin{equation*}
    P_{Pa(V')\cup I}(V') = \prod_i P_{Pa(C_i^+)\cup I}(C_i^+)
\end{equation*}
\end{lemma}

\begin{proof}[Proof Sketch]
 Similar to  the original c-factorization formula 
 $P(\mathcal{V}) = 
 \prod_i P_{Pa(C_i)} (C_i)$,
 we can factorize as   $P_{Pa(V') \cup I}(\mathcal{V}) = 
 \prod_i P_{Pa(C_i)\cup Pa(V')\cup I} (C_i)$. 
 Next, we can marginalize out unnecessary variables $S$   located outside of  $V'$ from both sides of this expression. The left hand side of the expression is then $P_{Pa(V') \cup I}(V')$  that is what we need.  For the right hand side, we can distribute the marginalization $\sum_{S}$ among all terms and obtain $ \prod_i \sum_{S(i)} P_{Pa(C_i)\cup Pa(V')\cup I} (C_i)$  from  $\sum_{S} \prod_i P_{Pa(C_i)\cup Pa(V')\cup I} (C_i)$. Finally for each product term $P_{Pa(C_i)\cup Pa(V')\cup I} (C_i)$, we remove $S(i)$ from $C_i$ to obtain $C_i^{+}$ and drop non parent interventions following do-calculus rule-3. This final right hand side expression becomes, $\prod_i P_{Pa(C_i^+)\cup I}(C_i^+)$. We provide the detailed proof below.
\end{proof}

\begin{proof}
$(G_{\overline{Pa(V')\cup I}})_{V'}$ and $(G_{\overline{I}})_{V'}$
have the same c-components according to Proposition~\ref{prop:same-cc}.
 According to Tian's factorization for causal effect identification~\citep{tian2002general}, we know that
\begin{equation}
\begin{split}
    P_{Pa(V') \cup I}(\mathcal{V}) &= 
    \prod_i P_{Pa(C_i)\cup Pa(V')\cup I} (C_i)
    \\ 
    &\text{[let $\eta = Pa(V') \cup I$, i.e., action-set]} \\
    \implies P_{\eta}(\eta) \times P_{\eta}(\mathcal{V} \setminus \eta|\eta) &= 
    \prod_i P_{Pa(C_i) \cup \eta} (C_i) \\
    \implies  P_{\eta}(\mathcal{V} \setminus \{Pa(V') \cup I\}) &= 
    \prod_i P_{Pa(C_i) \cup \eta} (C_i) \\
    \end{split}
\end{equation}
We ignore conditioning on action-set $\eta= Pa(V')\cup I$ since we are intervening on it. Now, we have a joint distribution of focus-set and remain-set with action-set as an intervention.
\begin{equation}
\begin{split}
    \implies  P_{\eta}(V' \cup S ) &=  
    \prod_i P_{Pa(C_i) \cup \eta} (C_i) \\ 
    \text{[Here, } 
    S \coloneqq \mathcal{V}\backslash \{V'\cup & Pa(V')  \cup I\}
    \implies
    \mathcal{V} \setminus \{Pa(V') \cup I\} = V' \cup S]
    \\ 
    \implies \sum_{S} P_{\eta}(V' \cup S ) &= \sum_{S} 
    \prod_i P_{Pa(C_i) \cup \eta} (C_i) \\ 
    \implies \sum_{S} P_{\eta}(V' \cup S ) &=
    \prod_i \sum_{S(i)} P_{Pa(C_i) \cup \eta} (C_i ) \\ 
    \text{[Since, $S(i)= S \cap C_i $} & \text{and $\forall (i,j), i\neq j, C_i\cap C_j =\emptyset \implies S_i \cap S_j = \emptyset$]} \\
    \end{split}
\end{equation}
Here, $\forall_i, S(i)$ are disjoint partitions of the variable set $S$ and contained in only c-component $C_i$, i.e, $S(i)= S \cap C_i $. Since $\forall_{i,j}, C_i\cap C_j =\emptyset$, this implies that $S_i \cap S_j = \emptyset$ would occur as well. Intuitively, remain-sets located in different c-components do not intersect. Therefore, each of the probability terms at R.H.S, $P_{Pa(C_i) \cup \eta} (C_i)$ is only a function of $S(i)$ instead of whole $S$. This gives us the opportunity to push the marginalization of $S(i)$ inside the product and marginalize the probability term. After marginalizing $S(i)$ from the joint, we define rest of the variables as active c-components $C^+_i$. 
The following Figure~\ref{proof-visualization} helps to visualize all the sets.
\begin{figure}[H]
  \centering
     \includegraphics[width=0.6\linewidth]{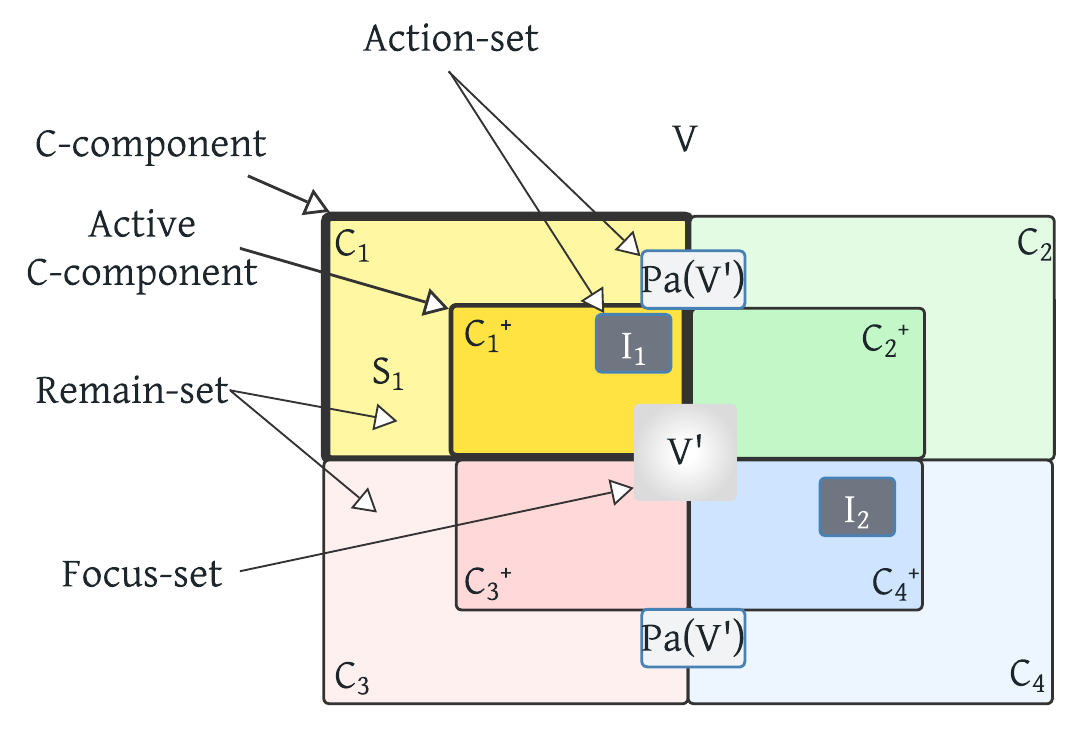}
   \caption{Visualization of focus-sets, action-sets, and remain-sets }
   \label{proof-visualization}
\end{figure}

We continue the derivation as follows:
    \begin{equation}
\begin{split}
    \implies P_{Pa(V') \cup I}(V') &=
    \prod_i P_{Pa(C_i)\cup Pa(V') \cup  I} (C_i^+)\\
     & \hspace{0mm} [Here, C_i^+ = C_i\setminus  S(i) \text{, i.e., active c-component: focus-set elements located in $C_i$}]\\
    &= \prod_i P_{Pa(C_i^+) \cup  I \cup \{Pa(C_i)\setminus Pa(C_i^+)\}  \cup Pa(V') } (C_i^+)\\
    &= 
    \prod_i P_{X \cup Z} (C_i^+)
    \hspace{3mm} \text{[Let, 
    $X= Pa(C_i^+) \cup I$ and 
    $Z = \{Pa(C_i) \cup Pa(V')\} \setminus X$]}
    \end{split}
\end{equation}
Here, we have variable set $C^+_i$ in the joint distribution. Now, if we intervene on the parents $Pa(C_i^+)$ and $I$, rest of the intervention which is outside $C^+_i$ becomes ineffective. Therefore, we have $X= Pa(C_i^+) \cup I$, the intervention which shilds the rest of the interventions, $Z = \{Pa(C_i) \cup Pa(V')\} \setminus X$. Therefore, we can apply do-calculus rule 3 on $Z$ and remove those interventions. Finally,
        \begin{equation}
\begin{split}
    \implies P_{Pa(V') \cup I}(V') &= 
    \prod_i P_{Pa(C_i^+) \cup I} (C_i^+) 
      \hspace{3mm}  \text{[We apply Rule $3$ since $C_i \indep Z | X_{G_{\overline{X}}} $]}
\end{split}
\end{equation}


%
\end{proof}

Corollary~\ref{thm-set-of-c-factorization}, suggests that Tian's factorization can be applied on the h-nodes of $H^I \in \mathcal{H}$.\\
\begin{corollary} 
\label{thm-set-of-c-factorization}
Consider a causal graph $G$. Let $\{C_i\}_{i\in [t]}$ be the c-components of $G_{\overline{I}}$. For some intervention target $I$, let
$H^I=(V_{\mathcal{H}^I},E_{\mathcal{H}^I})$ be the h-graph constructed by Algorithm \ref{appex:construct-H-graph} where
$V_{\mathcal{H}^I}= \{H^I_k\}_{k}$. Suppose $H_k^I$ is some node in $\mathcal{H}^I$. We have that $H_k^I=\{C_i\}_{i\in T_k^I}$ for some $T_k^I\subseteq [t]$. 
With slight abuse of notation we use $H_k^I$ interchangeably with the set of nodes that are in $H_k^I$.
Then,
\begin{equation}
    P_{ Pa(H_k^I) \cup I} (H_k^I) =\prod \limits_{i\in [t]} P_{Pa(C_i) \cup I }(C_i)
\end{equation}
\end{corollary}
\begin{proof}
Let, $V'=H_k^I$, $C^+_i= C_i \setminus \emptyset= C_i$. Then, this corollary is direct application of Lemma~\ref{thm:sub-graph-factorization}.
\end{proof}
\begin{figure}[H]
  \centering
     \includegraphics[width=1\linewidth]{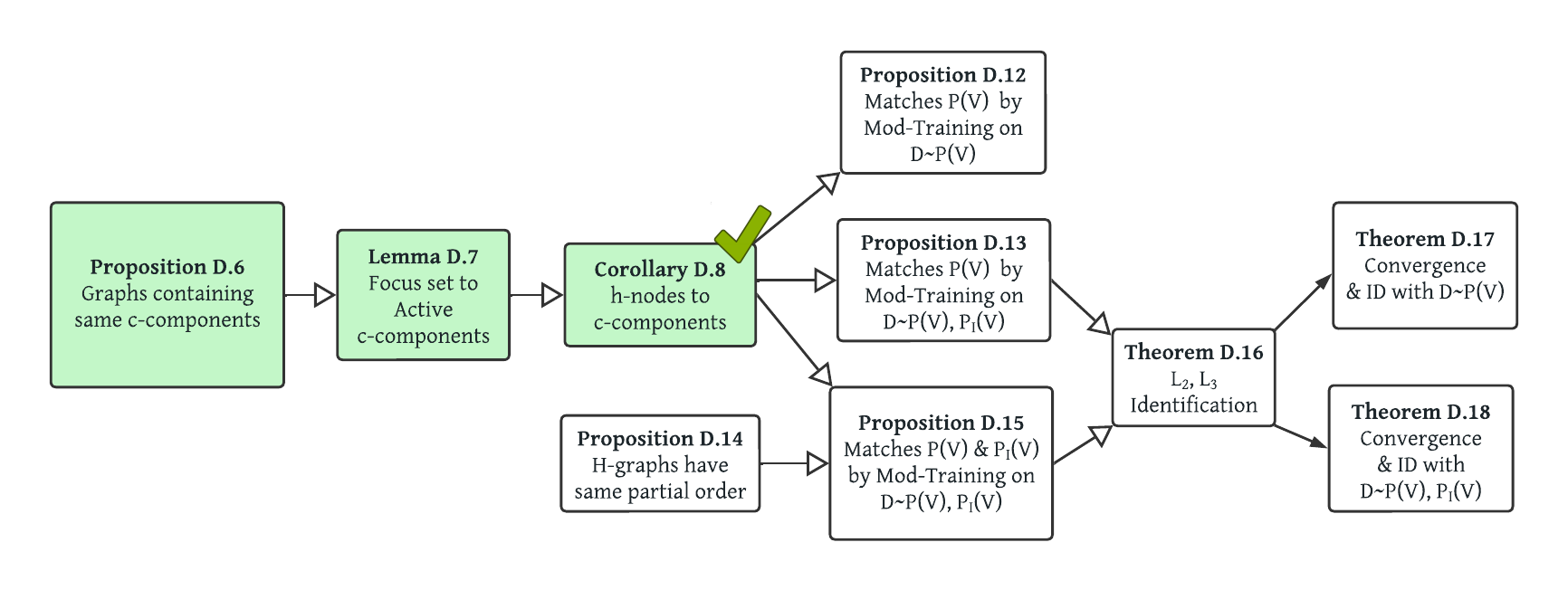}
   \caption{Flowchart of proofs }
   \label{proof-flow-D5}
\end{figure}
%
\subsection{Theoretical Proofs of Distributions Matching by \WG Modular Training
}  
\label{appendix:theoretical_proofs}
We provide Definition~\ref{def:HI-graph}: $\mathcal{H}^I$-graph here again.
\begin{definition}[$\mathcal{H}^I$-graph]
For a post-interventional graph $G_{\overline{I}}$, let the set of c-components in $G_{\overline{I}}$ be $\mathcal{C}=\{C_1,\hdots C_t\}$. Choose a partition  $\{H^I_k\}_k$ of $\mathcal{C}$ such that the $\mathcal{H}^I$-graph $\mathcal{H}^I=(V_{\mathcal{H}^I},E_{\mathcal{H}^I})$, defined as follows, is acyclic: 
$V_{\mathcal{H}^I}= \{H^I_k\}_k$ and 
for any $s,t$, $H^I_s \rightarrow H^I_t \in E_{\mathcal{H}^I}$, iff
$P(H^I_t|\doo(pa(H^I_t) \cap H^I_s)) \neq P(H^I_t|pa(H^I_t) \cap H^I_s)$, i.e., do-calculus rule-$2$ does not hold. Note that one can always choose a partition of $\mathcal{C}$ to ensure $\mathcal{H}^I$ is acylic: The $\mathcal{H}^I$ graph with a single node $H^I_1 =\mathcal{C}$ in $G_{\overline{I}}$.
Even though $\mathcal{H}^I$ for different $I$ might have different partial order, during training, every $\mathcal{H}^I$ follows the partial order of $\mathcal{H}^{\emptyset}$.
 Since its partial order is valid for other H-graphs as well.\\
\textbf{Training order, $\mathcal{T}$}:
  We define a training order, $\mathcal{T}=\{\sigma_0,\hdots, \sigma_m\}$ where $\sigma_i = \{H_k\}_k$. If $H^I_{k_1} \rightarrow H^I_{k_2}, H^I_{k_1}\in \sigma_i,   H^I_{k_2}\in \sigma_j$ then $i<j$.
 \end{definition}
%
 \begin{definition}[Notation for distributions]
$Q(.)$ is the observational distribution induced by the deep causal SCM.
$P(.)$ is the true (observational/interventional) distribution.
With a slight abuse of notation, if we have $P(\mathbf{V})$ and intervention $I$, then $P_I(\mathbf{V})$ indicates $P_I(\mathbf{V}\setminus I)$.
Algorithm~\ref{alg:appex-train-by-components} is said to have converged if training attains zero loss every time line~\ref{line:appex-call-trmod} is visited. 
 \end{definition}
 
 \begin{definition}[Ancestor set $\mathcal{A}_{I}$ in $G_{\overline{I}}$]
 Let parents of a variable set $\mathbf{V}$ be $Pa(\mathbf{V})= \bigcup\limits_{V \in \mathbf{V}} Pa(V) \setminus \mathbf{V}$.
 Now, for some h-node $H_k^I$ $\in \mathcal{H}^I$-graph, we define $\mathcal{A}_{I}\coloneqq$  the minimal subset of ancestors exists in the causal graph $G_I$ with intervention $I$  such that the following holds,
\begin{equation}
\label{eq:obs-anc}
p(H_n^I\cup \mathcal{A}_{I}| \doo(pa(H_n^I\cup \mathcal{A}_{I})), \doo(I)
=
p(H_n^I\cup \mathcal{A}_{I}|pa(H_n^I\cup \mathcal{A}_{I}), \doo(I))
\end{equation}
\end{definition}


For training any h-node in the training order $\mathcal{T}=\{\sigma_0,\hdots, \sigma_m\}$, i.e., $H_k^{\emptyset} \in \sigma_j$, $0 < j \leq m$, if only observational data is available, i.e., $I=\emptyset$, we search for an ancestor set $\mathcal{A}_\emptyset$ such that $\mathcal{A}_\emptyset$  satisfies modularity condition for $H_k^{\emptyset}$:
\begin{equation}
P(H_k^\emptyset\cup \mathcal{A}_{\emptyset}| \doo(pa(H_k^\emptyset \cup \mathcal{A}_{\emptyset}))
=
P(H_k^\emptyset \cup \mathcal{A}_{\emptyset}|pa(H_k^\emptyset \cup \mathcal{A}_{\emptyset}))
\end{equation}
Similarly, for $I\in An_G(H_k^I)$, i.e.,
intervention on ancestors,
we can learn $P_{pa(H^{\emptyset}_k)}(H^{\emptyset}_k)$ from available interventional datasets since $H^I_k= H^{\emptyset}_k$, i.e., contains the same c-factors, according to $\mathcal{H}^I$-graphs construction.
These c-factor distributions are identifiable from $P_I(V)$ as they can be calculated from the c-factorization of $P_I(V)$. Thus we have,
\begin{equation}
\label{eq:obs-id-from-itv}
   \begin{split}
        P_{pa(H^{I}_k)}(H^{I}_k)&= P_{pa(H^{\emptyset}_k)}(H^{\emptyset}_k)
    \end{split}
\end{equation}
Therefore, to utilize ancestor interventional datasets,
We search for smallest ancestor set $\mathcal{A}_I \subseteq An_{G_{\overline{I}}}(H_k^I)$ in $G_{\overline{I}}$ such that do-calculus rule-2 applies,
\begin{equation}
P(H_k^I\cup \mathcal{A}_{I}| \doo(pa(H_k^I\cup \mathcal{A}_{I})), \doo(I))=
P(H_k^I\cup \mathcal{A}_{I}|pa(H_k^I\cup \mathcal{A}_{I}), \doo(I))
\end{equation}
Then we can train the mechanisms in $H^{\emptyset}_k$ to learn the $P(\mathcal{V})$ c-factors by matching the following alternative distribution from $D^I$ dataset, 
\begin{equation}
\label{eq:obs-dist-match-with-intv}
\begin{split}
    P(H_k^I\cup \mathcal{A}_{I}|pa(H_k^I\cup \mathcal{A}_{I}), \doo(I))&=
    Q(H_k^I\cup \mathcal{A}_{I}|\doo(pa(H_k^I\cup \mathcal{A}_{I})), \doo(I))\\
    \implies P(H_k^I\cup \mathcal{A}_{I}| \doo(pa(H_k^I\cup \mathcal{A}_{I})), \doo(I)&=
    Q(H_k^I\cup \mathcal{A}_{I}|\doo(pa(H_k^I\cup \mathcal{A}_{I})), \doo(I))\\
\end{split}
\end{equation}
Matching the alternative distributions with $D^I$ will imply that we match $P(\mathcal{V})$ c-factor as well. Formally:
\begin{equation}
   \begin{split}
       Q_{pa(H^{I}_k)}(H^{I}_k) &= P_{pa(H^{I}_k)}(H^{I}_k) \\
      Q_{pa(H^{I}_k)}(H^{I}_k) &=P_{pa(H^{\emptyset}_k)}(H^{\emptyset}_k)
    \end{split}
\end{equation}


\subsubsection[Matching Observational Distributions with Modular Training on observational dataset]{Matching Observational Distributions with Modular Training on $D\sim P(\mathcal{V})$ }
Now, we provide the theoretical proof of the correctness of \WG Modular Training matching observational distribution by training on observational dataset $D^{\emp}.$
Since, we have access to only observational data we remove the intervention-indicating superscript/subscript and address $\mathcal{H}^{\emp}$ as $\mathcal{H}$, ancestor set $\mathcal{A}_I$ as $\mathcal{A}$ and dataset $D^{\emp}$ as $D$.
\begin{proposition}
\label{appex-th:obs-dist-matches-obsdata}
Suppose Algorithm \ref{alg:train-by-components}: \textbf{\WG Modular Training} converges
for each h-node in $\mathcal{H}^{\emptyset}$-graph constructed from $G=(\mathcal{V},\mathcal{E})$. Suppose the observational distribution induced by the deep causal model is $Q(\mathcal{V})$ after training on data sets $D\sim P(\mathcal{V})$.  Then,
\begin{equation}
    P(\mathcal{V})=Q(\mathcal{V})
\end{equation}
\end{proposition}

\begin{proof}[Proof Sketch]
	After expressing the observational distribution $P(V)$ as c-factorization expression, we can combine multiple c-factors located in the same h-node as 
	$
	\prod \limits_{C_i\in H^{}_k} 
	P_{pa(C_{i})}(C_{i}) 
	= P_{pa(H_k^{})}(H_k^{}) $
	 according to  Corollary \ref{thm-set-of-c-factorization}. Therefore, if we can match $P_{pa(H_k^{})}(H_k^{}), \forall k$, this will ensure that we have matched all c-factors $	P_{pa(C_{i})}(C_{i}), \forall i $ and as a result $P(V)$ as well.
	For the h-nodes which does not have any parents (i.e., root nodes) in the $\mathcal{H}$-graph,
	we know $P_{Pa(H_k^{})}(H_k^{}) = P(H_k^{}|Pa(H_k^{}))$ due to the construction of H-graph. Therefore, 	 \WG trains mechanism in those h-nodes by matching  $P(H_k^{}|Pa(H_k^{})) = Q_{Pa(H_k^{})}(H_k^{})$.

		For the h-nodes which are not root h-nodes in the $\mathcal{H}$-graph, we match $P_{pa(H_k^{})}(H_k^{})$ by matching an alternative distribution $P(H_k\cup \mathcal{A}_{}|\doo(Pa(H_k\cup \mathcal{A}_{})))$. This alternative distribution factorizes as $P(H_k\cup \mathcal{A}_{}|\doo(Pa(H_k\cup \mathcal{A}_{})))=P_{pa(H_k)}(H_k)* 
		\prod \limits_{H_S \in \mathcal{A}}  \prod\limits_{C_j^+\subseteq H^{}_S} P_{Pa(C^+_j)}(C^+_j) $.
		Therefore,
		$P_{Pa(H^{}_k)}(H^{}_k) = \frac{ P(H_k^{}\cup \mathcal{A}_{}|\doo(Pa(H_k^{}\cup \mathcal{A}_{})))}
		{   \prod \limits_{H_S \in \mathcal{A}}  \prod\limits_{C_j^+\subseteq H^{}_S} P_{Pa(C^+_j)  }(C^+_j)}$.
		The numerator is already matched as that is the alternative distribution we  have matched while training h-node $H_k$.  As we are following the topological order of the $\mathcal{H}$-graph, the denominator is matched while training the ancestor h-nodes of $H_k$. Therefore, \WG modular training matches   $P_{pa(H_k^{})}(H_k^{}), \forall k$ and thus $P(V)= Q(V)$. 
		We provide the detailed proof below.
\end{proof}
\begin{proof}

According to Tian's factorization we can factorize the joint distributions into c-factors as follows:
\begin{equation}
    P(\mathcal{V})= P(\mathcal{H}) =  
    \prod \limits_{H_k\in \mathcal{H}}
    \prod \limits_{C_i\in H_k} 
    P_{pa(C_{i})}(C_{i})
\end{equation}
We can divide the set of c-components $\mathcal{C}=\{C_1,\hdots C_t\}$
into disjoint partitions or h-nodes as $H_k^{}=\{C_i\}_{i\in T_k}$ for some $T_k\subseteq [t]$.
Following Corollary \ref{thm-set-of-c-factorization},
we can combine the c-factors in each partitions and rewrite it as:
\begin{equation}
\begin{split}
\label{eq:joint-to-hnode-obsdata}
 \prod \limits_{H_k\in \mathcal{H}^{}}
    \prod \limits_{C_i\in H^{}_k} 
    P_{pa(C_{i})}(C_{i}) 
    = P_{pa(H_0^{})}(H_0^{}) 
    \times P_{pa(H_1^{})}(H_1^{})\times \hdots 
    \times P_{pa(H_n^{})}(H_n^{})
\end{split}    
\end{equation}
%
%
Now, we prove that we match each of these terms according to the training order $\mathcal{T}$.
\par
\textbf{\underline{For any root h-nodes $H_k^{} \in \sigma_0$ }
:}\\
Due to the construction of $\mathcal{H}^{}$ graphs in Algorithm~\ref{appex:construct-H-graph}, the following is true for any root nodes, $H_k^{} \in \sigma_0$.
\begin{equation}
\label{eq:root-intv-cond-obs}
    P(H_k^{}|Pa(H_k^{})) = P_{Pa(H_k^{})}(H_k^{}) 
\end{equation}
\WG training convergence for the DCM in $H_k^{} \in \sigma_0$. (Algorithm~\ref{alg:train-by-components}, line~\ref{line:call-trmod})
ensures that the following matches:
\begin{equation}
\label{appex:eq-root-h-node-1}
\begin{split}
       P(H_k^{}|Pa(H_k^{})) &= Q_{Pa(H_k^{})}(H_k^{})\\
   \implies P_{Pa(H_k^{})}(H_k^{})&=  Q_{Pa(H_k^{})}(H_k^{})
\end{split}
\end{equation}

Since, Equation~\ref{eq:root-intv-cond-obs} is true, observational data is sufficient for training the mechanisms in $H_k^{} \in \sigma_0$. Thus, we do not need to train on interventional data.  

\textbf{\underline{For the h-node $H_k^{} \in \sigma_1$} }:\\
Now we show that  we can train mechanisms in $H_k^{}$ by matching $P(\mathcal{V})$ c-factors  
with $D^{}\sim P(\mathcal{V})$ data set.
Let us assume, $\exists \mathcal{A} \subseteq \sigma_0$ such that $\mathcal{A} = An(H_k)$,i.e., ancestors set of $H_k$ in the $\mathcal{H}$-graph that we have already trained with available $D$ dataset.
%
%
To apply Lemma \ref{thm:sub-graph-factorization} in causal graph $G$, consider $V'= H_k\cup \mathcal{A}_{}$ as the focus-set, $Pa(V')$ as the action-set. Thus, active c-components: 
$C^+_j \coloneqq C_j\cap V'$

Then we get the following:

\begin{equation}
    \begin{split}
        P(H_k\cup \mathcal{A}|\doo(Pa(H_k\cup \mathcal{A}_{}))) 
        &=\prod \limits_{C_i\in H^{}_{k} } P_{pa(C_i)}(C_i) \times \prod \limits_{H_S \in \{\mathcal{A}\}} \prod \limits_{C_j^+\subseteq H^{}_S} P_{Pa(C_{j}^+) }(C_{j}^+) 
        \\
        &\text{[{Here, 1st term is the factorization of the current h-node}} \\ 
        &\text{and 2nd term is the factorization of the ancestors set.]}
        \\
     \implies  P(H_k\cup \mathcal{A}_{}|\doo(Pa(H_k\cup \mathcal{A}_{})))   &=P_{pa(H_k)}(H_k)* 
       \prod \limits_{H_S \in \mathcal{A}}  \prod\limits_{C_j^+\subseteq H^{}_S} P_{Pa(C^+_j)}(C^+_j) 
          \end{split}
\end{equation}
Here according to Corollary~\ref{thm-set-of-c-factorization}, we combine the c-factors $P_{pa(C_i)}(C_i)$ for c-components in $H_k$  to form $P_{pa(H_k)}(H_k)$.
We continue the derivation as follows:
         \begin{equation}
         \label{eq:obs-joint-match-h1-obsdata}
    \begin{split}
        \implies P_{Pa(H^{}_k)}(H^{}_k) &= \frac{ P(H_k^{}\cup \mathcal{A}_{}|\doo(Pa(H_k^{}\cup \mathcal{A}_{})))}
        {   \prod \limits_{H_S \in \mathcal{A}}  \prod\limits_{C_j^+\subseteq H^{}_S} P_{Pa(C^+_j)  }(C^+_j)}\\
        \implies P_{Pa(H^{}_k)}(H^{}_k) &= \frac{ Q(H_k^{}\cup \mathcal{A}_{}|\doo(Pa(H_k^{}\cup \mathcal{A}_{})))}
        {
         \prod \limits_{H_S \in \mathcal{A}}  \prod\limits_{C_j^+\subseteq H^{}_S}
         Q_{Pa(C^+_j) }(C^+_j)}
           \end{split}
\end{equation}
Here the R.H.S numerator follows from previous line according to Equation~\ref{eq:obs-dist-match-with-intv}. 
For the denominator at R.H.S,  $\forall H_S \in  \mathcal{A}$, we have already matched $P(H_S\cup \mathcal{A}_{}| \doo(pa(H_S\cup \mathcal{A}_{})))$,   during training of $\mathcal{A}= An(H_k)$ h-nodes.
According to Lemma~\ref{thm:sub-graph-factorization}, matching these distribution is sufficient to match the distribution at R.H.S denominator.
Therefore,  our DCM will produce the same distribution as well. This implies that from Equation~\ref{eq:obs-joint-match-h1-obsdata} we get,
\begin{equation}
   \begin{split}
        P_{pa(H^{}_k)}(H^{}_k)&= Q_{pa(H^{}_k)}(H^{}_k)\\
        \implies P_{pa(H^{}_k)}(H^{}_k)&= Q_{pa(H^{}_k)}(H^{}_k) \hspace{5mm} \text{[According to Equation~\ref{eq:obs-id-from-itv}]}
    \end{split}
\end{equation}

Similarly, we train each h-node following the training order $\mathcal{T}$ and match the distribution in Equation~\ref{eq:joint-to-hnode-obsdata}.
This finally shows that,
\begin{equation}
\begin{split}
    P(V)= \prod \limits_{j \leq n} P_{pa(H_j^{})}(H_j^{})=
    \prod \limits_{j \leq n} Q_{pa(H_j^{})}(H_j^{})= Q(V)
\end{split}    
\end{equation}
\end{proof}

\subsubsection[Matching Observational Distributions with Modular Training on interventional dataset]{Matching Observational Distributions with Modular Training on $\mathbf{D}\sim P_I(\mathcal{V}), \forall I\in \mathcal{I}$ }

Now, we provide the theoretical proof of the correctness of \WG Modular Training matching observational distribution from multiple datasets $D^I, \forall I\in \mathcal{I}$.

\emph{\textbf{Notations:}}
When we consider multiple interventions $I \in \mathcal{I}$, we add $I$ as subscript to each notation to indicate the intervention that notation correspond to.
The following notations are mainly used  in Proposition~\ref{appex-th:obs-dist-matches}, Proposition~\ref{appex-lemma:consistent-hgraph} and Proposition~\ref{appex-th:intv-dist-matches}.
\begin{itemize}
	\item $\mathbf{D}^I$: the interventional dataset collected with intervention on node $I$.
	\item $P_{I}(V), Q_{I}(V)$: the interventional joint distribution after intervening on node $I$ representing respectively the real data and the  generated data produced from the \WG DCM.
	\item $G_{\bar{I}}$ and $\mathcal{H}^I$: the causal graph and the H-graph after $\doo(I)$ intervention.  $\mathcal{H}^{\emptyset}$ or only $\mathcal{H}$  implies the H-graph for the original causal graph. 
	\item ${H}^I_k$: the $k$-th hnode in the $\mathcal{H}^I$-graph.
	\item $\mathcal{A}_{I}$: the ancestor set in $G_{\bar{I}}$-graph, required to construct the alternative distribution for the c-component in consideration. Thus, $\mathcal{A}_{\emptyset}$ or only $\mathcal{A}$ refers to the observational case.
	\item $\sigma_0, \sigma_1$: the root hnodes and the non-root hnodes in  the H-graph in consideration.
\end{itemize}

\begin{proposition}
\label{appex-th:obs-dist-matches}
Suppose Algorithm \ref{alg:appex-train-by-components}: \textbf{\WG Modular Training} converges
for each h-node in $\mathcal{H}^{\emptyset}$-graph constructed from $G=(\mathcal{V},\mathcal{E})$. Suppose the observational distribution induced by the deep causal model is $Q(\mathcal{V})$ after training on data sets $\mathbf{D}^I, \forall I\in \mathcal{I}$.  Then,
\begin{equation}
    P(\mathcal{V})=Q(\mathcal{V})
\end{equation}
\end{proposition}

\begin{proof}[Proof Sketch]
	The proof of this Proposition follows the same route as Proposition~\ref{appex-th:obs-dist-matches-obsdata}. In both cases, \WG matches the observational distribution $P(V)$. The only difference between the two setups is that \WG has access to multiple interventional datasets in this setup which enables matching observational distribution efficiently by utilizing a smaller ancestor set with the joint.
    An important fact is that even if we have access to $\doo(I), \forall I\in \mathcal{I}$ datasets and we construct multiple $\mathcal{H}^{I}$-graphs, we still follow the topological order of $\mathcal{H}^{\emptyset}$-graph, i.e, H-graph with no intervention. This is valid according to Proposition~\ref{appex-lemma:consistent-hgraph} since a topological order of  $\mathcal{H}^{\emptyset}$ works for all  $\mathcal{H}^I$-graphs even though  $\mathcal{H}^I$ are sparser. Also, any node in $H^I_k$ contains the same set of nodes as in $H^{\emptyset}_k$ for all $k$.

We can consider any h-node to be either a root h-node or a non-root h-node.
	Since for the root h-nodes, the ancestor set is empty, we follow the same approach as the observational case and the proof of correctness follows from Proposition~\ref{appex-th:obs-dist-matches-obsdata}. 
 Now,  suppose a h-node is not a root node and intervention $I$ is not located inside it. To match the alternative distribution, instead of searching for the ancestor set in only  $H^{\emptyset}$-graph created for observational data, \WG looks at all  $H^{I}$-graphs created based on intervention $I$ and chooses the smallest ancestor set. We assume that a smaller ancestor set will make it easy to match the corresponding alternative distribution.
 	More precisely, 
	instead of matching $P(H_k\cup \mathcal{A}_{\emptyset}|\doo(Pa(H_k\cup \mathcal{A}_{\emptyset})))$ from observational $H$-graph, \WG matches  $P(H_k\cup \mathcal{A}_{I}|\doo(Pa(H_k\cup \mathcal{A}_{I})), \doo(I))$ where $A_I$ is smallest across all $H^I$-graphs.
 Since intervention $I$ is not located inside the h-node, $P(H_k\cup \mathcal{A}_{I}|\doo(Pa(H_k\cup \mathcal{A}_{I})), \doo(I)) = P(H_k\cup \mathcal{A}_{I}|Pa(H_k\cup \mathcal{A}_{I}), I)$, i.e, the interventional alternative distribution is same as an observational conditional distribution. Thus, we train on the interventional dataset to match $P_{pa(H^{I}_k)}(H^{I}_k)$ which is equivalent to $P_{pa(H^{\emptyset}_k)}(H^{\emptyset}_k)$ for h-nodes that contain no intervention. 
 
 Following the above approach, to match $P_{pa(H^{\emptyset}_k)}(H^{\emptyset}_k)$ for all h-nodes will eventually match $P(V)$.
	The rest follows the same proof as Proposition~\ref{appex-th:obs-dist-matches-obsdata}.
	We provide the detailed proof below. 
\end{proof}

\begin{proof}

According to Tian's factorization we can factorize the joint distributions into c-factors as follows:
\begin{equation}
    P(\mathcal{V})= P(H^{\emptyset}) =  
    \prod \limits_{H_k^\emptyset\in H^{\emptyset}}
    \prod \limits_{C_i\in H^{\emptyset}_k} 
    P_{pa(C_{i})}(C_{i})
\end{equation}
We can divide the set of c-components $\mathcal{C}=\{C_1,\hdots C_t\}$
into disjoint partitions or h-nodes as $H_k^{\emp}=\{C_i\}_{i\in T_k}$ for some $T_k\subseteq [t]$.
Following Corollary \ref{thm-set-of-c-factorization},
we can combine the c-factors in each partitions and rewrite it as:
\begin{equation}
\begin{split}
\label{eq:joint-to-hnode}
 \prod \limits_{H_k^\emptyset\in H^{\emptyset}}
    \prod \limits_{C_i\in H^{\emptyset}_k} 
    P_{pa(C_{i})}(C_{i}) 
    = P_{pa(H_0^{\emptyset})}(H_0^{\emptyset}) 
    \times P_{pa(H_1^{\emptyset})}(H_1^{\emptyset})\times \hdots 
    \times P_{pa(H_n^{\emptyset})}(H_n^{\emptyset})
\end{split}    
\end{equation}
%
%
Now, we prove that we match each of these terms according to the training order $\mathcal{T}$.
\par
\textbf{\underline{For any root h-nodes $H_k^{\emptyset} \in \sigma_0$ }
:}\\
Due to the construction of $\mathcal{H}^{\emptyset}$ graphs in Algorithm~\ref{appex:construct-H-graph}, the following is true for any root nodes, $H_k^{\emptyset} \in \sigma_0$.
\begin{equation}
\label{eq:root-intv-cond}
    P(H_k^{\emptyset}|Pa(H_k^{\emptyset})) = P_{Pa(H_k^{\emptyset})}(H_k^{\emptyset}) 
\end{equation}
\WG training convergence for the DCM in $H_k^{\emptyset} \in \sigma_0$. (Algorithm~\ref{alg:appex-train-by-components}, line~\ref{line:appex-call-trmod})
ensures that the following matches:
\begin{equation}
\label{appex:eq-root-h-node-2}
\begin{split}
       P(H_k^{\emptyset}|Pa(H_k^{\emptyset})) &= Q_{Pa(H_k^{\emptyset})}(H_k^{\emptyset})\\
   \implies P_{Pa(H_k^{\emptyset})}(H_k^{\emptyset})&=  Q_{Pa(H_k^{\emptyset})}(H_k^{\emptyset})
\end{split}
\end{equation}

Since, Equation~\ref{eq:root-intv-cond} is true, observational data is sufficient for training the mechanisms in $H_k^{\emptyset} \in \sigma_0$. Thus, we do not need to train on interventional data.  

\textbf{\underline{For the h-node $H_k^{\emptyset} \in \sigma_1$} }:\\
Now we show that  we can train mechanisms in $H_k^{\emptyset}$ by matching $P(\mathcal{V})$ c-factors  
with either \Li or \Lii datasets.
Let us assume, $\exists \mathcal{A}_I \subseteq \sigma_0$ such that $\mathcal{A}_I = An_{G_{\overline{I}}}(H_k^{I})$,i.e., ancestors set of $H_k^{I}$ in the $\mathcal{H}^{I}$-graph that we have already trained with available $D^I$ dataset.
%
%
To apply Lemma \ref{thm:sub-graph-factorization} in $G_{\overline{I}}$ with $|I|\geq 0$, consider $V'= H_k^I\cup \mathcal{A}_{I}$ as the focus-set, $\{Pa(V') \cup I\}$ as the action-set. Thus, active c-components: 
$C^+_j \coloneqq C_j\cap V'$

Then we get the following:

\begin{equation}
    \begin{split}
        P(H_k^I\cup \mathcal{A}_{I}|\doo(Pa(H_k^I\cup \mathcal{A}_{I})), \doo(I)) 
        &=\prod \limits_{C_i\in H^{I}_{k} } P_{pa(C_i)}(C_i) \times \prod \limits_{H_S^I \in \{\mathcal{A}_I\}} \prod \limits_{C_j^+\subseteq H^{I}_S} P_{Pa(C_{j}^+) \cup I}(C_{j}^+) 
        \\
        &\text{[{Here, 1st term is the factorization of the current h-node}} \\ 
        &\text{and 2nd term is the factorization of the ancestors set.]}
        \\
     \implies  P(H_k^I\cup \mathcal{A}_{I}|\doo(Pa(H_k^I\cup \mathcal{A}_{I})), \doo(I))   &=P_{pa(H^I_k)}(H^I_k)* 
       \prod \limits_{H_S^I \in \mathcal{A}_I}  \prod\limits_{C_j^+\subseteq H^{I}_S} P_{Pa(C^+_j) \cup I}(C^+_j) 
          \end{split}
\end{equation}
Here according to Corollary~\ref{thm-set-of-c-factorization}, we combine the c-factors $P_{pa(C_i)}(C_i)$ for c-components in $H^I_k$  to form $P_{pa(H^I_k)}(H^I_k)$.
We continue the derivation as follows:
         \begin{equation}
         \label{eq:obs-joint-match-h1}
    \begin{split}
        \implies P_{Pa(H^{I}_k)}(H^{I}_k) &= \frac{ P(H_k^{I}\cup \mathcal{A}_{I}|\doo(Pa(H_k^{I}\cup \mathcal{A}_{I})), \doo(I))}
        {   \prod \limits_{H_S^I \in \mathcal{A}_I}  \prod\limits_{C_j^+\subseteq H^{I}_S} P_{Pa(C^+_j)  \cup I}(C^+_j)}\\
        \implies P_{Pa(H^{I}_k)}(H^{I}_k) &= \frac{ Q(H_k^{I}\cup \mathcal{A}_{I}|\doo(Pa(H_k^{I}\cup \mathcal{A}_{I})), \doo(I))}
        {
         \prod \limits_{H_S^I \in \mathcal{A}_I}  \prod\limits_{C_j^+\subseteq H^{I}_S}
         Q_{Pa(C^+_j)  \cup I}(C^+_j)}
           \end{split}
\end{equation}
Here the R.H.S numerator follows from previous line according to Equation~\ref{eq:obs-dist-match-with-intv}. 
For the denominator at R.H.S, the intervention is an ancestor of the current hnode, i.e., $I\in \{An(H_k^I)\setminus H_k^I\}$.
Now, $\forall H_S^I \in  \mathcal{A}_I$, we have already matched $P(H_S^I\cup \mathcal{A}_{I}| \doo(pa(H_S^I\cup \mathcal{A}_{I})), \doo(I))$,   during training of $\mathcal{A}_I= An(H_k^I)$ h-nodes.
According to Lemma~\ref{thm:sub-graph-factorization}, matching these distribution is sufficient to match the distribution at R.H.S denominator.
Therefore,  our DCM will produce the same distribution as well. This implies that from Equation~\ref{eq:obs-joint-match-h1} we get,
\begin{equation}
   \begin{split}
        P_{pa(H^{I}_k)}(H^{I}_k)&= Q_{pa(H^{I}_k)}(H^{I}_k)\\
        \implies P_{pa(H^{\emptyset}_k)}(H^{\emptyset}_k)&= Q_{pa(H^{\emptyset}_k)}(H^{\emptyset}_k) \hspace{5mm} \text{[According to Equation~\ref{eq:obs-id-from-itv}]}
    \end{split}
\end{equation}

Similarly, we train each h-node following the training order $\mathcal{T}$ and match the distribution in Equation~\ref{eq:joint-to-hnode}.
This finally shows that,
\begin{equation}
\begin{split}
    P(V)= \prod \limits_{j \leq n} P_{pa(H_j^{\emptyset})}(H_j^{\emptyset})=
    \prod \limits_{j \leq n} Q_{pa(H_j^{\emptyset})}(H_j^{\emptyset})= Q(V)
\end{split}    
\end{equation}
\end{proof}

\begin{figure}[H]
  \centering
     \includegraphics[width=1\linewidth]{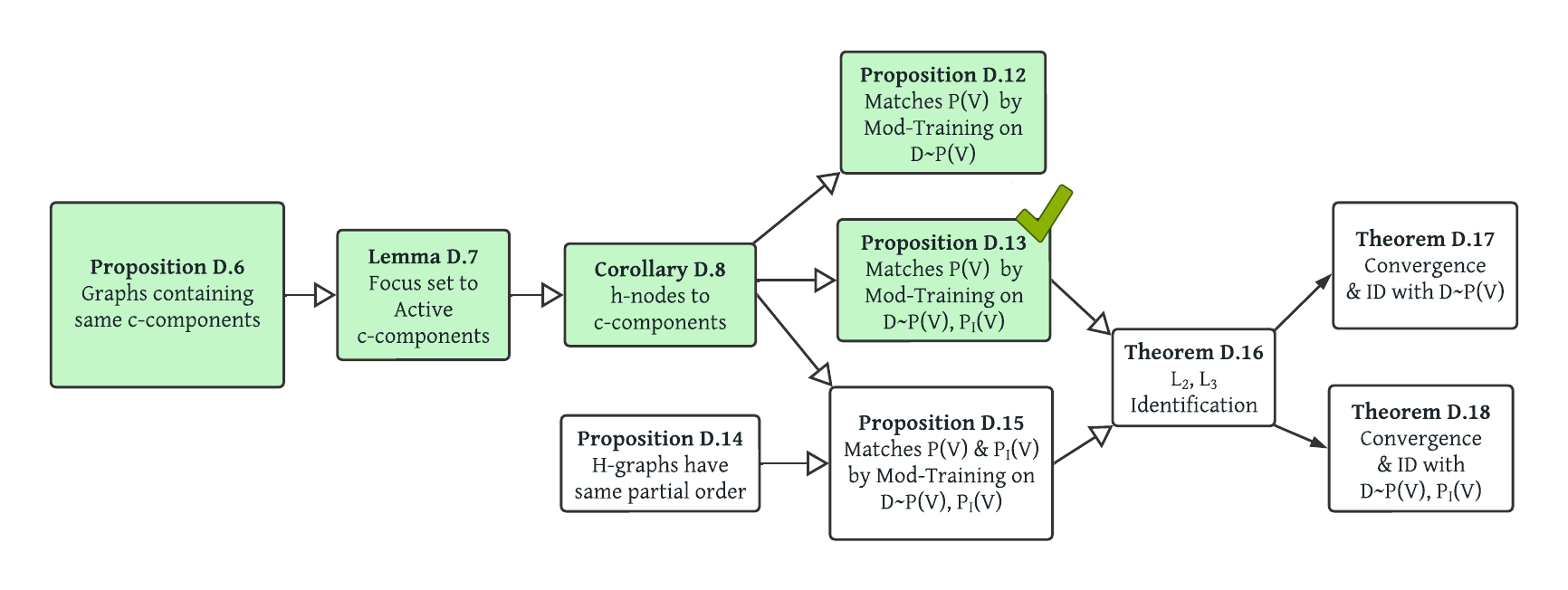}
   \caption{Flowchart of proofs }
   \label{proof-flow-D13}
\end{figure}

\subsubsection[Matching Interventional Distributions with Modular Training]{Matching Interventional Distributions with Modular Training on $\mathbf{D}\sim P_I(\mathcal{V}), \forall I\in \mathcal{I}$ }


Before showing our algorithm correctness with both observational and interventional data, we first discuss the DAG property of $\mathcal{H}$-graphs. Please check  the \textbf{notations} in the previous section defined for multiple interventions.

 \begin{proposition}
 \label{appex-lemma:consistent-hgraph}

Any $\mathcal{H}$-graph constructed according to Definition~\ref{def:HI-graph} is a directed acyclic graph (DAG) and a common partial order $\mathcal{T}$, exists for all $\mathcal{H}^I$-graphs, $\forall I\in \mathcal{I}$.



 \end{proposition}
 \begin{proof}
 We construct the $\mathcal{H}$-graphs following Algorithm~\ref{appex:construct-H-graph}. By checking the modularity condition we add edges between any two h-nodes. However, 
 if we find a cycle $H^I_j, H^I_{k} \rightarrow \hdots \rightarrow H^I_j$, then we combine all h-nodes in the cycle and form a new h-node in $\mathcal{H}^I$.
 This new h-node contains the union of all outgoing edges to other h-nodes.
 Therefore, at the end of the algorithm, the final $\mathcal{H}$-graph, $\mathcal{H}^I$ will always be a directed acyclic graph. 
 Note that one can always choose a partition of the c-components $\mathcal{C}$ to ensure $\mathcal{H}^I$ is acylic: The $\mathcal{H}^I$ graph with a single node $H_1^I =\mathcal{C}$.
 \par
 Next in Algorithm~\ref{alg:appex-train-by-components},
 training is performed  according to the partial order of $\mathcal{H}^{\emptyset}$ which corresponds to the original graph $G$ without any intervention. This is the most dense $\mathcal{H}$-graph and thus imposes the most restrictions in terms of the training order. 
 Let $I$ be an intervention set. 
 For any intervention $I$, suppose $\mathcal{H}^I$-graph is obtained from $G_{\overline{I}}$ and $I$ is located in $H^{\emptyset}_k$ h-node of $\mathcal{H}^{\emp}$-graph obtained from $G$. 
 The only difference between $\mathcal{H}^{\emptyset}$ and $\mathcal{H}^{I}$ is that the h-node $H^{\emptyset}_k$ might be split into multiple new h-nodes in $\mathcal{H}^{I}$ and some edges with other h-nodes that were present in $\mathcal{H}^{\emptyset}$, might be removed in $\mathcal{H}^{I}$. 
 \par
 However, according to Algorithm~\ref{alg:appex-train-by-components}, we do not split these new h-nodes rather bind them together to form $H^{I}_k$ that contain the same nodes as $H^{\emptyset}_k$. Therefore, no new edge is being added among other h-nodes. This implies that the partial order of $\mathcal{H}^{\emptyset}$ is also valid for $\mathcal{H}^{I}$.
 %
 %
%
After intervention no new edges are added to the constructed $\mathcal{H}$-graphs, thus we can safely claim that, 
  \begin{equation}
     An_{G_{\overline{I}}}(H_k^{I})  \subseteq An_G(H_k^{\emptyset}) , \forall I \in \mathcal{I}
 \end{equation}
 Since all $\mathcal{H}$-graphs are DAGs and the above condition holds, any valid partial order for $\mathcal{H}^{\emptyset}$ is also a valid partial order for all $\mathcal{H}^{I}, \forall I\in \mathcal{I}$, i.e., they have a common valid partial order.
 

 
 \end{proof}
 

In general, for $I\in  H_k^I$, i.e., when the intervention is inside $H_k^I$,
we utilize interventional datasets and search for minimum size variable set $\mathcal{A}_I \subseteq An_{G_{\overline{I}}}(H_k^I)$ in $G_{\overline{I}}$ such that do-calculus rule-2 satisfies,
\begin{equation}
P(H_k^I\cup \mathcal{A}_{I}|pa(H_k^I\cup \mathcal{A}_{I}), \doo(I))
=P(H_k^I\cup \mathcal{A}_{I}| \doo(pa(H_k^I\cup \mathcal{A}_{I})), \doo(I))
\end{equation}
Then we can train the mechanisms in $H^{I}_k$ to match the following distribution,
\begin{equation}
\label{eq:intv-dist-match-with-intv}
\begin{split}
    P(H_k^I\cup \mathcal{A}_{I}|pa(H_k^I\cup \mathcal{A}_{I}), \doo(I))&=
    Q(H_k^I\cup \mathcal{A}_{I}|\doo(pa(H_k^I\cup \mathcal{A}_{I})), \doo(I))\\
    \implies P(H_k^I\cup \mathcal{A}_{I}| \doo(pa(H_k^I\cup \mathcal{A}_{I})), \doo(I)&=
    Q(H_k^I\cup \mathcal{A}_{I}|\doo(pa(H_k^I\cup \mathcal{A}_{I})), \doo(I))
\end{split}
\end{equation}

\begin{proposition}
\label{appex-th:intv-dist-matches}
Suppose Algorithm \ref{alg:appex-train-by-components}: \textbf{\WG Modular Training} converges
for each h-node in $\mathcal{H}^{\emptyset}$-graph constructed from $G=(\mathcal{V},\mathcal{E})$. Suppose the interventional distribution induced by the deep causal model is $Q_I(V)$ after training on data sets $\mathbf{D}^I, \forall I\in \mathcal{I}$. 
then,
\begin{equation}
    P_I(V)=Q_I(V)
\end{equation}
\end{proposition}

\begin{proof}[Proof Sketch]
	The proof of this Proposition  follows the same route as Proposition~\ref{appex-th:obs-dist-matches-obsdata}. 
	However, we have now access to both observational and interventional datasets and \WG is trained on all these datasets modularly to match every interventional joint distribution.
	An important fact is that even if we have access to $\doo(I), \forall I\in \mathcal{I}$ datasets and we construct multiple $\mathcal{H}^{I}$-graphs, we still follow the topological order of $\mathcal{H}^{\emptyset}$-graph, i.e, H-graph with no intervention. This is valid according to Proposition~\ref{appex-lemma:consistent-hgraph} since a topological order of  $\mathcal{H}^{\emptyset}$ works for all  $\mathcal{H}^I$-graphs even though  $\mathcal{H}^I$ are sparser. Also, any node in $H^I_k$ contains the same set of nodes as in $H^{\emptyset}_k$ for all $k$.

Tian's factorization allows us to express the interventional joint distribution $P_I(V)$ in terms of multiple c-factors.
We divide the c-components corresponding to these c-factors  into two sets. Set-1: the c-component containing the intervention and the c-components in the same h-node. Set-2: the rest of the c-components without any intervention.  We combine the c-factors in both sets as  $H_k^{I}$ and $H_{k'}^I\in \{\mathcal{H}^{I}\setminus H_{k}^I\}$ . Therefore, according to Corollary~\ref{thm-set-of-c-factorization}, $P_{I}(V)$ can be written as:  $P_{pa(C_{i}) \cup I}(C_{i}) \times 
\prod \limits_{H_{k}^I\in \mathcal{H}^{I}}
\prod \limits_{C_{i'}\in H_{k}^I} P_{pa(C_{i'})}(C_{i'})= P_{pa(H_k^{I}) \cup I}(H_k^{I}) 
\times \prod \limits_{H_{k'}^I\in \{\mathcal{H}^{I}\setminus H_{k}^I\}} P_{pa(H_{k'}^{I})}(H_{k'}^{I})$. During the modular training with interventional datasets, \WG matches each of these c-factors and thus matches the interventional joint distribution.

We can consider  any h-node $H^I_k$ as  $H^I_k \in \sigma_0$, i.e., to be either a root h-node  of $\mathcal{H}^I$ or   $ H^I_k \in \sigma_1$ i.e., to be a non-root h-node of $\mathcal{H}^I$. 
For both of these cases, we follow the same approach as the observational case except the fact that we consider h-nodes in the $\mathcal{H}^I$ graph (but the same topological order as $\mathcal{H}^{\emptyset}$), the ancestor set $A_{I}$ in  $G_{\bar{I}}$ and the $\doo(I)$ dataset while matching the interventional distribution for h-nodes.
 Now, for  $H^I_k \in \sigma_0$,  by the construction of the $\mathcal{H}^I$ graph, we can say $ P(H_k^{I}|Pa(H_k^{I}), \doo(I)) = P_{Pa(H_k^{I})\cup I}(H_k^{I}) $. 
 Thus, we match $P_{Pa(H_k^{I})\cup I}(H_k^{I})$ by training the DCM mechanisms in $H^I_k$ by matching $P(H_k^{I}|Pa(H_k^{I}), \doo(I)) = Q_{Pa(H_k^{I}) \cup I}(H_k^{I})$.

For h-nodes $H^I_k \in \sigma_1$, we perform modular training to train these mechanisms by matching an alternative interventional joint distribution $  P(H_k^I\cup \mathcal{A}_{I}|\doo(Pa(H_k^I\cup \mathcal{A}_{I})), \doo(I)) $ with the $\doo(I)$ interventional data. This alternative distribution can be expressed as:
$P_{pa(H^I_k) \cup I}(H^I_k) \times 
\prod \limits_{H_S^I \in \mathcal{A}_I}  \prod\limits_{C_{i'}^+\in H^{I}_S} P_{Pa(C^+_{i'})}(C^+_{i'}) $.
Here the first term correspond to the distribution involving the current h-node $H^I_k$ we are training. The second term corresponds to the partial c-factors located in the ancestors $\mathcal{A}_I$. They are partial $C_i^+$ because $\mathcal{A}_I$ are ancestors of $H^I_k$ in $G_{\bar{I}}$ satisfying the modularity condition~\ref{def:modularity-condition-I}  and  not necessarily containing the full c-component $C_i$. We can equivalently write: $ P_{Pa(H^{I}_k) \cup I}(H^{I}_k) = \frac{ P(H_k^{I}\cup \mathcal{A}_{I}|\doo(Pa(H_k^{I}\cup \mathcal{A}_{I})), \doo(I))}
{   \prod \limits_{H_S^I \in \mathcal{A}_I}  \prod\limits_{C_j^+\subseteq H^{I}_S} P_{Pa(C^+_j)}(C^+_j)}$.
We  match the numerator at  the current training step. Since we follow the topological order of the H-graph, the denominator distributions are matched while training the ancestor h-nodes mechanisms in $\mathcal{H}^I$. Therefore, \WG DCM can match the interventional distribution $P_{Pa(H^{I}_k) \cup I}(H^{I}_k)$. More precisely, $Q_{Pa(H^{I}_k) \cup I}(H^{I}_k)= P_{Pa(H^{I}_k) \cup I}(H^{I}_k)$.

\WG follows the topological order of $\mathcal{H}^{\emptyset}$ and trains all mechanisms in any $H^{\emptyset}_k$. While training the $k$-th h-node, \WG enforces the mechanisms in the h-node to learn all interventional distribution  $P_{Pa(H^{I}_k) \cup I}(H^{I}_k), \forall I\in \mathcal{I}$.
 Therefore, after training the last node in the topological order, \WG modular training matches the joint interventional distribution $P_I(V)$.
We provide the detailed proof below.

%
%

\end{proof}

\begin{proof}
Suppose, intervention $I$ belongs to a specific c-component $C_i$, i.e., $I\in C_i$.
According to Tian's factorization, we can factorize the $\doo(I)$ interventional joint distributions for $G_{\overline{I}}$ causal graph, into c-factors as follows:
\begin{equation}
\label{eq:fact-intv-dist}
    P_I(V)= P_I(\mathcal{H}^{I}) = P_{pa(C_{i}) \cup I}(C_{i}) \times 
    \prod \limits_{H_{k}^I\in \mathcal{H}^{I}}
    \prod \limits_{C_{i'}\in H_{k}^I} P_{pa(C_{i'})}(C_{i'})
\end{equation}
The difference between the c-factorization for $P(V)$ and $P_I(V)$ is that when intervention $I$ is located inside c-component $C_i$, we have $P_{pa(C_i) \cup I}(C_i)$ instead of $P_{pa(C_i)}(C_i)$.
We can divide c-components $\mathcal{C}=\{C_1,\hdots C_t\}$ into disjoint partitions or h-nodes as $H_k^{\emp}=\{C_i\}_{i\in T_k}$ for some $T_k\subseteq [t]$. 

Let, the c-component $C_i$ that contains intervention $I$ belong to hnode $H_k^I$, i.e., $C_i \in H_k^I$.
Following Corollary \ref{thm-set-of-c-factorization},
we can combine the c-factors in each partitions and rewrite R.H.S of Equation~\ref{eq:fact-intv-dist} as:
\begin{equation}
\begin{split}
\label{eq:intv-joint-to-hnode}
P_{pa(C_{i}) \cup I}(C_{i}) \times 
    \prod \limits_{H_{k}^I\in \mathcal{H}^{I}}
    \prod \limits_{C_{i'}\in H_{k}^I} P_{pa(C_{i'})}(C_{i'})    
    = P_{pa(H_k^{I}) \cup I}(H_k^{I}) 
   \times \prod \limits_{H_{k'}^I\in \{\mathcal{H}^{I}\setminus H_{k}^I\}} P_{pa(H_{k'}^{I})}(H_{k'}^{I})
\end{split}    
\end{equation}

%
Now, we prove that we match each of these terms in Equation~\ref{eq:intv-joint-to-hnode} according to the training order $\mathcal{T}$.\\
\textbf{\underline{For any root h-nodes $H_k^{I} \in \sigma_0$}}:\\
Due to the construction of $H^{I}$ graphs in Algorithm~\ref{appex:construct-H-graph}, the following is true for any root nodes, $H_k^{I} \in \sigma_0$.
\begin{equation}
\label{eq:intv-root-intv-cond}
    P_I(H_k^{I}|Pa(H_k^{I})) = P_{Pa(H_k^{I})\cup I}(H_k^{I}) 
\end{equation}
\WG training convergence for the mechanisms in $H_k^{I} \in \sigma_0$. Algorithm~\ref{alg:appex-train-by-components}, line~\ref{line:appex-call-trmod}
ensures that the following matches:
\begin{equation}
\label{appex:intv-eq-root-h-node}
\begin{split}
       P_I(H_k^{I}|Pa(H_k^{I})) &= Q_{Pa(H_k^{I}) \cup I}(H_k^{I})\\
   \implies P_{Pa(H_k^{I}) \cup I}(H_k^{I})&=  Q_{Pa(H_k^{I}) \cup I}(H_k^{I})
\end{split}
\end{equation}

\textbf{\underline{For the h-node $H_k^{I} \in \sigma_1$ with $I \in H^I_k$}}:\\
Now we show that  we can train mechanisms in $H_k^{I}$ by matching $P_I(\mathcal{V})$ c-factors  
with \Li and \Lii datasets.
Let us assume, $\exists \mathcal{A}_I \subseteq \sigma_0$ such that $\mathcal{A}_I = An_{G_{\overline{I}}}(H_k^{I})$, i.e., ancestors of $H_k^{I}$ in the $\mathcal{H}^{I}$-graph that we have already trained with available $D^I$ dataset, $\forall I\in \mathcal{I}$. 

To apply Lemma \ref{thm:sub-graph-factorization} in $G_{\overline{I}}$ with $|I|\geq 0$, consider $V'= H_k^I\cup \mathcal{A}_{I}$ as the focus-set, $\{Pa(V') \cup I\}$ as the action-set. Thus, active c-components: 
$C^+_j \coloneqq C_j\cap V'$. We apply the lemma as below:
\begin{equation}
    \begin{split}
        P(H_k^I\cup \mathcal{A}_{I}|\doo(Pa(H_k^I\cup \mathcal{A}_{I})), \doo(I)) 
        &=
        \prod \limits_{C_i\in H^{I}_{k} }
        P_{pa(C_i)\cup I}(C_i)
        \times \prod \limits_{H_S^I \in \mathcal{A}_I }
        \prod \limits_{C_{i'}^+\in H^{I}_S} P_{Pa(C_{i'}^+)}(C_{i'}^+) 
\end{split}
\end{equation}
Here, the 1st term is the factorization of the current h-node
and the 2nd term is the factorization of the ancestors set.
The intervened variable $I$ is located in the current h-node $H_k^{I}$. Therefore, the factorized c-components, i.e., $C_i\in H^I_k$ has $I$ as intervention along with their parent intervention. The above equation implies:
\begin{equation}
    \begin{split}
      P(H_k^I\cup \mathcal{A}_{I}|\doo(Pa(H_k^I\cup \mathcal{A}_{I})), \doo(I))   &=P_{pa(H^I_k) \cup I}(H^I_k) \times 
       \prod \limits_{H_S^I \in \mathcal{A}_I}  \prod\limits_{C_{i'}^+\in H^{I}_S} P_{Pa(C^+_{i'})}(C^+_{i'}) 
\end{split}
\end{equation}
According to Corollary~\ref{thm-set-of-c-factorization}, we combine the c-factors $P_{pa(C_i)\cup I}(C_i)$ for c-components in $H^I_k$ to form $P_{pa(H^I_k) \cup I}(H^I_k)$.
We continue the derivation as follows:
         \begin{equation}
         \label{eq:intv-joint-match-h1}
    \begin{split}
        \implies P_{Pa(H^{I}_k) \cup I}(H^{I}_k) &= \frac{ P(H_k^{I}\cup \mathcal{A}_{I}|\doo(Pa(H_k^{I}\cup \mathcal{A}_{I})), \doo(I))}
        {   \prod \limits_{H_S^I \in \mathcal{A}_I}  \prod\limits_{C_j^+\subseteq H^{I}_S} P_{Pa(C^+_j)}(C^+_j)}\\
        \implies P_{Pa(H^{I}_k) \cup I}(H^{I}_k) &= \frac{ Q(H_k^{I}\cup \mathcal{A}_{I}|\doo(Pa(H_k^{I}\cup \mathcal{A}_{I})), \doo(I))}
        {
         \prod \limits_{H_S^I \in \mathcal{A}_I}  \prod\limits_{C_j^+\subseteq H^{I}_S}
         Q_{Pa(C^+_j)}(C^+_j)}
           \end{split}
\end{equation}
Here the R.H.S numerator follows from previous line according to Equation~\ref{eq:intv-dist-match-with-intv} since training has converged for the current h-node.
For the R.H.S, denominator,
$\forall H_S^I \in \mathcal{A}_I$ appear before $H_k^I$ in the partial order. 
When we trained h-nodes $H_S^I \in \mathcal{A}_I$ on $P(\mathcal{V})$ and $P_I(\mathcal{V})$ datasets, we matched the joint distribution $P(H_S^I\cup \mathcal{A}_{I}| \doo(pa(H_S^I\cup \mathcal{A}_{I})), \doo(I)), \forall H_S^I \in \mathcal{A}_I$. 
%
According to Lemma~\ref{thm:sub-graph-factorization}, matching these distribution is sufficient to match the distribution at the R.H.S denominator.
Therefore,  our DCM will produce the same distribution as well. This implies that from Equation~\ref{eq:intv-joint-match-h1} we get,
\begin{equation}
   \begin{split}
        P_{pa(H^{I}_k) \cup I}(H^{I}_k)&= Q_{pa(H^{I}_k) \cup I}(H^{I}_k)\\
    \end{split}
\end{equation}
Similarly, we train each h-node following the training order $\mathcal{T}$ and match the distribution in Equation~\ref{eq:intv-joint-to-hnode}. We train the c-factor that contains interventions with our available interventional dataset and the c-factors that do not include any interventions can be trained with $P(V)$ dataset.
This finally shows that,
\begin{equation}
\begin{split}
    P_I(V)=&P_{pa(H_k^{I}) \cup I}(H_k^{I}) 
   \times \prod \limits_{H_{k'}^I\in \{H^{I}\setminus H_{k}^I\}} P_{pa(H_{k'}^{I})}(H_{k'}^{I})\\
   =&Q_{pa(H_k^{I}) \cup I}(H_k^{I}) 
   \times \prod \limits_{H_{k'}^I\in \{H^{I}\setminus H_{k}^I\}} Q_{pa(H_{k'}^{I})}(H_{k'}^{I})\\
   =& Q_I(V)
\end{split}    
\end{equation}
\end{proof}

\begin{figure}[H]
  \centering
     \includegraphics[width=1\linewidth]{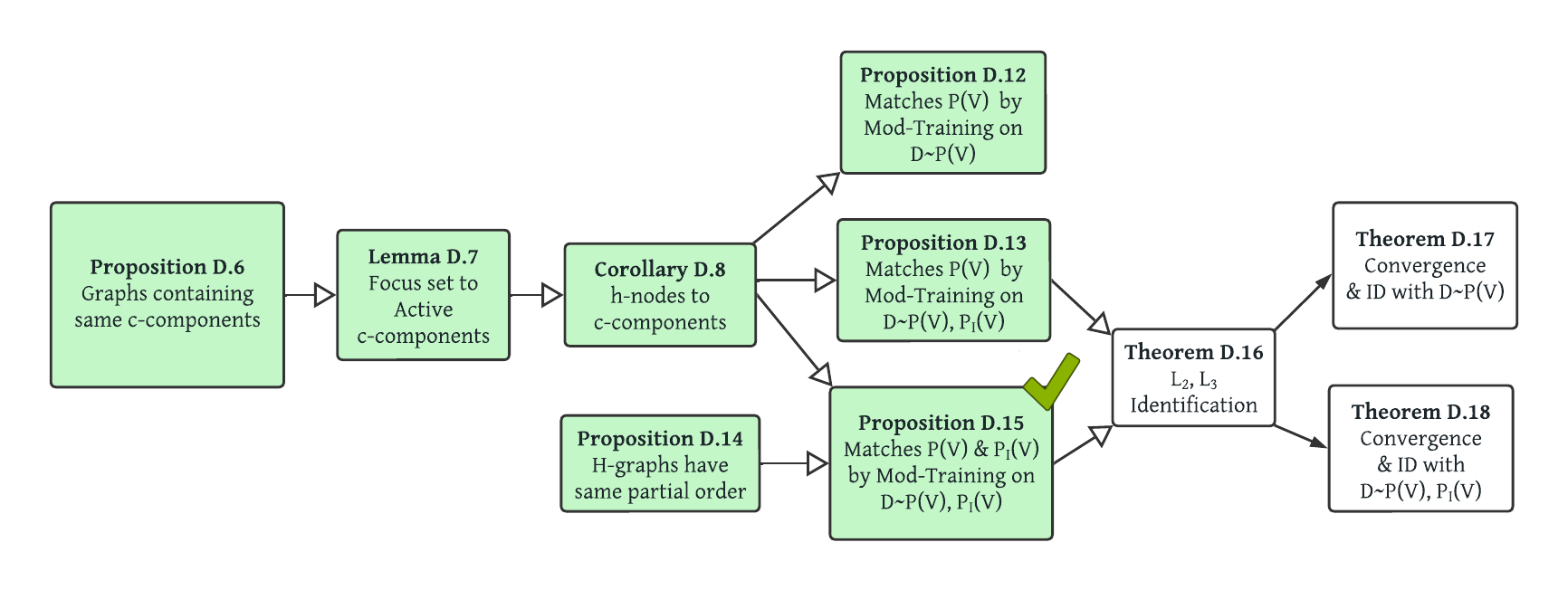}
   \caption{Flowchart of proofs }
   \label{proof-flow-D15}
\end{figure}
\subsection{Identifiability of Algorithm~\ref{alg:appex-train-by-components}:\WG Modular Training}
\label{appex:converge-alg-3-proof}
\begin{theorem}
\label{appex-th:intervention-identifiability}
Let $\mathcal{M}_1$ be the true SCM and Algorithm \ref{alg:appex-train-by-components}: \textbf{\WG Modular Training} converge
for each h-node in $\mathcal{H}$ constructed from $G=(\mathcal{V},\mathcal{E})$ after training on
data sets $\mathbf{D}=  \{\mathcal{D}^I\}_{\forall I\in \mathcal{I}}$ and output the DCM $\mathcal{M}_2$. Then for any $\mathcal{L}_2$ causal query $\mathcal{K}_{\mathcal{M}_1}(\mathcal{V})$, identifiable from $\mathbf{D}$, $\mathcal{K}_{\mathcal{M}_1}(\mathcal{V})= \mathcal{K}_{\mathcal{M}2}(\mathcal{V})$ holds.
\end{theorem}

\begin{proof}
Let $\mathcal{M}_1 = (G=(\mathcal{V},\mathcal{E}), \mathcal{N}, \mathcal{U}, \mathcal{F}, P(.) )$ be the true SCM and $\mathcal{M}_2 = (G, \mathcal{N}', \mathcal{U}', \mathcal{F}', Q(.))$ be the deep causal generative model 
represented by Modular-DCM. 
For any $H^I_k \in \mathcal{H}^I, I\in \mathcal{I}$,  we observe the joint distribution $P(H^I_k \cup \mathcal{A}^I \cup Pa(H^I_k \cup \mathcal{A}^I), \doo(I))$ in the input $D^I$ datasets.
Thus we can train all the mechanisms in the current h-node $H^I_k$ by matching the following distribution from the partially observable datasets:
 \begin{equation}
\begin{split}
\label{Thm:modular-train-converge}
    P(H_k^I\cup \mathcal{A}_{I}|pa(H_k^I\cup \mathcal{A}_{I}), \doo(I))=
    Q(H_k^I\cup \mathcal{A}_{I}|\doo(pa(H_k^I\cup \mathcal{A}_{I})), \doo(I))
\end{split}
\end{equation}
Now, as we are following a valid partial order of the $\mathcal{H}^\emptyset$-graph to train the h-nodes, we train the mechanisms of each h-node to match the input distribution only once and do not update it again anytime during the training of rest of the network. As we move to the next h-node of the partial order for training, we can keep the weights of the Ancestor h-nodes fixed and only train the current one and can successfully match the joint distribution in Equation~\ref{Thm:modular-train-converge}.
In the same manner, we would be able to match the distributions for each h-node and reach convergence for each of them.
\WG Training convergence implies that $Q_I(\mathcal{V})=P_I(\mathcal{V}), \forall I\in \mathcal{I}$ i.e., for all input dataset distributions.
Therefore, according to Theorem~\ref{th:appex-identifiability}, \WG is capable of producing samples from correct interventional that are identifiable from the input distributions.
\end{proof}

\begin{theorem}
\label{th:modular-train-converges-obsdata}
Suppose Algorithm \ref{alg:train-by-components}: \textbf{\WG Modular Training} converges
for each h-node in the $\mathcal{H}$-graph constructed from $G=(\mathcal{V},\mathcal{E})$ and after training on observational dataset $D\sim P(\mathcal{V})$, the observational distribution induced by the DCM is $Q(V)$.  Then, we have $i) P(V)=Q(V)$, and $ii)$ for any $\mathcal{L}_2$  causal query $\mathcal{K}_{\mathcal{M}_1}(\mathcal{V})$ that is identifiable from $\mathbf{D}$, we have  $\mathcal{K}_{\mathcal{M}_1}(\mathcal{V})= \mathcal{K}_{\mathcal{M}2}(\mathcal{V})$
\end{theorem}
\begin{proof}
Theorem~\ref{th:main:modular-train-converges} is restated here. 
The first part of the theorem is proved in Proposition~\ref{appex-th:obs-dist-matches-obsdata}.
The second part can be proved with Theorem~\ref{appex-th:intervention-identifiability}.
\end{proof}

\begin{theorem}
\label{th:appex-modular-train-converges}
Suppose Algorithm \ref{alg:appex-train-by-components}: \textbf{\WG Modular Training} converges
for each h-node in the $\mathcal{H}^\emptyset$-graph constructed from $G=(\mathcal{V},\mathcal{E})$ and after training on observational and interventional datasets $\mathbf{D}^I\sim P_I(\mathcal{V}) \forall I\in \mathcal{I}$, the distribution induced by the DCM is $Q_I(V), \forall I\in \mathcal{I}$.  Then, we have $i) P_I(V)=Q_I(V)$, and $ii)$ for any $\mathcal{L}_2$ causal query $\mathcal{K}_{\mathcal{M}_1}(\mathcal{V})$ that is identifiable from $\mathbf{D}^I, \forall I\in \mathcal{I}$, we have  $\mathcal{K}_{\mathcal{M}_1}(\mathcal{V})= \mathcal{K}_{\mathcal{M}2}(\mathcal{V})$
\end{theorem}
\begin{proof}

The first part of the theorem is proved in Proposition~\ref{appex-th:obs-dist-matches} and Proposition~\ref{appex-th:intv-dist-matches}. Then it is a direct implication of Theorem~\ref{appex-th:intervention-identifiability}, This theorem is equivalent to Theorem~\ref{th:main:modular-train-converges} if we consider $\mathcal{I}=\{\emp\}$.
\end{proof}

\begin{figure}[H]
  \centering
     \includegraphics[width=1\linewidth]{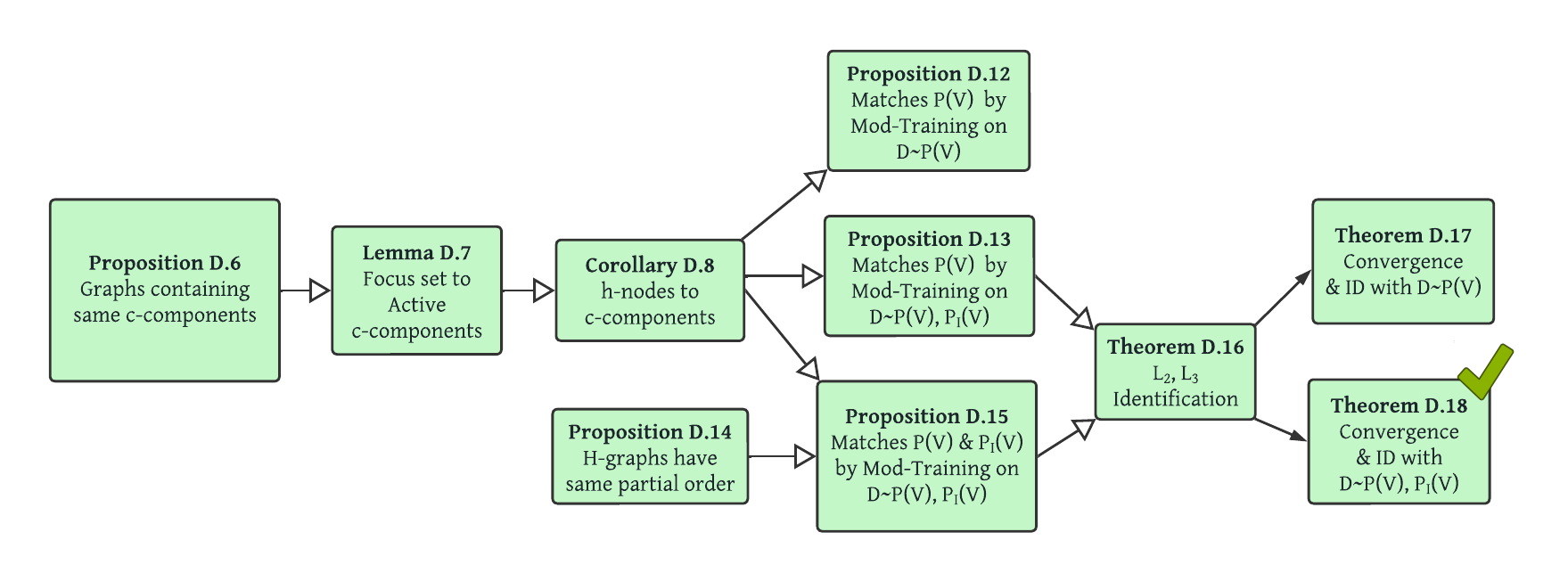}
   \caption{Flowchart of proofs }
   \label{proof-flow-D18}
\end{figure}

\section{ Modular Training on Different Graphs}
\label{appex:section:different graphs}
\subsection{Modular Training Example}
\label{appex:sec-long-hgraph}
\begin{figure}[H]
  \centering
     \includegraphics[width=1\linewidth]{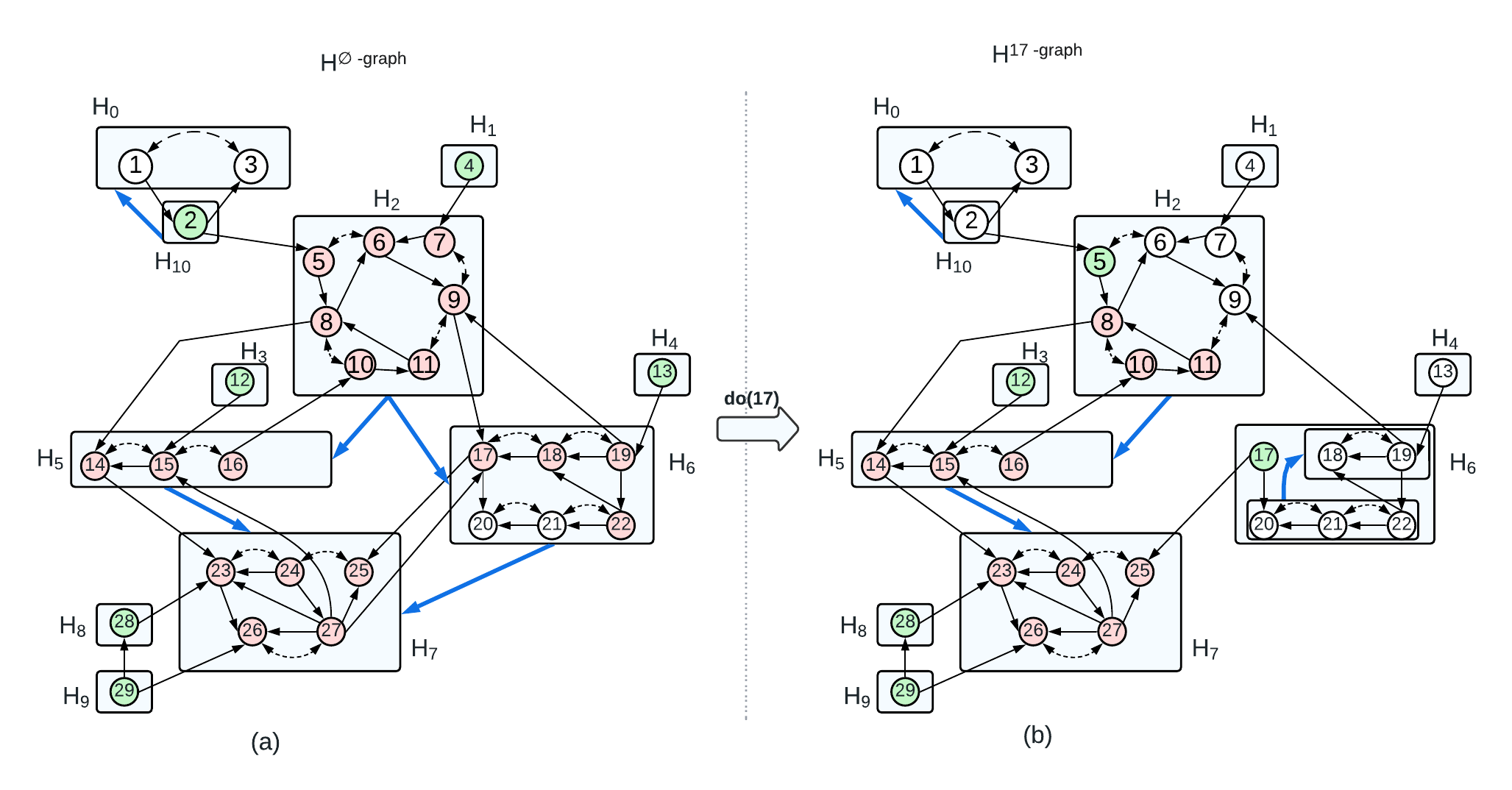}
   \caption{$\mathcal{H}^\emptyset$-graph and $H^{17}$-graph construction}
   \label{appex:fig-long-hgraph}
\end{figure}

In Figure~\ref{appex:fig-long-hgraph}, we construct the $\mathcal{H}^\emptyset$-graph as below. We describe the H-graph edges (thick blue edges) and the backdoor path (thin black edges) responsible for those edges.\\
$H_{10} \rightarrow H_0: 3 \leftarrow 2 \leftarrow 1$\\
$H_2 \rightarrow H_5: 14 \leftarrow 8\leftarrow 11 \leftarrow 10 \leftarrow16,\\
H_2\rightarrow  H_6: 17 \leftarrow 9 \leftarrow 19,\\
H_5 \rightarrow H_7: 23 \leftarrow 14 \leftarrow 15 \leftarrow 27,\\
H_6 \rightarrow H_7: 25 \leftarrow 17 \leftarrow 27$

In Figure~\ref{appex:fig-long-hgraph}, we construct the $H^{17}$-graph as below:\\
$H_{10} \rightarrow H_0: 3 \leftarrow 2 \leftarrow 1$\\
$H_2 \rightarrow H_5: 14 \leftarrow 8\leftarrow 11 \leftarrow 10 \leftarrow16,\\
H_5 \rightarrow H_7: 23 \leftarrow 14 \leftarrow 15 \leftarrow 27,$\\
Now, notice that due to $do(17)$, $H^{17}_6$ gets splitted into two new h-nodes, $[18,19]$ and $[20,21,22]$ with a new edge $[20,21,22] \rightarrow [18,19]$. However, according to our H-graph construction algorithm, we keep these two new h-nodes of $H^{17}$ combined inside $H^{17}_6$ same as $\mathcal{H}^\emptyset$-graph. Therefore, $\mathcal{H}^\emptyset$ and $H^{17}$'s common partial order does not change.

For training $H^\emptyset_{7}$ node: $\{23, 24, 25, 26, 27\}$, we match the following distribution found by applying do-calculus rule 2.
\begin{equation}
\label{eq:larger-obs-expression}
    \begin{split}
        P( 23, 24, 25, 26, 27, 14, 15, 16, 17, 18, 19, 22,5, 6, 7, 8, 9, 10, 11, |\doo(2, 12, 13, 28, 29, 4))
    \end{split}
\end{equation}
In Figure~\ref{appex:fig-long-hgraph}(a), joints are shown as red nodes and their parents as green nodes.
However, consider, we have both observational and interventional datasets from $P(\mathcal{V})$ and $P(\mathcal{V}|\doo(17))$ and we have already trained all the ancestor h-nodes of $H^{17}_{7}$. Then
we can train the mechanisms that lie in $H^\emptyset_{7}$ to learn both observational and interventional distribution by matching a smaller joint distribution compared to Equation~\ref{eq:larger-obs-expression}:
\begin{equation}
    \begin{split}
        P(23, 24, 25, 26, 27, 8, 10, 11, 14, 15, 16|\doo(12, 28, 29, 5, 17))
    \end{split}
\end{equation}
In Figure~\ref{appex:fig-long-hgraph}(b), joints are shown as red nodes and their parents as green nodes.
We see that the number of red nodes is less for $H^{17}$ graph compared to $\mathcal{H}^\emptyset$ graph when we were matching the mechanisms in h-node, $H^\emptyset_{7}$.
%
%
%
%


\section{ Experimental Analysis}
\label{appex:experimental-analysis}
In this section, we provide implementation details and algorithm procedures of our \WG training.

\subsection{Training Details and Compute}
\label{sec: train-det-comp}

We performed our experiments on a machine with an RTX-3090 GPU. The experiments took 1-4 hours to complete. We ran each experiment for $300$ epochs. We repeated each experiment multiple times to observe the consistent behavior. Our datasets contained $20-40K$ samples, and the batch\_size was $200$, and we used the ADAM optimizer. For evaluation, we generated 20k fake samples after a few epochs and calculated the target distributions from these 20k fake samples and 20k real samples. We calculated TVD and KL distance between the real and the learned distributions. For Wassertein GAN with gradient penalty, we used LAMBDA\_GP=$10$. We had learning\_rate = $5 * 1e-4$. We used Gumbel-softmax with a temperature starting from $1$ and decreasing it until $0.1$. We used different architectures for different experiments since each experiment dealt with different data types: low-dimensional discrete variables and images. Details are provided in the code. For low-dimensional variables, we used two layers with 256 units per layer
and with BatchNorm and ReLU between each layer. Please check our code for architectures of other neural networks such as encoders and image generators
%
%
%

%

\subsection{Complexity Evaluation}
\label{appex:complex-eval}

Suppose, a causal graph has $N$
 variables. Without modularization, we have to match the joint distribution containing $N$  (might be large) number of low and high dimensional variables in a single training phase. Matching that joint distribution with deep-learning models, and a complicated confounded causal structure could be difficult since we are attempting to minimize a very complicated loss function for a very large neural network. Our proposed method allows us to reduce the complexity of this problem tremendously by modularizing the training process to c-components. The size of a c-component is generally a lot smaller than the whole graph. Thus, even though we have to train mechanisms in a c-component together and match a joint distribution involving high and low dimensional variables, the complexity will be much lower. Without our approach, there is no existing work that can modularize and simplify the training process for a causal graph with latents.

To achieve a deep causal generative model (DCM), given a causal graph of $N$ nodes, it is required to train $N$ neural networks. However, our nearest benchmark NCM, trains all $N$ networks together at the same time.
While our method trains only the networks that belong to a single h-node. Thus, during a training phase, the maximum number of networks NCM has to train together is $O(N)$ and in our case, it is $O(|${Largest h-node}$|)$ which is in most cases $O(|$Largest c-component$|)$.

\subsection{Image Mediator Experiment}
\label{sub:appex-img-med}

In this section, we provide additional information about the experiment described in Section~\ref{sec:image-mediator}.
The front-door graph has been instrumental for a long time in the causal inference literature. However, it was not shown before that modular training with high dimensional data was possible, even in the front door graph. This is why we demonstrate the utility of our work on this graph. 

\begin{figure}[H]
\captionsetup[subfigure]{justification=centering}
\begin{subfigure}{0.22\linewidth}
\centering
\includegraphics[width=1.0\linewidth]{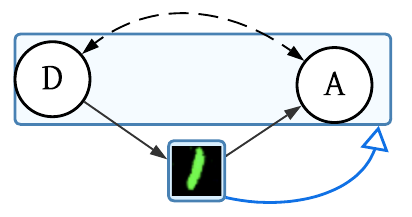}
      	\caption{Frontdoor causal graph w/ image mediator}
   \label{fig:appex-mnist-frontdoor}
\end{subfigure}
\hspace{4mm}
\centering
\begin{subfigure}{0.4\linewidth}
\includegraphics[width=0.9\linewidth]{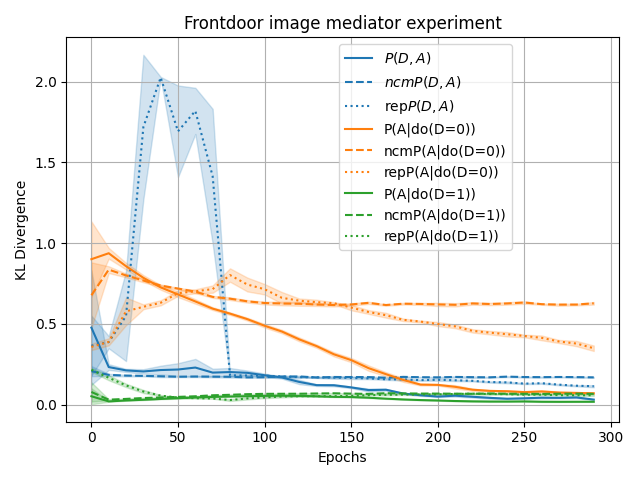}
	\caption{Training converges matching $P(D,A)$ and $P(A|do(D))$}
\label{fig:appex-image-mediator-benchmark}
\end{subfigure}

\begin{subfigure}{0.32\linewidth}
\centering
\includegraphics[width=1.0\linewidth]{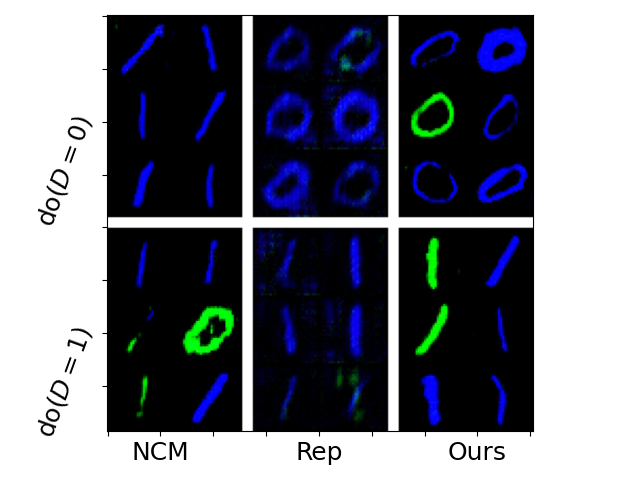}
\centering
      	\caption{MNIST Image generation  \\ comparison}
   \label{fig:digit-benchmarks2}
\end{subfigure}
\begin{subfigure}{0.32\linewidth}
\centering
\includegraphics[width=1.0\linewidth]{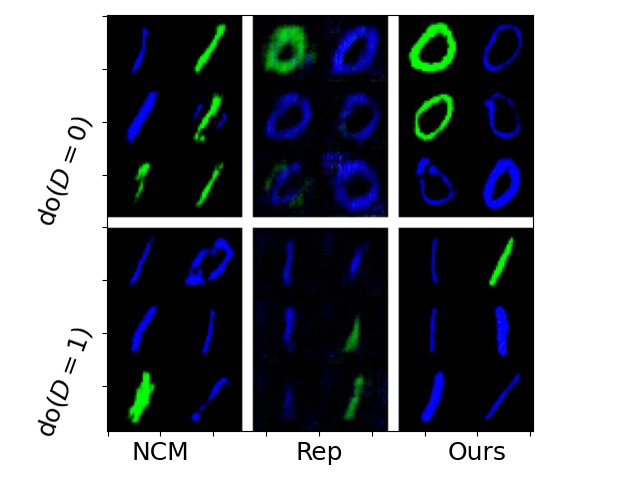}
      	\caption{MNIST Image generation  \\comparison}
   \label{fig:digit-benchmarks3}
\end{subfigure}
\begin{subfigure}{0.32\linewidth}
\centering
\includegraphics[width=1.0\linewidth]{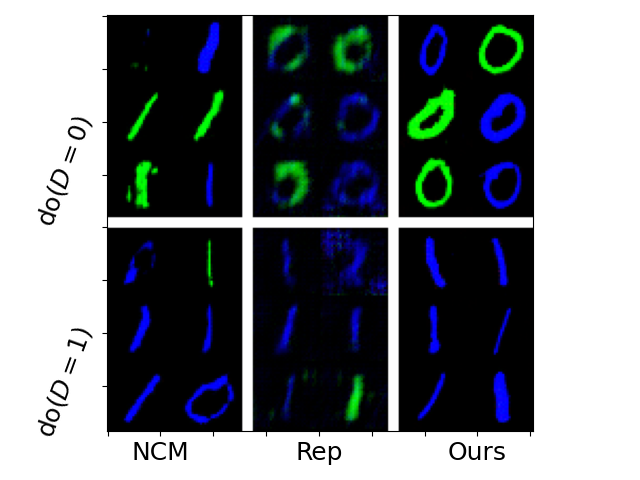}
      	\caption{MNIST Image generation \\ comparison}
   \label{fig:digit-benchmarks4}
\end{subfigure}
    \caption{Modular Training on frontdoor causal graph with training order:  $\{I\}\rightarrow \{D,A\}$}
    \label{fig:appex-colored-mnist-1-experiment}
\end{figure}

We have domain $D=[0,1]$, Image size=$3\times32\times32$ and $C=[0,1,2]$.  Let $U_0,e_1,e_2,e_3$ are randomly generate exogenous noise. $D=  U_0 + e_1, Image= f_2(D,e_2), A= f_3(Image, e_3, U_0)$. $f_2$ is a function which takes $D$ and $e_2$ as input and produces different colored images showing $D$ digit in it. $f_3$ is a classifier with random weights that takes $U_0, e_3$ and $Image$ as input and produces $A$ such a way that $|P(A|do(D=0)) - P(A|D=0)| , |P(A|do(D=1))-P(A|do(D=0))| and |P(A|D=1) - P(A|D=0)|$ is enough distant.
The digit color can be considered as exogenous noise.
The target is to make sure that the backdoor edge $D\leftrightarrow A$ and the causal path from $D$ to $A$ is active.
Since we have access to $U_0$ as part of the ground truth, we can calculate the true value of $P(A|do(D))$ with the backdoor criterion~\citep{pearl1993bayesian}:
\begin{equation*}
  P(A|do(D))= \int_{U_0} P(A|D,U_0) P(D|U_0)  
\end{equation*}
During training, $U_0$ is unobserved but still, the query is identifiable with the front door criterion~\citep{pearl2009causality}. $\mathrm{Image}$ is a mediator here.
\begin{equation*}
P(A|do(D))= \int_{Image} P(Image|D) \sum_{D'} P(A|D',Image)P(D')    
\end{equation*}

However, this inference is not possible with the identification algorithm since it requires image distribution. 
But Modular-DCM can achieve that by producing $\mathrm{Image}$ samples instead of learning the explicit distribution. %
If we can train all mechanisms in the \WG DCM to match $P(D, A, I)$, we can produce correct samples from $P(A|do(D))$. 
We construct the \WG architecture with a neural network $\mathbb{G}_D$ having fully connected layers to produce $D$, a deep convolution GAN $\mathbb{G}_I$ to generate images, and a classifier $\mathbb{G}_A$ to classify MNIST images into variable $A$ such that $D$ and $A$ are confounded.
Now, for this graph, the corresponding $\mathcal{H}$-graph is $[I]\rightarrow[D,A]$. 
Thus, we first train $\mathbb{G}_{I}$ by matching $P(I|D)$. 
Next, to train $\mathbb{G}_D$ and $\mathbb{G}_A$, we should match the joint distribution $P(D, A, I)$ since $\{I\}$ is ancestor set $\mathcal{A}$ for c-component $\{D, A\}$.
GAN convergence becomes difficult using the joint distribution loss since the losses generated by low and high dimensional variables are not easily comparable and it is non-trivial to find a correct re-weighting of such different loss terms. 
To the best of our knowledge, no current causal effect estimation algorithm can address this problem since there is no estimator that does not contain explicit image distribution, which is practically impossible to estimate.
To deal with this problem, we map samples of $I$ to a low-dimensional representation, $RI$ with a trained encoder and match $P(D, RI, A)$ instead of $P(D, Image, A)$. 

Note that, we use the mechanism training order $[I] \rightarrow [D, A]$ specified by the H-graph (Algorithm~\ref{construct-H-graph}) to match the joint distribution $P(D, Image, A)$. It is not feasible to follow any other sequential training order such as $[D] \rightarrow [Image] \rightarrow [A]$ as training them sequentially with individual losses can not hold the dependence in $D \leftrightarrow A$.
We compare our performance with NCM in Figure~\ref{fig:appex-colored-mnist-1-experiment}. We implemented NCM on our architectures as it could not be directly used for images. 
For estimating the FID scores, we generated 2050 samples from each method and calculated the FID score compared to the original images using a method proposed in~\cite{Seitzer2020FID}.

\subsection{MNIST Diamond Graph}
Here we discuss the data generating process of the MNIST diamond graph.
We have considered matching the joint distribution for the following diamond graph.
$I_1 \rightarrow Digit \rightarrow I_2 \rightarrow Color ; I_1 \leftrightarrow Color \leftrightarrow Digit$.
Here $I_1$ and $I_2$ are image nodes and the rest are discrete.
$I_1, Digit, Color$ belong to the same c-component.
To generate semi-synthetic data for this graph, we first uniformly sample $U_1$ and $U_2$ where $I_1 \leftarrow U_1 \rightarrow Digit$ and $Digit \leftarrow U_2 \rightarrow Color$. 
Next, we set  $I_1.color$ according to $U_1$. Then we pick a digit image from the MNIST dataset of $I_1.digit$ and color it with $I_1.color$.   Next we generate values for $Digit$ consistent with $I_1.digit$ while adding some confounding variable $U_2$. We pick another MNIST image with $Digit$ and color it with some random color. Finally we set the value of $Color$ with $I_2.color$ and $U_2$.


\subsection{Performance on Real-world COVIDx CXR-3 Dataset}
\label{sec:appex-covid}

\begin{figure}[t!]
\begin{subfigure}{0.25\linewidth}
\centering
\includegraphics[width=0.7\linewidth]{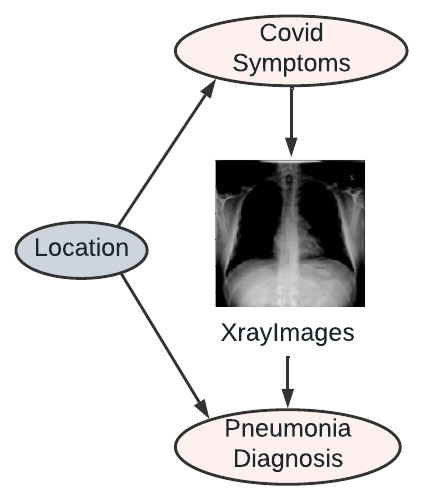}
      	\caption{COVIDx CXR-3 graph}
   \label{fig:appex-xray-graph}
\end{subfigure}
\begin{subfigure}{0.35\linewidth}
\centering
\includegraphics[width=0.5\linewidth]{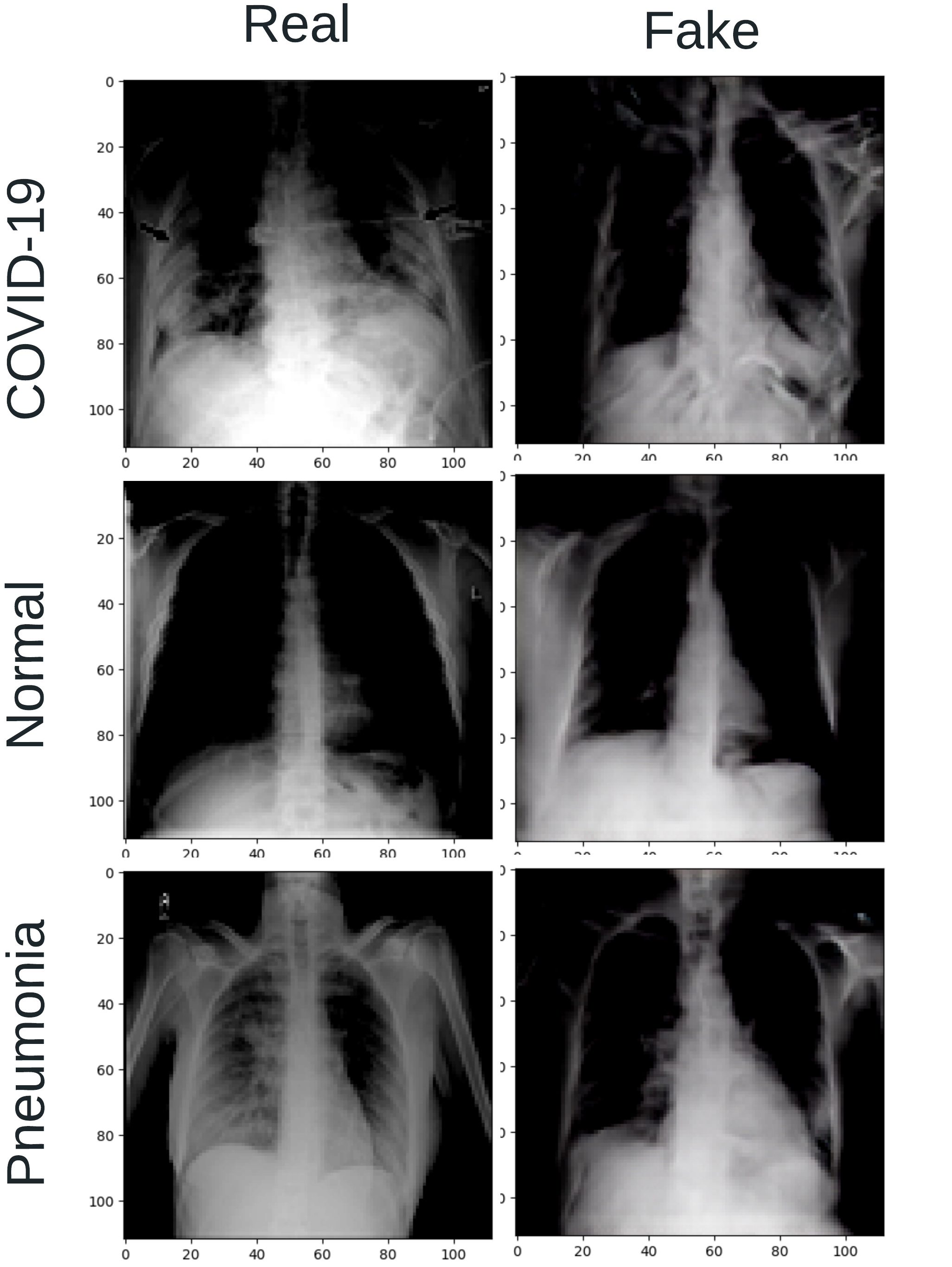}
      	\caption{Real vs fake}
   \label{fig:xray-real-fake}
\end{subfigure}
\begin{subfigure}{0.45\linewidth}
\includegraphics[width=0.8\linewidth]{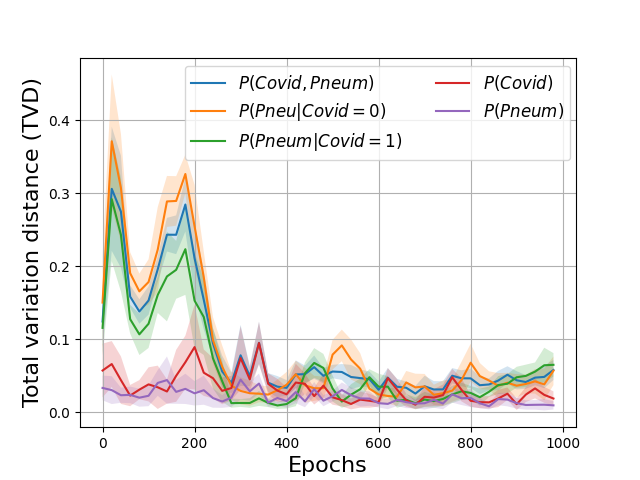}
 \caption{Training converges with low TVD}
\label{fig:xray-tvd}
\end{subfigure}
     \caption{\WG converges with pre-trained model on COVIDx CXR-3 dataset.
    }
    \label{fig:xray-image-experiment}
\end{figure}

\subsubsection{Real-world COVIDx CXR-3 Dataset}
To demonstrate the convergence behavior of \WG on real high-dimensional datasets, we conduct a case study with the COVIDx CXR-3~\citep{Wang2020} dataset in this section. This dataset contains 30,000 chest X-ray images ($Xray$) with Covid $(C)$ and pneumonia $(N)$ labels from over 16,600 patients located in 51 countries.
Even though there is no ground truth causal graph associated with this dataset, we consider the same motivational setup we discussed in Figure~\ref{fig:xray-a}: $C\rightarrow Xray \rightarrow N, C \leftrightarrow N$. 
We assume the graph to be consistent with the dataset since it does not impose conditional independence restrictions on the joint distribution $P(C, Xray,N)$. 
Therefore, we expect our modular training algorithm to correctly match the observational joint distribution. 
We discuss the reasoning behind each edge in Appendix~\ref{sec:appex-covid}.
We aim to learn that if a patient is randomly picked and 
intervened with Covid (hypothetically), how likely will they be diagnosed with pneumonia, i.e., ${P(N|do(C))}$?
To match the joint distribution ${P(C, Xray, N)}$, we follow the modular training order: $[\mathbb{G}_{Xray}] \rightarrow [\mathbb{G}_{C}, \mathbb{G}_{N}]$. Instead of training $\mathbb{G}_{Xray}$ from scratch, we use a pre-trained model~\citep{Carbone2023} that can be utilized to produce Xray images corresponding to ${C\in [0,1]}$ input. Next, we train $\mathbb{G}_{C}$ and $\mathbb{G}_{N}$ together since they belong to the same c-component. Since the joint distribution contains both low and high-dimensional variables, we map $\mathrm{Xray}$ to a low-dimensional representation $\mathrm{Rxray}$ with an encoder and match ${P(C, Rxray, N)}$.
\par
\emph{Evaluation:}
Figure~\ref{fig:xray-real-fake} shows images for the original dataset (left) and output images (right) from the pre-trained model.
In Figure~\ref{fig:xray-tvd}, we plot the total variation distance (TVD) of ${P(C), P(N)}$, ${P(N|C), P(N, C)}$. We observe that TVD for all distributions is decreasing.  The average treatment effect, i.e., the difference between $E[P(N|do(C=1))]$ and  $E[P(N|do(C=0))]$ is in $[0.05, 0.08]$ after convergence. This implies that intervention with Covid increases the likelihood of being diagnosed with Pneumonia. However, these results are based on this specific COVIDx CXR-3 dataset and should not be used to make medical inferences without expert opinion.

\subsubsection{Detailed discussion on Covidx CXR-3}

In this section, we provide some more results of our experiment on COVIDx CXR-3 dataset~\citep{Wang2020}. This dataset contains 30,000 chest X-ray images with Covid $(C)$ and pneumonia $(N)$ labels from over 16,600 patients located in 51 countries.
The X-ray images are of healthy patients $(C=0, N=0)$, patients with non-Covid pneumonia $(C=0, N=1)$, and patients with Covid pneumonia $(C=1, N=1)$. 
X-ray images corresponding to COVID non-pneumonia $(C=1, N=0)$ are not present in this dataset as according to health experts those images do not contain enough signal for pneumonia detection. However, to make the GAN training more smooth we replaced a few $(C=1, N=1)$ real samples with $(C=1, N=0)$  dummy samples. We also normalized the X-ray images before training.

Note that the causal effect estimates obtained via this graph may not reflect the true causal effect since the ground truth graph is unknown and there may be other violations of assumptions such as distribution shift and selection bias.   
 In order to demonstrate the convergence behavior of \WG on real high-dimensional datasets, we consider the causal graph shown in Figure~\ref{fig:appex-xray-graph}.
 However, observe that this graph does not impose conditional independence restrictions on the joint distribution $P(C, Xray,N)$. 
If our mentioned assumptions (including no selection bias, etc.) are correct, we expect \WG to correctly sample from interventional distribution after training by Theorem \ref{th:main:modular-train-converges}. 
Therefore, we expect our modular training algorithm to correctly match the observational joint distribution.

Our reasoning for using this causal graph is as follows: we can assume that Covid symptoms determine the X-ray features and the pneumonia diagnosis is made based on the X-rays. Thus we can add direct edges between these variables.
A patient’s location is hidden and acts as a confounder because a person's socio-economic and health conditions in a specific location might affect both the likelihood of getting Covid and being properly diagnosed with Pneumonia by local health care. The X-ray images are done by chest radiography imaging examination. Due to the standardization of equipment, we assume the difference in X-ray data across hospital locations is minor and can be ignored. Thus,  $\mathrm{Location} \not\rightarrow \mathrm{XrayImages}$.

To obtain the low dimensional representation of both real and fake X-ray images, we used a Covid conditional trained encoder. Instead of training $\mathbb{G}_{Xray}$ from scratch, we use a pre-trained model~\citep{Carbone2023} that can be utilized to produce Xray images corresponding to ${C\in [0,1]}$ input. Note that, this pre-trained model takes value 0 for Covid, 1 for normal, and 2 for Pneumonia as input and produces the corresponding images. 
If a fake Covid sample indicates Covid=1, we map it to the 0 input of the pre-trained GAN.
If a fake Covid sample indicates Covid=0, this might be either mapped to 1 (normal) or 2 (Pneumonia). Instead of randomly selecting the value, we use the real Pneumonia sample to decide this (either 1 or 2). After that, we produce X-ray images according to the decided input values.
Since we are using the GAN-generated fake samples for Covid=1, the computational graph for auto grad is not broken. Rather the mentioned modification can be considered as a re-parameterization trick.


\begin{figure}[t!]
\begin{subfigure}{0.33\linewidth}
    \includegraphics[width=1\linewidth]{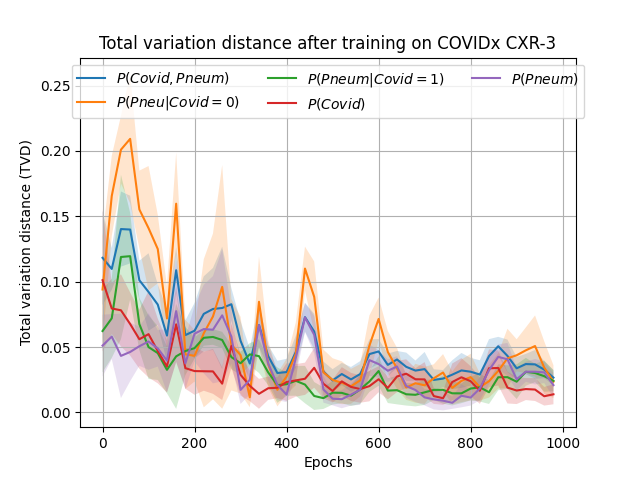}
\end{subfigure}
\begin{subfigure}{0.32\linewidth}
    \includegraphics[width=1\linewidth]{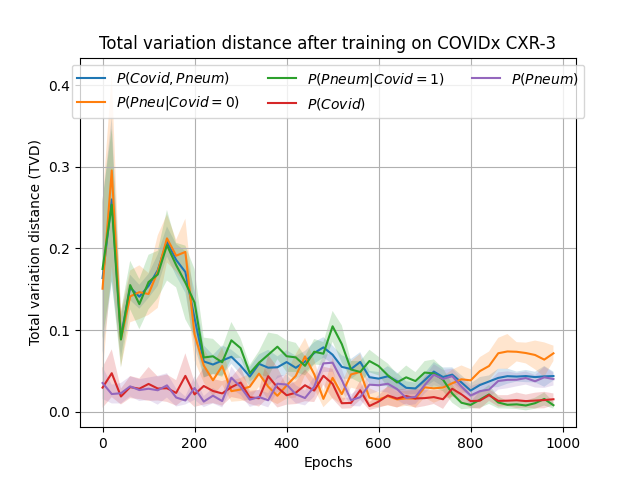}
\end{subfigure}
\begin{subfigure}{0.32\linewidth}
    \includegraphics[width=1\linewidth]{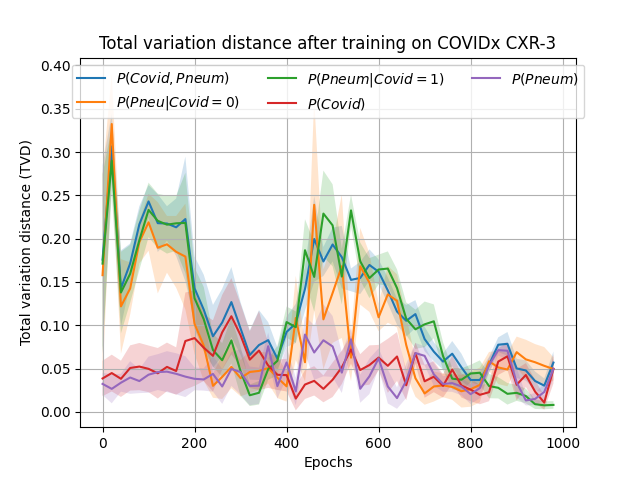}
\end{subfigure}
    \caption{Total variation distance plots show \WG converges  on COVIDx CXR-3 dataset. (consecutive 20 epochs were averaged)}
    \label{appex-fig:xray-tvd-plots}
\end{figure}

\begin{figure}[t!]
    \centering
    \includegraphics[width=0.8\linewidth]{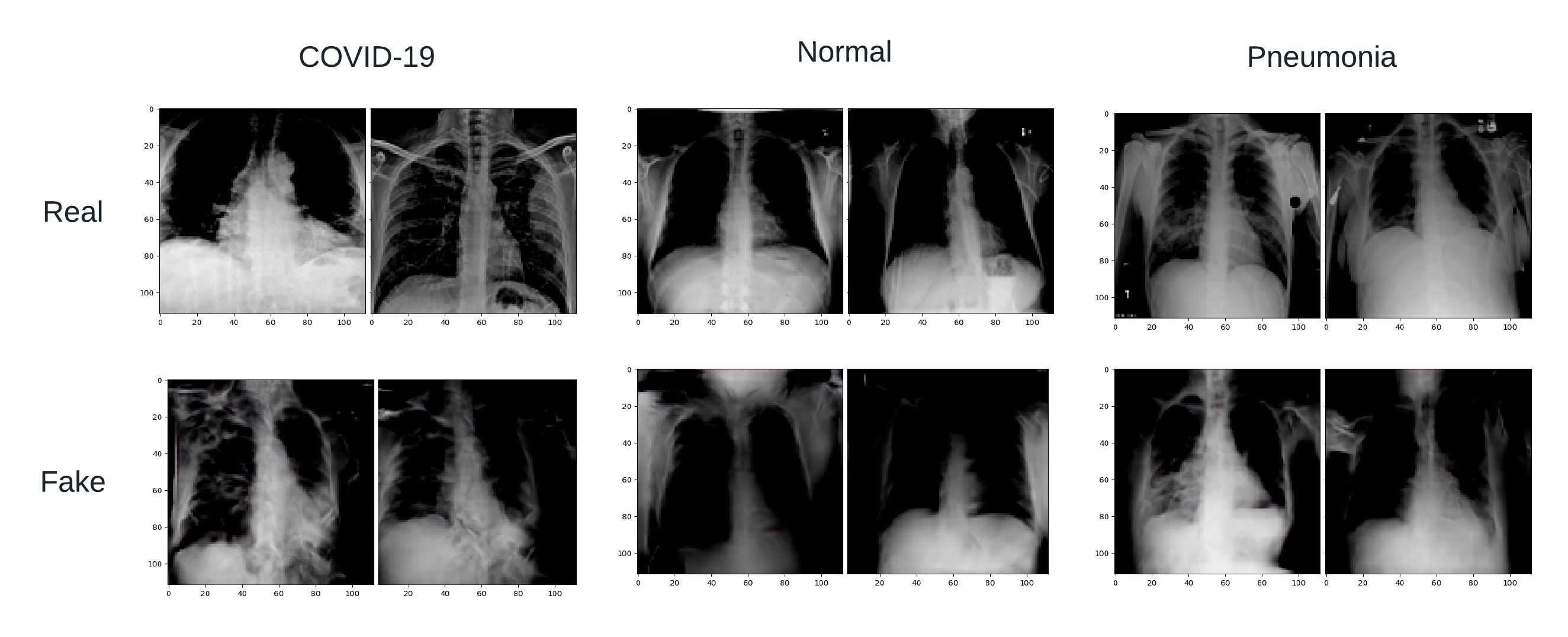}
    \caption{Real images from dataset vs pre-trained GAN generated images}
    \label{fig:enter-label}
\end{figure}

\subsection{Invariant Prediction on CelebA-HQ}
To reflect the distribution shift in $P(Sex)$, we divide image samples from the CelebA-HQ dataset into train and test domains as the table in Figure~\ref{fig:celeb-graph-dist} and the actual number of samples are given in Table~\ref{tab:celeba-dataset}. In the test domain, $P(Sex)$ changes while $P(Eyeglass)$ stays fixed.

The prediction of $Eyeglass$ from $Image$ is done by learning the probability distribution $P(Sex|Image)$. Note that, we would like the prediction of $Eyeglass$ to be independent of $Sex$ and $Domain$.  This might not work if we learn $P(Eyeglass|Image)$ or $P(Eyeglass|Image, Sex)$ since conditioning will make the prediction depend on $Domain$. Thus, we train \CLb classifier on $\mathcal{D}[Eyeglass, Image]\sim P(Eyeglass,Image|\doo(Sex))$ following the approach suggested in~\cite{subbaswamy2019preventing}.

Since $\{Eyeglass,Sex\}$ and $\{Image\}$ belong to different c-components, we can train models $\mathbb{G}_{Eyeglass}, \mathbb{G}_{Sex}$ together and use a pre-trained model for $\mathbb{G}_I$.
Therefore, we only have to train $\mathbb{G}_A$ and $\mathbb{G}_S$. We utilize \WG's ability to incorporate a pre-trained image generation model, \IG~\citep{shen2020interpreting} which can generate impressive human faces in its causal generative models. 
We generate $10k$ samples of $[Eyeglass', Image']\sim P(Eyeglass,Image|\doo(Sex))$. Intervention on $Sex$ attribute will make $Eyeglass$ independent from both $Sex$ and $Domain$. 
Finally, the prediction would be independent of the distribution shift in $P(Sex|Domain)$.

We used the \IG that uses pre-trained StyleGAN from the repository: \url{https://github.com/genforce/interfacegan}.
To filter incorrect images, we used a pre-trained classifier to  from this repository: \url{https://github.com/clementapa/CelebFaces_Attributes_Classification} to filter the inconsistent images generated from \IG.

\begin{table}[H]
    \caption{Number of samples in training and test dataset}
    \label{tab:celeba-dataset}
\centering
\begin{tabular}{|c|cc|}
\hline
          & \multicolumn{2}{c|}{Train}                 \\ \hline
          & \multicolumn{1}{c|}{Eyeglass=0} & Eyeglass=1 \\ \hline
Sex=0   & \multicolumn{1}{c|}{3200}       & 100    \\ \hline
Sex=1 & \multicolumn{1}{c|}{ 1000}       & 1080     \\ \hline
          & \multicolumn{2}{c|}{Test}                  \\ \hline
Sex=0   & \multicolumn{1}{c|}{400}       & 180     \\ \hline
Sex=1 & \multicolumn{1}{c|}{600}       & 100     \\ \hline
\end{tabular}
\end{table}

\begin{figure}[H]
  \centering
     \includegraphics[width=1\linewidth]{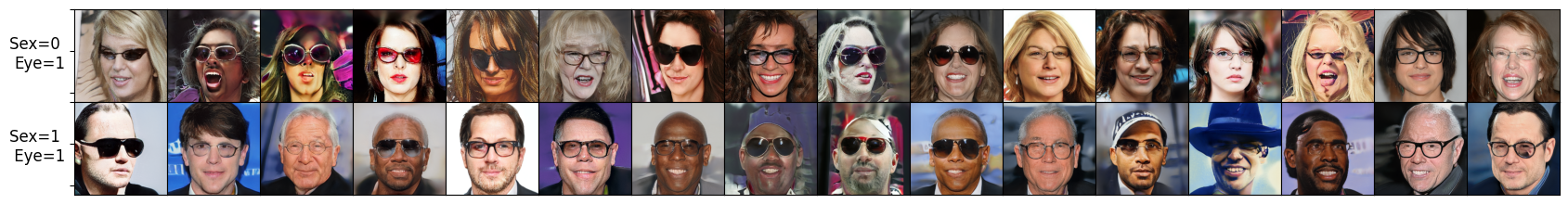}
          \includegraphics[width=1\linewidth]{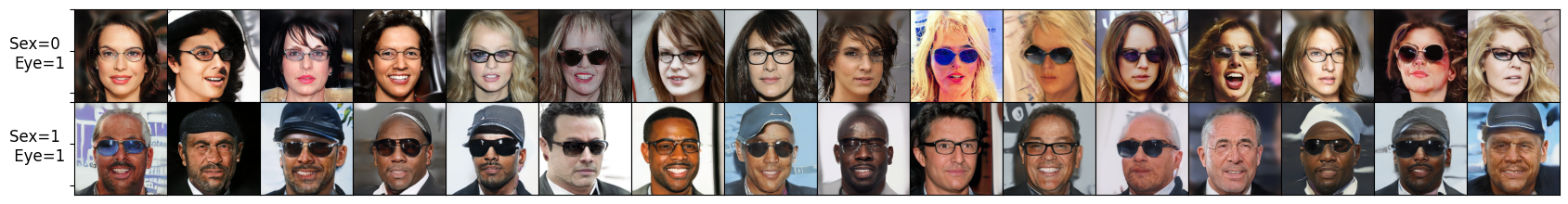}
               \includegraphics[width=1\linewidth]{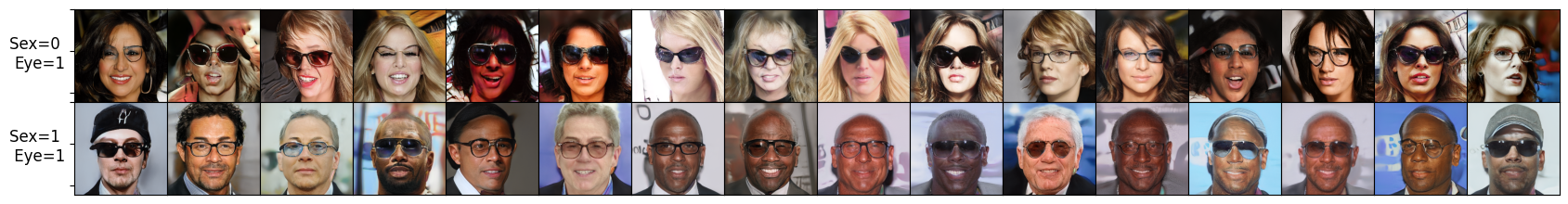}
   \caption{Image samples generated by \IG for $P(Image|Sex, Eyeglass=1)$}
   \label{appex:eye1-sample}
\end{figure}


\begin{figure}[H]
  \centering
     \includegraphics[width=0.8\linewidth]{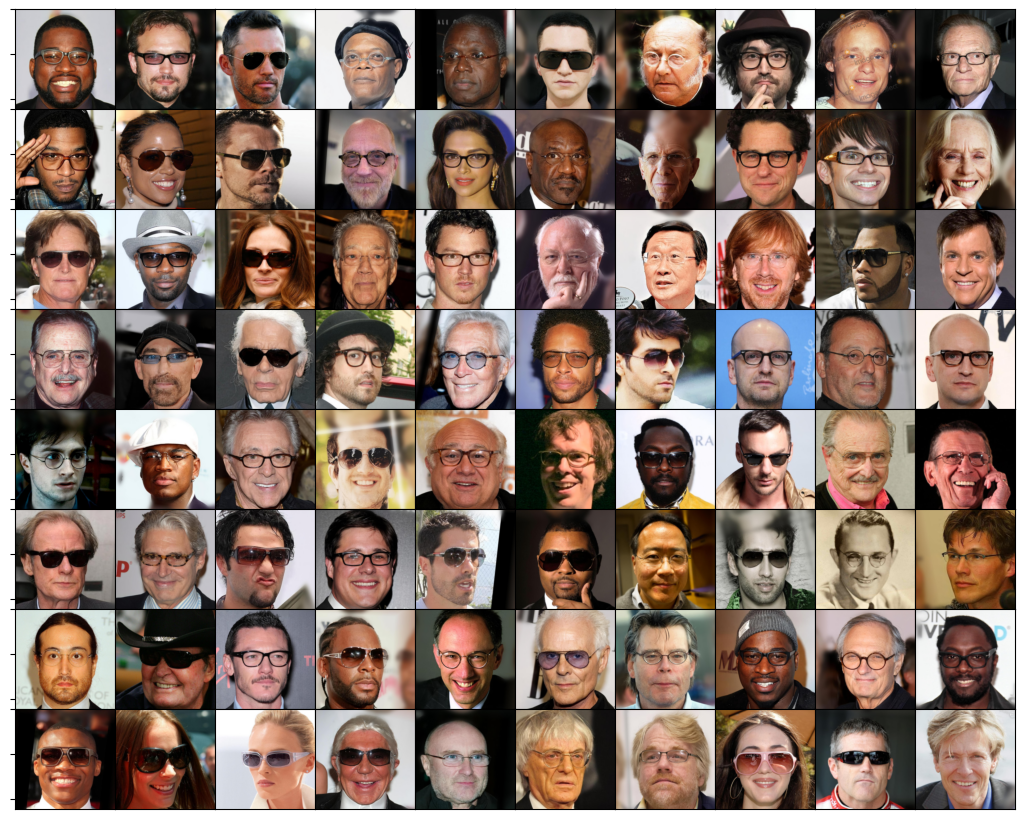}
   \caption{Image samples for $P(Image|Eyeglass=1)$ from the CelebA-HQ dataset. We observed $P(Sex=1|Eyeglass=1)$ is around 0.91 implying that in the training data, there is a high correlation between Sex=Male and wearing eyeglass.}
   \label{appex:i_given_eye}
\end{figure}

\subsection{Asia/Lung Cancer Dataset }
\label{appex-sec-asia-experiment}
\begin{figure}[H]
\begin{subfigure}{0.25\linewidth}
\hspace{-0mm}
   \includegraphics[width=1.0\linewidth]{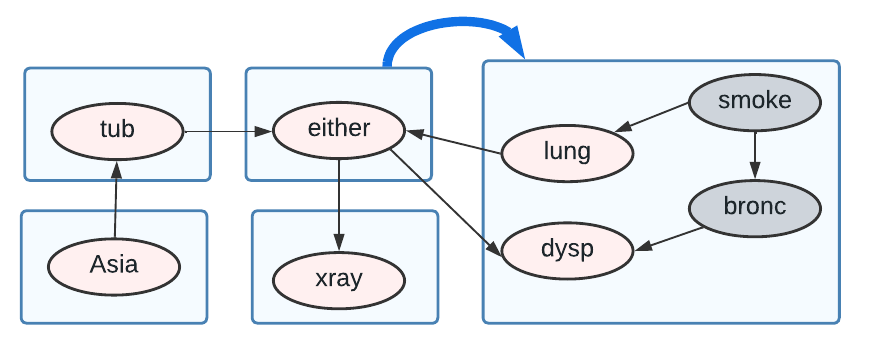}
   \caption{Asia causal graph}
   \label{asia-graph}
   \end{subfigure}
\hspace{2mm}
\begin{subfigure}{0.35\linewidth}
  \centering
    \includegraphics[width=1\textwidth]{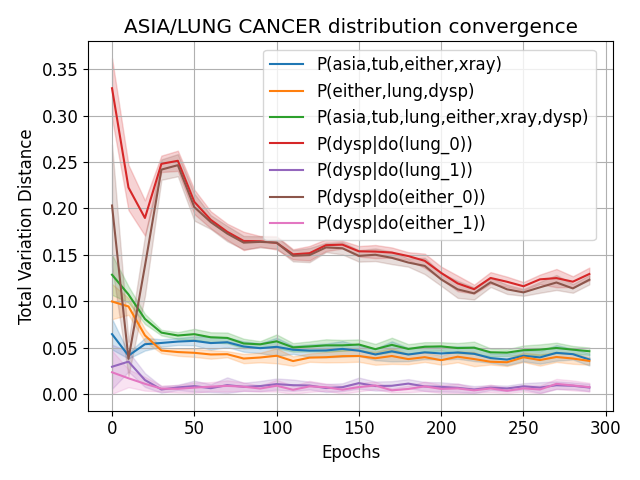}
    \caption{Asia Total Variation Distance}
    \label{tvd-asia}
    \end{subfigure}
\begin{subfigure}{0.35\linewidth}
 \includegraphics[width=1.0\linewidth]{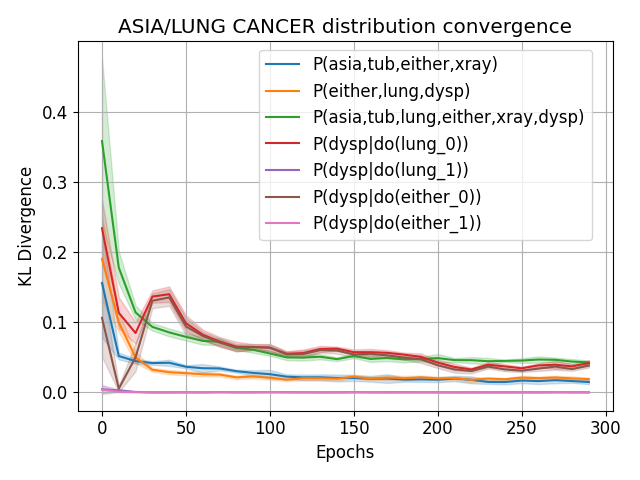}
   \caption{KL divergence of Asia experiment}
   \label{kl-asia}
\end{subfigure}
    \caption{Modular Training on Asia Dataset}
\end{figure}

\par \emph{\textbf{Asia Dataset.}} 
We evaluate our algorithm performance on ASIA Dataset from bnlearn repository \citep{scutari2021bayesian}. The purpose of this experiment is to show that modular training can learn the joint distribution of the Asia dataset formed as semi-Markovian and correctly produces samples from identifiable $\mathcal{L}_2$ distributions.
To check the effectiveness of \WG for a semi-Markovian causal model, we hide "smoke" and "bronc" variables in the observational dataset as shown in Figure~\ref{asia-graph}. This action gives us a causal graph with a latent confounder between the "lung" and the "dysp" variables. 
The $\mathcal{H}$ graph nodes are indicated by the square box containing the variables. According to the algorithm, all $\mathcal{H}$-nodes are disconnected except $[either]\rightarrow [lung,dysp]$. Therefore, we first start training the mechanisms of $asia, tub, either, xray$ and then separately but in parallel train the mechanisms of $lung,dysp$. 
Here we can also use pre-trained $either$ while we train $lung,dysp$ to match the distribution $P(lung,dysp, either|tub)$. 
For evaluation, we generated samples from $P(dysp|\doo(lung))$ and $P(dysp|\doo(either))$ distributions from Modular-DCM. We can calculated $P(dysp|\doo(lung))$ with front-door adjustment and $P(dysp|\doo(either))$ with back-door adjustment using the real dataset samples. In Figure~\ref{tvd-asia},~\ref{kl-asia}, we can see that our partial training is working well with all of the distributions converging to low TVD and KL loss.

\subsection{Real-world: Sachs Protein Dataset}
\label{appex:sachs-section}

%
%
   %
%
%
%

%
\begin{figure}[t!]


\begin{subfigure}{0.2\linewidth}
   \centering
\includegraphics[width=1\linewidth]{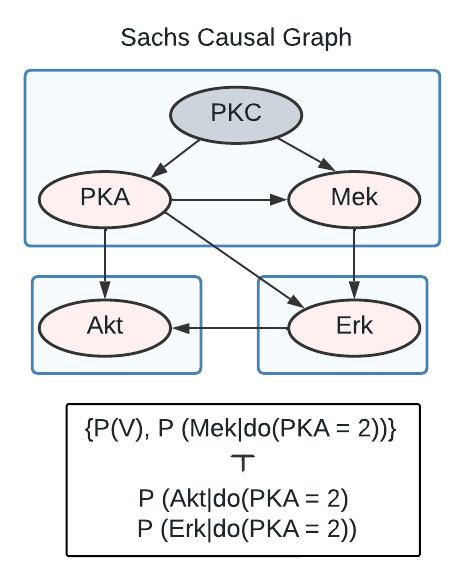}
      	\caption{Causal graphs}
   \label{sachs-graph}
\end{subfigure}
\begin{subfigure}{0.4\linewidth}
   \centering
\includegraphics[width=1\linewidth]{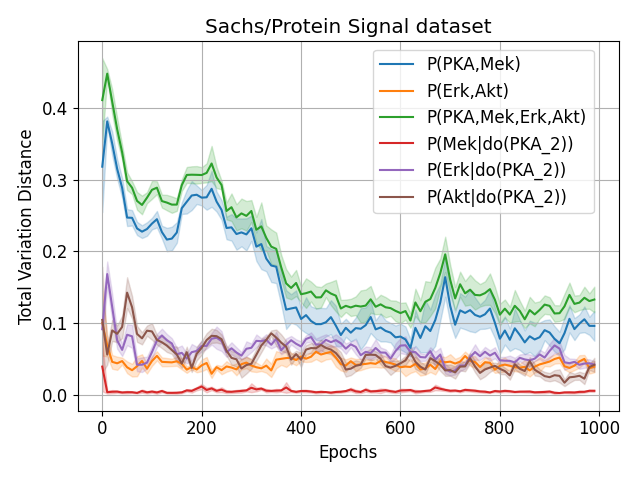}
      	\caption{TVD Convergence on Sachs dataset}
   \label{fig:sachs-tvd}
\end{subfigure}
\begin{subfigure}{0.4\linewidth}
   \centering
\includegraphics[width=1\linewidth]{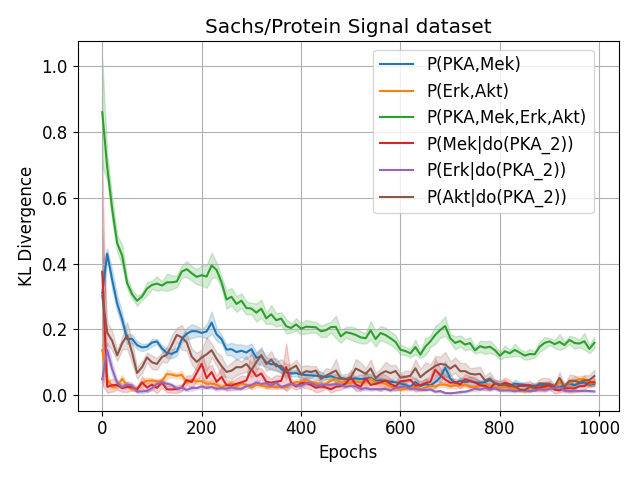}
      	\caption{KL convergence on Sachs dataset}
   \label{fig:sachs-KL}
\end{subfigure}
%
%
\begin{subfigure}{0.4\linewidth}
   \centering
\includegraphics[width=0.9\linewidth]{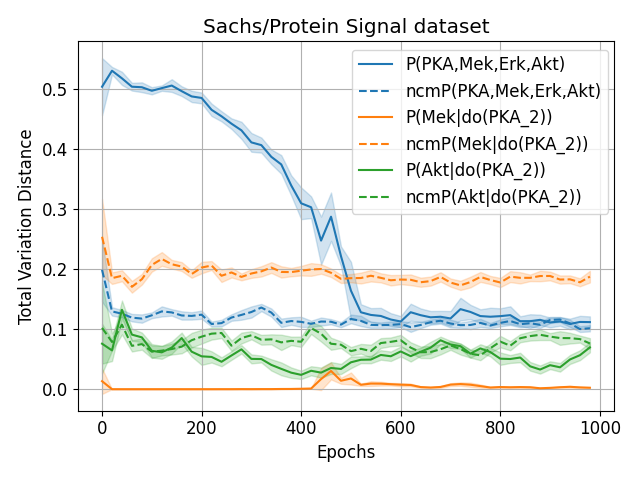}
      	\caption{TVD convergence on Sachs dataset}
   \label{fig:sachs-tvd-benchmarks}
\end{subfigure}
\begin{subfigure}{0.45\linewidth}
   \centering
\includegraphics[width=0.8\linewidth]{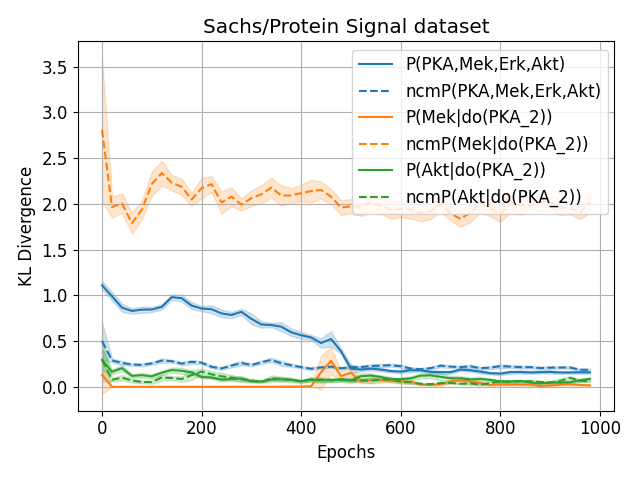}
      	\caption{KL convergence on Sachs dataset}
   \label{fig:sachs-kl-benchmarks}
\end{subfigure}
  \caption{Benchmark and Real-world datasets}
\end{figure}
For completeness, we test both \WG and NCM performance on a low-dimensional real-world Sachs dataset~\citep{sachs2005causal}, which contains a protein signaling causal graph and is 
%
given in Figure~\ref{sachs-graph}.
The goal is to illustrate \WG's capability of utilizing multiple partial $\mathcal{L}_1, \mathcal{L}_2$ datasets.
We considered the observational dataset $D_1\sim P(PKA, Mek, Erk, Akt)$ and the interventional dataset $D_2\sim P(Mek|\doo(PKA=2))$. 
The interventional dataset with $PKA=2$ is chosen since it has a large number of samples. 
Here we intentionally hide variable $PKC$ and considered it as a confounder. Hence, $P(Mek|\doo(PKA=2))$, $P(Akt|\doo(PKA=2)$ and $P(Erk|\doo(PKA=2))$ are non-identifiable from only $P(V)$. 
According to Corollary~\ref{id-intv},
$P(V), P(Mek|\doo(PKA=2))$ make these distributions identifiable.
More precisely, if we have access to $P(V)$ and $P_{PKA}(Mek)$ only, then its sufficient to identify, $P_{PKA}(Mek, Erk, Akt)$.
We have datasets $D_1, D_2$ that are sampled from $P(V), P(Mek|\doo(PKA=2))$ for training.
If we convert the Sachs graph into the train graph $\mathcal{H}$ in  Figure~\ref{sachs-graph},
we see that for only the interventional dataset, we have to train the mechanism of Mek. This is because other variables except Mek belong to different hnodes or training components. Here,
\begin{equation}
    \begin{split}
        &P_{PKA}(Mek, Erk, Akt)\\
        &= P_{PKA}(Mek) P_{PKA}(Erk|Mek) P_{PKA} (Akt|Erk, Mek)\\
        &= P_{PKA}(Mek) P(Erk|Mek, PKA) P(Akt | PKA,Erk, Mek)
    \end{split}
\end{equation}
Therefore, we train \WG i.e., the DCM to match both $P(V)$ and $P(Mek|\doo(PKA))$. We $i)$ first train
$\mathbb{G}_{H_0}=[\mathbb{G}_{PKA}, \mathbb{G}_{Mek}]$ with both $D_1$ and $D_2$, $ii)$ next, train $\mathbb{G}_{H_1 \cup H_2}= [\mathbb{G}_{Erk}, \mathbb{G}_{Akt}]$ with only $D_1$.
In Figure~\ref{fig:sachs-tvd} and ~\ref{fig:sachs-KL},
\WG converges by training on both $P(V)$ and $P(Mek|\doo(PKA=2))$ datasets. We compared the distributions $P(Akt|\doo(PKA=2)$ and $P(Erk|\doo(PKA=2))$ implicit in \WG generated samples with the Sachs \Lii-dataset distributions and observed them matching with low TVD and KL loss. 
Even though, we dont observe $Erk$ and $Akt$ in $D_2$, with modular training, we can still train the mechanisms with $P(V)$ and sample correctly from their \Lii- distributions. This reflects the transportability of Modular-DCM. During \WG training, we can use pre-trained models of $\{PKA, Mek\}, \{Akt\}$ or $\{Erk\}$ since they are located in different hnodes.

\textbf{\emph{Sachs dataset performance comparison with NCM}:}
In Figure~\ref{fig:sachs-tvd-benchmarks} and Figure~\ref{fig:sachs-kl-benchmarks}, we compare and show the convergence of both \WG and NCM with respect to total variation distance and KL-divergence. We observe that for low-dimensional variables, we perform similarly to DCM or better in some cases. However, they do not have the ability to utilize pre-trained models like we do.
Besides, unlike NCM, we do not need to run the algorithm again and again for each identifiable queries. Thus, when queries are identifiable, our algorithm can be utilized as an efficient method to train on datasets involving low-dimensional variables.

\section{  Algorithms \& Pseudo-codes }

\begin{algorithm}[H]
\footnotesize
\caption{\footnotesize Construct\_Hgraph($G$)}
\begin{algorithmic}[1]
 \STATE {\bfseries Input:} Causal Graph $G$
\STATE $\mathcal{C} \leftarrow \mathrm{get\_ccomponents}(G)$
\STATE Create nodes $H_k=C_k$ in $\mathcal{H}$, $\forall C_k \in \mathcal{C} $
\FOR{{\bfseries each} $H_s,H_t \in \mathcal{H}$ such that $s\neq t$}
\IF{$P(H_t|\doo(pa(H_t) \cap H_s))$ \\
$\neq P(H_t|pa(H_t) \cap H_s)$}
\label{alg2:do-calculus-check}
\STATE $\mathcal{H}.add(H_s \rightarrow H_t)$
\ENDIF
\ENDFOR
\STATE $\mathcal{H} \leftarrow$  $\mathrm{Merge}$($\mathcal{H}$, $cyc$), $\forall cyc \in Cycles(\mathcal{H})$
\label{alg:h-graph-mergecycles}
\STATE {\bfseries Return:} $\mathcal{H}$
\end{algorithmic}
\label{construct-H-graph}
\end{algorithm}

\begin{algorithm}[H]
\small
\caption{isIdentifiable($G, \mathcal{I}, query$)}
	\label{procedure2}
\begin{algorithmic}[1]
\STATE {\bfseries Input:} Causal Graph $G=(\mathcal{V}$, $\mathcal{E})$,
Interventions = $I$,
Causal query distribution= $query$
\IF{type($query$)=Interventional}
\STATE {\bfseries{Return}} Run\_ID($G,query$) or hasSurrogates($G,query,I$)
\ENDIF
\end{algorithmic}
\end{algorithm}

\begin{algorithm}[H]
\small
\caption{RunGAN($G, \mathbb{G}, V_{\mathbf{K}}, I , N$)}
	\label{alg:RunGAN}
\begin{algorithmic}[1]
\STATE {\bfseries Input:}Causal Graph $G=(\mathcal{V}$, $\mathcal{E})$, DCM $\mathbb{G}$, target variable set $V_{\mathbf{K}}$, Intervention $I$, Pre-defined noise $N$.
\FOR{$V_i, V_j \in V_{\mathbf{K}}$ such that $i<j$}
    \IF{$V_i, V_j$ has latent confounder}
   \STATE $z \sim p(z)$\\
   \STATE $conf[V_i] \gets Append(conf[V_i],z) $\\
   \STATE $conf[V_j] \gets Append(conf[V_j],z) $
    \hspace{4mm}// Assigning same confounding noise [fix for multiple confounders]
     \ENDIF
\ENDFOR
\FOR{$V_i \in V_{\mathbf{K}}$ in causal graph,$G$ topological order}
\IF{$V_i \in I.keys()$}
 \STATE $v_i= I[V_i]$ \hspace{4mm}// Assigning intervened value 
\ELSE
 \STATE $par= get\_parents(V_i, G$)\\
\IF{$V_i \in N.keys()$}
 \STATE $exos, conf, gumbel = N[V_i]$
\ELSE
 \STATE $exos \sim p(z)$\\
 \STATE $conf=conf[V_i] $\\
 \STATE $gumbel= \emptyset $. \hspace{3mm} //New Gumbel noise will be assigned during forward pass
\ENDIF
 \STATE $v_i= \mathbb{G}_{\theta_i}(exos, conf, gumbel, \mathbf{\hat{v}}_{par})$
\ENDIF
 \STATE $\mathbf{\hat{v}} \gets Append(\mathbf{\hat{v}}, v_i)$\\
\ENDFOR
 \STATE {\bfseries{Return}} Samples \textbf{v} or Fail
\end{algorithmic}
\end{algorithm}

\begin{algorithm}[H]
\small
\caption{Evaulate\_GAN($G, \mathbb{G}, \mathcal{I}, query$)}
	\label{alg:EvaulateCausalGAN}
\begin{algorithmic}[1]
\STATE {\bfseries Input:}Causal Graph $G=(\mathcal{V}$, $\mathcal{E})$, DCM= $\mathbb{G}$,
Available Interventions = $\mathcal{I}$,
Causal query distribution=$query$
\IF{isIdentifiable$(G, \mathcal{I}, query)$ = False}
	\STATE {\bfseries Return:} Fail
	\ENDIF
	\IF{type(query)= observation}
	\STATE $Y= Extract(query)$
    \STATE $samples \gets $  RunGAN($G, \mathbb{G}, [Y], \emptyset,  \emptyset$)
    \ELSIF{type(query)= Intervention}
    \STATE $Y, (X,x):=$ Extract$(query)$
    \STATE $samples \gets $  RunGAN($G,\mathbb{G}, [Y], \{X:x\}, \emptyset $)
    \ENDIF
    \STATE {\bfseries{Return}} $samples$
\end{algorithmic}
\end{algorithm}

\end{document}